\documentclass{article}

\usepackage{microtype}
\usepackage{graphicx}
\usepackage{booktabs} % for professional tables

\usepackage{hyperref}

\usepackage{subcaption}

\usepackage[accepted]{icml2025}

\usepackage{amsmath}
\usepackage{amssymb}
\usepackage{mathtools}
\usepackage{amsthm}

\theoremstyle{plain}
\newtheorem{theorem}{Theorem}[section]

\theoremstyle{definition}

\theoremstyle{remark}

\usepackage[textsize=tiny]{todonotes}

\def \inputset {\mathcal{X}}
\def\reasonset{\mathcal{Z}}

\def \reasontask {f}

\def \evalloss {l}

\def \norm#1{\|#1\|_2}

\def \model {f_{\theta}}

\def\vx{{\textbf{x}}}
\def\vy{{\textbf{y}}}

\def \reasoningtrace {CoT}

\def\simpletohard{S2H generalization}
\def\textgeneralizable{S2H-generalizing}
\def\nontextgeneralizable{non S2H-generalizing}

\DeclareMathAlphabet{\mathbfsf}{\encodingdefault}{\sfdefault}{bx}{n}
\DeclareMathAlphabet{\mathsfit}{\encodingdefault}{\sfdefault}{m}{sl}
\DeclareMathAlphabet      {\mathsb}{OT1}{cmss}{sbc}{n}

\def\smfttext{$\mathsb{Text}$}
\def\smftimage{$\mathsb{Image}$}
\def\imageviatext{$\mathsb{Image\:\!\text{-}via\:\!\text{-}\!Text}$}
\def\imagetext{$\mathsb{Text\text{+}\;\!\:\!\!Image}$}

\def\mixft{$\mathsb{Mix}$}
\def\cmft{$\mathsb{Mix}\textbf{+}$}
\def\cimageviatext{\imageviatext$\textbf{+}$}
\def\alcmft{$\mathsb{Align}\:\!\text{-}\;\!\!\mathsb{Mix}\textbf{+}$}

\def\tw#1{$\mathsb{(TW)}$#1}
\def\al#1{$\mathsb{Align}\:\!\text{-}\;\!\!$#1}

\def\contmixft{$\mathsb{(TW)}$ \mixft}
\def\contcmft{$\mathsb{(TW)}$ \cmft}
\def\contalcmft{$\mathsb{(TW)}$ \alcmft}

\def \evalitoa {l_{(I; S)}}

\def \evalitothard {l^{(H)}_{(I\#; T)}}
\def \evalittoahard {l^{(H)}_{(I, \#T; S)}}
\def \evalitoahard {l^{(H)}_{(I; S)}}

\def\data{\vx}
\def\datalabel{\vy}

\def\datatext{\vx^{(t)}}
\def\dataimage{\vx^{(i)}}

\newcounter{resultcounter}

\newcommand{\tableread}{{\emph{Table Readout}}}
\newcommand{\gridnav}{{\emph{Grid Navigation}}}
\newcommand{\visualanalogy}{{\emph{Visual Analogy}}}
\newcommand{\consecutivetableread}{{\emph{Consecutive Table Readout}}}

\newcommand{\simple}{{\textsc{simple}}}
\newcommand{\hard}{{\textsc{hard}}}

\usepackage{minitoc}
\usepackage{hyperref}
\usepackage{url}
\usepackage{natbib}
\usepackage{microtype}
\usepackage{graphicx}
\usepackage{booktabs} % for professional tables

\usepackage{subcaption}

\usepackage{algorithm}
\usepackage{algorithmic}

\usepackage{amsmath}
\usepackage{amssymb}
\usepackage{mathtools}
\usepackage{amsthm}
\usepackage{subcaption}

\usepackage{pifont}
\theoremstyle{plain}

\theoremstyle{definition}
\theoremstyle{remark}

\usepackage{todonotes}
\usepackage{graphicx}
\usepackage{subcaption}
\usepackage{multirow}

\usepackage{varioref}
\usepackage{hyperref}
\usepackage{xcolor}
\hypersetup{
    colorlinks,
    linkcolor={red!50!black},
    citecolor={blue!50!black},
    urlcolor={blue!80!black}
}
\usepackage[capitalize,noabbrev]{cleveref}
\usepackage[shortlabels]{enumitem} 

\definecolor{mgreen}{RGB}{57, 197, 157}

\usepackage[most]{tcolorbox}
\tcbset{
  observation/.style={
    enhanced,
    colback=yellow!20,
    colframe=orange!70!black,
    boxrule=1pt,
    arc=4pt,
    outer arc=4pt,
    boxsep=5pt,
    left=4pt,
    right=4pt,
    top=4pt,
    bottom=4pt,
    fonttitle=\bfseries,
  }
}

\icmltitlerunning{Generalizing from SIMPLE to HARD Visual Reasoning: Can We Mitigate Modality Imbalance in VLMs?}

\begin{document}

\twocolumn[
\icmltitle{Generalizing from SIMPLE to HARD Visual Reasoning: \\ Can We Mitigate Modality Imbalance in VLMs?}

%% For partial TOC
\doparttoc 
\faketableofcontents 
% \part{} 

\icmlsetsymbol{equal}{*}

\begin{icmlauthorlist}
\icmlauthor{Simon Park}{equal,pli}
\icmlauthor{Abhishek Panigrahi}{equal,pli}
\icmlauthor{Yun Cheng}{equal,pli}
\icmlauthor{Dingli Yu}{pli}
\icmlauthor{Anirudh Goyal}{meta}
\icmlauthor{Sanjeev Arora}{pli}
\end{icmlauthorlist}

\icmlaffiliation{pli}{Princeton Language and Intelligence, Princeton University}
\icmlaffiliation{meta}{Meta AI}

\icmlcorrespondingauthor{}{\{juhyunp, ap34, yc6206\}@princeton.edu}

\vskip 0.3in
]

\printAffiliationsAndNotice{\icmlEqualContribution}

\begin{abstract}

\looseness-1
Vision Language Models (VLMs) are impressive at visual question answering and image captioning. But they underperform on multi-step visual reasoning---even compared to LLMs on the same tasks presented in text form---giving rise to perceptions of {\em modality imbalance} or {\em brittleness}. Towards a systematic study of such issues, we introduce a synthetic framework for assessing the ability of VLMs to perform algorithmic visual reasoning, comprising three tasks: {\tableread}, {\gridnav}, and {\visualanalogy}. Each has two levels of difficulty, {\simple} and {\hard}, and even the {\simple} versions are difficult for frontier VLMs. We propose strategies for training on the {\simple} version of tasks that improve performance on the corresponding {\hard} task, i.e., simple-to-hard (S2H) generalization. This controlled setup, where each task also has an equivalent text-only version, allows a quantification of the modality imbalance and how it is impacted by training strategy. We show that 1) explicit image-to-text conversion is important in promoting {\simpletohard} on images, by transferring reasoning from text; 2) conversion can be internalized at test time. We also report results of mechanistic study of this phenomenon. We identify measures of gradient alignment that can identify training strategies that promote better {\simpletohard}. Ablations highlight the importance of chain-of-thought \footnote{Code is available at \href{https://github.com/princeton-pli/VLM_S2H/}{VLM-S2H PLI codebase}}.

\end{abstract}

\section{Introduction}
\label{sec:intro}
\looseness-1
Many Vision Language Models (VLMs) (e.g., LLaVA-series \citep{liu2023llava,liu2023LLaVA1.5,liu2024llavanext}) fuse an LLM with visual encoders which allows them to harness the impressive reasoning abilities of pre-trained LLMs towards solving visual reasoning tasks~\citep{monajatipoor-etal-2023-metavl,carbune2024chart,zhang2024vision}. However, VLMs are usually felt to exhibit more {\em brittle} reasoning than the underlying LLM, and recent works have tried to understand this as a \textbf{modality imbalance} problem \citep{peng2022balanced, pmlr-v162-huang22e, Fan_2023_CVPR, Wei_2024_CVPR}. For example, presenting the task in an image form can lead to a lower performance than when the same task is presented in a text form \citep{zhang2023lost,zhang2024mathverse, wang2024picture, zhang2024cross, fu2024isobench}. Mitigating this modality imbalance is still an open problem. 
\begin{figure*}[t]
    \centering
    \includegraphics[width=0.99\linewidth]{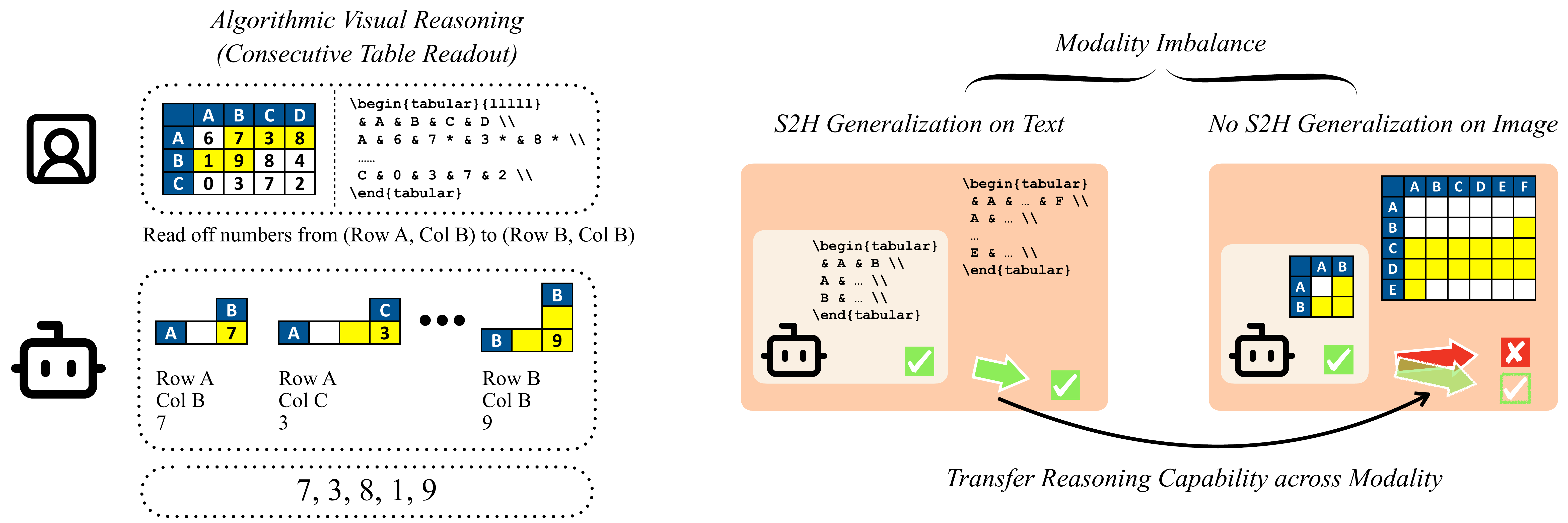}
    \captionof{figure}{ \textbf{(Left) Example Data Point for {\consecutivetableread}.} Input table can be provided as an image or LaTeX code. The task is to sequentially read numbers from a start cell to an end cell in row major order. \textbf{(Right)  Illustration of Key Concepts} using examples from {\consecutivetableread}. We observe that current models can S2H generalize on text -- when trained to read short sequences from small LaTeX-formatted tables, the models can read longer paths from larger tables, also provided in LaTeX code. However, they fail to length-generalize on images. To address the generalization gap and imbalanced learning of different modalities, our goal is to transfer the generalization behavior from text to image modality.}
    \label{fig:continual_table_readout_example_images}
\end{figure*}

\looseness-1
Here, we introduce a concrete methodology to precisely study such issues. First, we design visual tasks where the image information relevant to the task can also be represented as text (e.g., LaTeX code).  This allows a direct comparison of the effect of training strategies in individual modalities and combinations. Second, to allow a clear comparison of different training strategies, we measure the {\em brittleness} of learning with \textbf{simple-to-hard (S2H) generalization}, where models are trained on {\simple} examples of a task and evaluated on {\hard} examples. 

\looseness-1
We create a set of synthetic tasks\footnote{Creating such tasks was more nontrivial than expected, for reasons described in \cref{sec:gen-issue}.} that involve algorithmic visual reasoning \citep{ghosal2024language,cherian2023deep,zhang2024if}: {\tableread} (reading out table entries in an order specified visually), \gridnav{} (finding valid paths through grid-like structures while avoiding obstacles), and \visualanalogy{} (identifying logical patterns across sets of abstract visual examples and applying analogical reasoning). Each task requires many reasoning steps while dynamically shifting attention over a sequence of small regions in the image. {\simple} and {\hard} examples differ in the length and complexity of the necessary reasoning steps. 

The {\simple} tasks are difficult for current frontier VLMs such as GPT-4o and Claude-3.5 Sonnet \citep{achiam2023gpt, anthropic2024claude35sonnet} (\cref{app:other_models}). Since we work with smaller open-parameter models, our methodology consists of using supervised training to precisely inject capability at a task in one modality and then study how variations in training affect the gap in {\simpletohard} between modalities. Since the tasks are difficult for frontier VLMs, we expect the takeaways from our study to be of broader interest.

\looseness-1

\noindent{{\bf Illustrative example of {\consecutivetableread}}:  
\looseness-1
Given a table of numbers and indices of two table cells $(i, j)$ and $(k, l)$, the model needs to output every table entry between these two cells in a row-major order. The input table can be provided as an image or as text (i.e., LaTeX code), allowing the kind of study sketched in \cref{fig:continual_table_readout_example_images}. In the {\simple} task, the length of the output sequence is $5$ to $10$, whereas in the {\hard} task, it can be as long as $30$. Therefore, {\simpletohard} here is a type of \textit{length generalization}, a well-studied concept in LLMs \citep{zhou2024LengthGeneralization}. SFT on $8 \times 10^4$ {\simple}-text examples yields $80\%$ accuracy on {\hard}-text examples. However, training on the \simple-image examples results in only $20\%$ accuracy on the {\hard}-image examples. The $60\%$p difference is a measure of the {\em modality gap} or {\em modality imbalance}.

\subsection{Paper Overview} 
\looseness-1
We study training strategies that incorporate various types of supervision: text-based, image-based, and combinations of the two (\cref{sec:sup_types}). We find that the most reliable way to alleviate the gap is to teach the model {\bf image reasoning via text conversion} --- explicitly extracting information from the image in text form before generating the solution using CoT. Specifically, we find: (i) for tasks where the model exhibits {\simpletohard} in the text modality, training on {\bf image reasoning via text conversion} greatly helps to mitigate the gap (\cref{sec:consecutive_read_exp}); (ii) for tasks where the {\simpletohard} failed in both modalities, applying the idea from (i) while also injecting reasoning capability on the {\hard} task in the text modality leads to {\simpletohard} in the image modality (\cref{sec:tr_ar_gn}). The findings in (ii) should be interpreted as suggesting that simple image-to-text conversion could be a promising intervention to reduce modality imbalance in future VLMs whose base LLM does exhibit {\simpletohard} in the text modality. 

\looseness-1
A surprising finding is that even though explicitly training on image-to-text conversion seems necessary for {\simpletohard}, the final trained model can generate the correct solution without explicitly extracting the image content as text: the image-to-text conversion skill gets internalized! (This also greatly reduces the inference-time cost.) Therefore, we try to understand the effectiveness of this key intervention at the level of training gradients. We find that gradients from {\simple}-image reasoning examples can help reduce loss on {\hard}-image inputs with the above intervention (\cref{sec:grad_align_main}); this gradient alignment merits further study. 

\looseness-1
On tasks where we need to inject reasoning capability on the \hard{} task, our findings about gradients inspired a more effective two-phase training (\cref{sec:alcmft}). The first phase teaches the model to do image reasoning via text conversion on a few {\simple} examples. We find that inclusion of this phase substantially improves gradient alignment in the earlier phases of training, when gradients have larger norms, which allows for more effective {\simpletohard} on the image modality. This finding is in accord  with previous empirical evidence that highlights the importance of visual-language alignment in VLM training~\citep{fan2024pre}. 
\begin{figure*}[!t]
    \centering
    \includegraphics[width=0.97\linewidth]{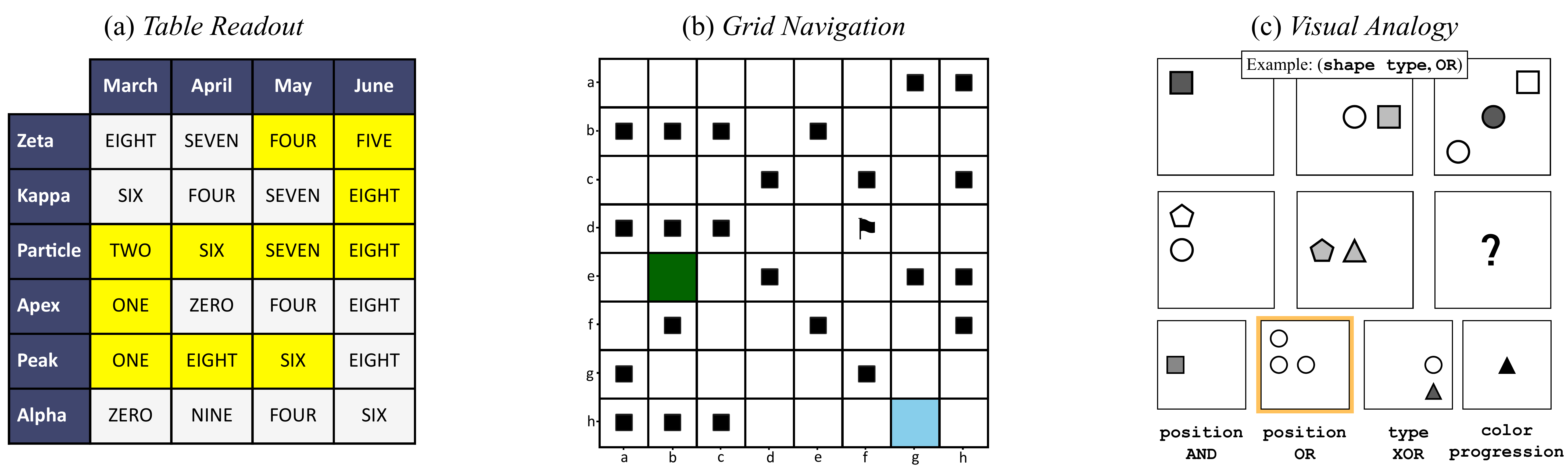}
    \vspace{-3mm}
    \captionof{figure}{ \textbf{Illustration of our synthetic tasks:} 
    \tableread{} involves reading numbers along a specified path in a table. \gridnav{} involves navigating a grid to collect objects while avoiding obstacles. \visualanalogy{} involves solving analogical reasoning queries using two in-context examples. \textbf{More details on {\visualanalogy}:} For the example of {\visualanalogy} above, we only include one in-context example for simplicity and provide annotations for clarity. In the first row, the first cell contains a rectangle, whereas the second and third cells contain a circle and a rectangle. Therefore, the in-context example is consistent with applying the {\sc OR} relation along the shape type domain. The model then needs to identify the correct option that corresponds to applying the {\sc OR} relation to the first two cells of the query along some (potentially different) domain. See \cref{app:synthetic} for non-annotated example images that are provided to the model.}
    \label{fig:example_images}
\end{figure*}%

\section{General Setup}

\subsection{Model}\label{sec:model}
In line with \citet{shi2024EAGLE} that show the benefit of combining multiple image encoders in VLMs, we trained \textit{Eagle-X2-Llama3-8B}, a variation of \textit{Eagle-X5} that uses Llama3-8B-Instruct \citep{dubey2024llama} as the LLM backbone and CLIP-448 \citep{radford2021learning} and ConvNeXt \citep{liu2022convnet} as visual encoders. Since the original paper found only minor benefit beyond the two encoders, we do not use all five visual encoders. See \cref{sec:training_details} for more details on the training. In \cref{app:qwen}, we replicate some of the experiments on \textit{Qwen2.5-VL-3B-Instruct} and \textit{7B-Instruct} \citep{bai2025qwen25vltechnicalreport} and observe consistent results.

\subsection{Tasks}
\label{sec:tasks_desc}
We briefly describe the tasks that we consider and the {\simple} and {\hard} setup of each task below (summarized in \cref{tab:avr_summary}; fully detailed in \cref{app:synthetic}). 

\begin{itemize}[leftmargin=*]
    \item \textbf{{\tableread}}: 
    \looseness-1
    The model sequentially reads numbers along a highlighted path in a table (given in either image or its LaTeX code). {\simple} examples consist of $1$–$4$ linear segments in spiral or sinusoidal path patterns with an average length of $12$ (\cref{fig:table-id}). {\hard} examples consist of $>4$ linear segments, featuring longer and arbitrary compositions of spiral or sinusoidal path patterns with an average length of $35$ (\cref{fig:table-ood}).

    \item \textbf{{{\consecutivetableread}}}:
    \looseness-1
    This is a variant of \textit{Table Readout}, modified to make the reasoning simpler\footnote{There is a fixed underlying rule for where the next cell should be. The model doesn't need to make a decision at each step.}. The model sequentially reads numbers in a \textit{row major order}. The number of cells to read in {\simple} and {\hard} examples is respectively $5$-$10$ and $25$-$30$. Only for this task, we additionally prepare a set of \textsc{medium} difficulty level, where the number of cells to read is $15$-$20$. Training on {\simple} examples and evaluating on \textsc{medium} examples can also measure {\simpletohard}.
    
    \item \textbf{{\gridnav}}: 
    \looseness-1
    The model navigates in a 2D grid (given in either image or its LaTeX code) from a designated start cell to an end cell while collecting all specified objects and avoiding obstacles. {\simple} examples contain $1$–$2$ objects and $1$ type of obstacle (\cref{fig:grid-id}). {\hard} examples involve $\geq2$ distinct objects and $\geq3$ types of obstacles (\cref{fig:grid-ood}). The task can be solved by depth-first search (DFS). Recent works \citep{kim2024language, wu2024vsp, wang2024picture} explored similar synthetic tasks in LLM and VLM evaluation.
    
    \item \textbf{{\visualanalogy}}: 
    \looseness-1
    The model reasons about attributes and relations between geometric figures in a puzzle (given in the image or text description). It analyzes two in-context examples and applies an analogous reasoning to choose $1$ from $4$ options to complete the query. {\simple} puzzles have examples and query vary along the \emph{same} attribute following a common relation (\cref{fig:ar-domain-transfer-heldout-id}). {\hard} puzzles have examples and query vary along \emph{different} attributes following a relation, and the combinations of attribute and relation \emph{held-out} from training (\cref{fig:ar-domain-transfer-heldout-ood}). This task is adapted from \citet{barrett2018measuring} and \citet{hill2019learning}.
    
    \item \textbf{{\em Pattern-Heldout Visual Analogy}}:
    \looseness-1
    This is a variant of \textit{Visual Analogy}, modified to make the reasoning simpler. See \cref{sec:compositional_VA} for more details. 
    
\end{itemize}

\subsection{Training Data}
Formally, we let $\reasontask: \inputset \to \reasonset$ denote a reasoning task, where $\inputset$ refers to a set of input data, further split into $\inputset_{\simple}$ and $\inputset_{\hard}$, and $\reasonset$ refers to a set of answers. Each input $\data \in \inputset$ can be presented in text format $\datatext$ or in image format $\dataimage$. 

For each pair of data $\data$ and solution $\reasontask(\data)$, we also create a chain-of-thought reasoning trace, which we denote by $\reasoningtrace (\data)$. We also define a prompt $P_{convert}$ that we optionally prepend at the start of chain-of-thought to signal explicit image-to-text conversion on image input\footnote{e.g., ``Convert the provided image to text''}. Hence, our training dataset is defined by input $\data$ (which can be given either as $\datatext$ or $\dataimage$), chain-of-thought $\reasoningtrace (\data)$, and the final answer $\reasontask(\data)$.

For each task $f$, we use the same Python script and a fixed template to generate all tuples ($\datatext$, $\dataimage$, $\reasoningtrace (\data)$, $\reasontask(\data)$).

\subsection{Types of Supervision}
\label{sec:sup_types}
Our controlled experiments study the effect of the following types of supervision on {\simple} examples during training:

\begin{enumerate}[(a), leftmargin=*]
    \item \smfttext{} supervision: given a text input $\datatext \in \inputset_{{\simple}}$, we train on the gold output containing a chain-of-thought trace $\reasoningtrace (\data)$ and the final answer $\reasontask(\data)$.
     
    \item \smftimage{} supervision: given an image input $\dataimage \in \inputset_{{\simple}}$, we train on the gold output containing a chain-of-thought trace $\reasoningtrace (\data)$ and the final answer $\reasontask(\data)$.

    \item \looseness-1\imageviatext{} supervision: given image input $\dataimage \in \inputset_{{\simple}}$, we train on the gold output containing the conversion prompt $P_{convert}$, converted text $\datatext$, a chain-of-thought trace $\reasoningtrace (\data)$, and the final answer $\reasontask(\data)$. 

    \item \imagetext{} supervision: we train on an equal mix of \smfttext{} and \smftimage{} supervisions.

    \item  \mixft{} supervision: we train on an equal mix of \smfttext{}, \smftimage{}, and \imageviatext{} supervisions.
\end{enumerate}

We train the model on one of the above supervision types with auto-regressive loss ($\evalloss$) that takes in the model's logits on an input example and returns the average loss on a selected set of tokens. For example, for \smftimage{} supervision, we will represent the input example as $\{\dataimage, \reasoningtrace (\data), \reasontask(\data)\}$, and compute the loss on $\{\reasoningtrace (\data), \reasontask(\data)\}$. During the evaluation, we test whether the model predicts $\reasontask(\data)$ correctly for a given input. 

In \cref{sec:tr_ar_gn}, we will adapt some of the above supervision strategies to also include {\hard} {\smfttext} supervision\footnote{Identical as {\smfttext} except we use a {\hard}-text example, i.e., $\datatext \in \inputset_{{\hard}}$. We note the subtle difference between a ``{\hard}-text example,'' which refers to the data $\datatext$, and ``{\hard} {\smfttext},'' which is a type of supervision with a prescribed (input, output) structure: $\{\datatext, \reasoningtrace (\data), \reasontask(\data)\}$. Similarly, a ``{\hard}-image example'' is distinct from {\hard} {\smftimage}.}. The adapted supervision strategies will have a $\textbf{+}$ sign appended to represent this additional component (e.g., {\cmft} adapted from {\mixft} supervision).
\section{Modality Imbalance in {\consecutivetableread}} \label{sec:consecutive_read_exp}

\looseness-1
We use {\consecutivetableread} introduced in \cref{sec:intro} and \cref{sec:tasks_desc} to illustrate the {\simpletohard} gap between different modalities and propose training strategies needed to address it. We compare different types of supervision by training on the prescribed {\simple} examples and measuring the improvements on the exact match accuracy\footnote{Correctness requires all generated numbers to be in the correct order; see \cref{sec:evaluation_details}.} on two different difficulty levels: (a) {\textsc{medium}}: reading $15$–$20$ consecutive numbers and (b) {\hard}:  $25$–$30$ numbers (more challenging). 

\looseness-1
To demonstrate the modality imbalance, we compare \smfttext{} and \smftimage{} supervision. \Cref{fig:exact_match_consecutiveread} shows that the {\simpletohard} gap between the two is substantial. For {\hard}, while {\smfttext} supervision achieves $80\%$ accuracy on {\hard}-text examples, {\smftimage} supervision achieves only $20\%$ on {\hard}-image examples.

\begin{figure}[t]
    \centering
    \includegraphics[width=0.97\linewidth]{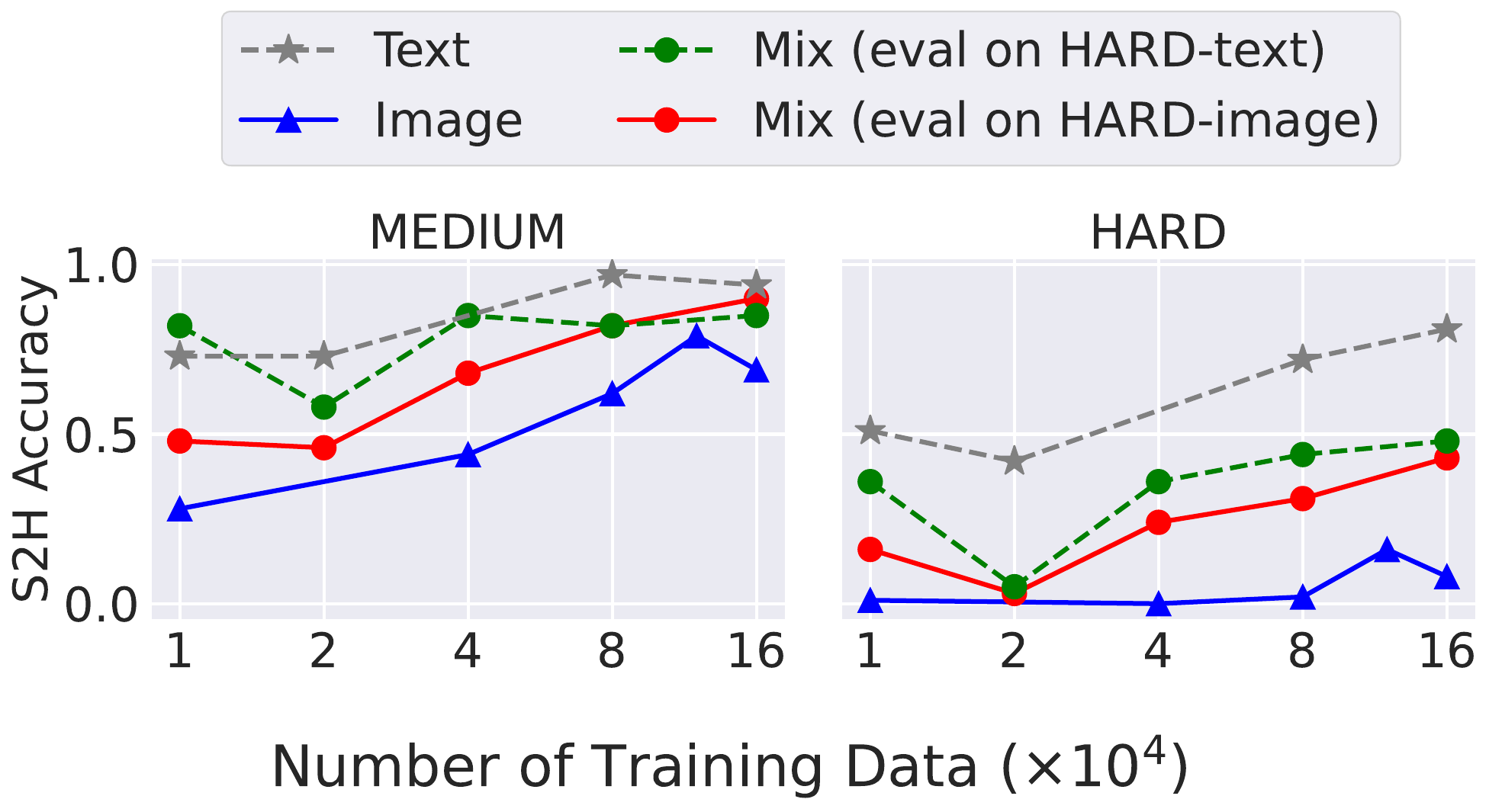}
    \caption{\looseness-1\textbf{S2H Generalization} of different supervisions for {\consecutivetableread} to {\textsc{medium}} (left) and {\hard} (right) examples. {\simpletohard} on text of {\smfttext} ($\star$) outperforms {\simpletohard} on image of {\smftimage} ($\blacktriangle$), highlighting modality imbalance. {\mixft} ($\bullet$) mitigates this imbalance.}
    \label{fig:exact_match_consecutiveread}
\end{figure}

\begin{figure}[t]
    \centering
    \includegraphics[width=0.97\linewidth]{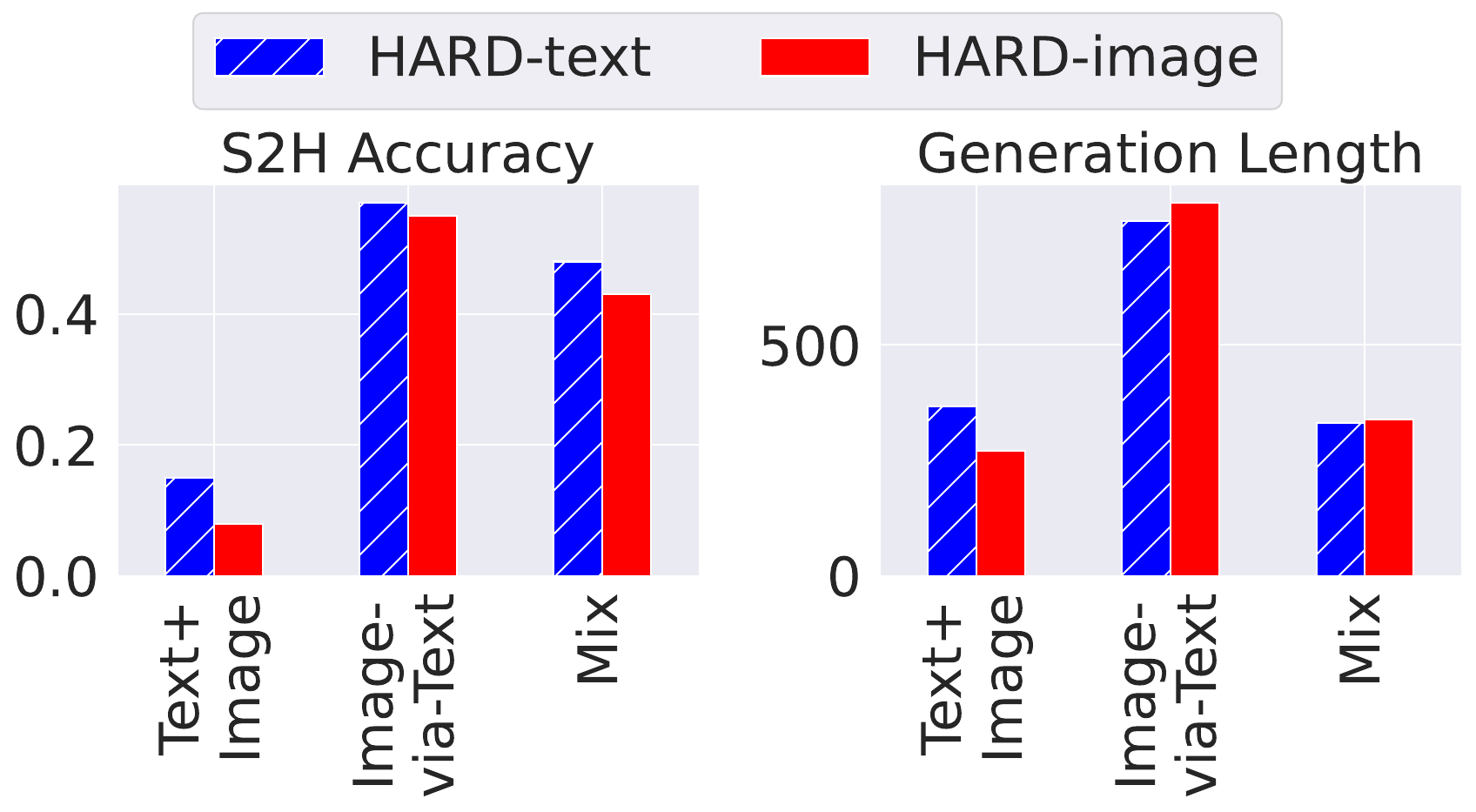}
    \caption{\textbf{Effect of \imageviatext{} on {\consecutivetableread}:} S2H Generalization (left) and Generation Length (right) for {\hard} task. Number of training data is $16 \times 10^4$. \imagetext{} underperforms \mixft{} and \imageviatext{} supervision. \imageviatext{} supervision improves performance slightly but at the cost of longer generation due to explicit image-to-text conversion at inference.}
    \label{fig:gen_length_consecutiveread}
\end{figure}

\looseness-1
In order to reduce the gap, we consider training strategies that can leverage strong {\simpletohard} of {\smfttext} supervision to help {\simpletohard} of {\smftimage} supervision. Two candidates are \imagetext{} supervision, which simply mixes in \smfttext{} and \smftimage{} supervision, and \imageviatext{} supervision, which trains the model to first convert the image input to its text format and then output the solution. \imagetext{} supervision induces the model to implicitly make the connection that the image and text formats are equivalent, while \imageviatext{} supervision makes this connection explicit.  We compare the two training strategies in \cref{fig:gen_length_consecutiveread} and show that \imageviatext{} supervision shows much better performance on {\hard}-images. 

\looseness-1
However, \imageviatext{} supervision has a key drawback: trained models have significantly higher inference costs, since the conversion of image to text before generating the solution leads to $3\times$ longer outputs, which limits the real-world practicality. To address this, we propose \textbf{{\mixft}} supervision, which combines \imageviatext{} and \imagetext{} supervision. This teaches the model to align the modalities, while also teaching it to not always rely on the image-to-text conversion. 

\looseness-1
\begin{tcolorbox}[observation]
{\mixft} can mitigate the modality imbalance by improving {\simpletohard} on images, while maintaining inference cost.
\end{tcolorbox}

{\mixft} supervision retains most of the {\simpletohard} performance of \imageviatext{} supervision while reducing generation length by directly solving reasoning tasks from images (\cref{fig:gen_length_consecutiveread}). In \cref{fig:exact_match_consecutiveread} (left), we show that it can almost completely match the {\simpletohard} performance of {\smfttext} supervision for {\textsc{medium}}. On {\hard} level, even though it does not fully close the gap between text and image input (\cref{fig:exact_match_consecutiveread}, right), the gap can be further reduced with a short text-only warm-up training. We discuss this further in \cref{sec:ablations}.

\paragraph{Consistent results across tasks:}
\looseness-1
In \cref{sec:compositional_VA}, we show similar results on {\em Pattern-Heldout Visual Analogy}.
\section{ Full Study: {\tableread},  {\gridnav}, {\visualanalogy}}
\label{sec:tr_ar_gn}

\looseness-1
% We now consider three \textit{{\nontextgeneralizable}} synthetic tasks, where our model struggles to generalize to {\hard} instances in either modality after training on the corresponding {\simple} examples: {\tableread}, {\gridnav}, and {\visualanalogy} (\cref{fig:example_images}). 

We now consider our three main tasks: {\tableread}, {\gridnav}, and {\visualanalogy} (\cref{fig:example_images}). These tasks require the model to generalize to {\hard} examples by composing reasoning patterns learned from {\simple} training examples, which has been known to be difficult for LLMs \cite{yu2023skill, zhao2024can, wu2024conceptmix, huang2023t2i, dziri2024faith}.

These tasks are considered \textit{{\nontextgeneralizable}} because the model struggles to generalize to {\hard} instances after being trained on {\simple} examples. In any of the three settings, training with {\smfttext}, {\smftimage}, and {\mixft} supervision (which only include {\simple} examples) cannot achieve more than $25\%$ {\simpletohard} on either text or image.

The failure to S2H-generalize in either input modality highlights the insufficient general reasoning capacity of existing models on these tasks. We then adapt {\mixft} supervision to include {\hard} {\smfttext} in training and measure whether the improved performance on {\hard}-text can result in better S2H generalization in the image modality.

\begin{figure*}[!ht]
    \centering
    \includegraphics[width=0.90\linewidth]{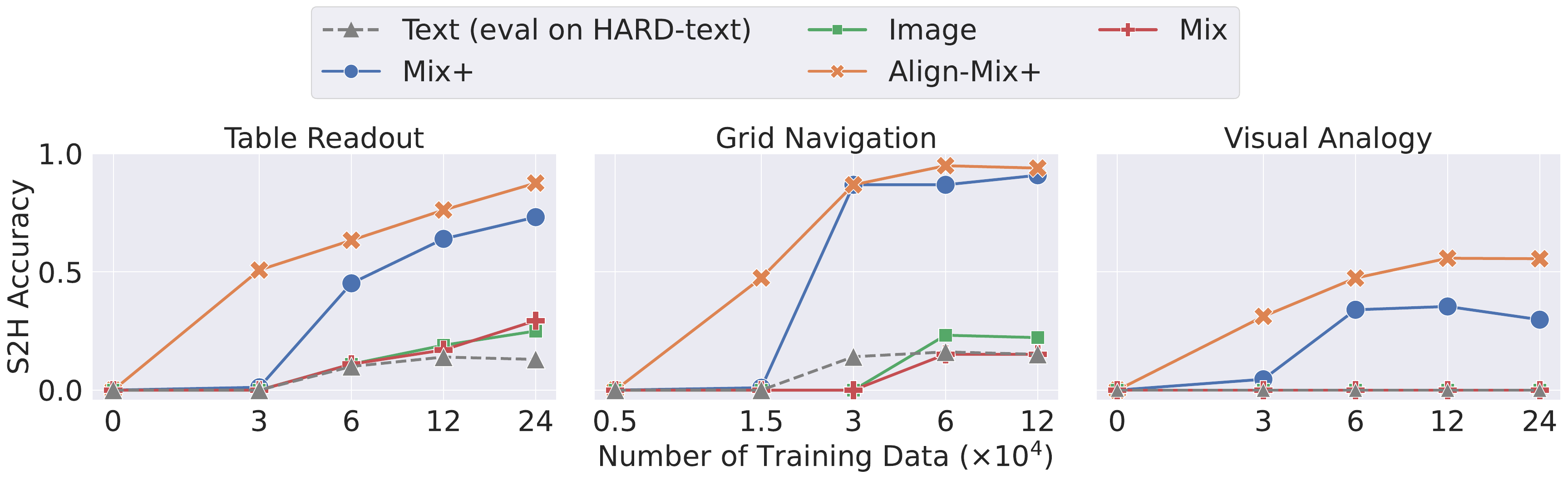}
    \caption{\textbf{Results on {\nontextgeneralizable} tasks:} We report the S2H generalization on image on {\tableread} (left), {\gridnav} (middle), and {\visualanalogy} (right). S2H generalization on text from {\smfttext} supervision serves as a reference (in gray dashed line). {\smfttext}, {\smftimage}, and {\mixft} supervisions fail to generalize, highlighting the gap between {\simple} and {\hard} examples. {\cmft} improves performance, while {\alcmft} further enhances generalization with an initial alignment phase. }
    \label{fig:main-results}
\end{figure*}

% \subsection{Improved text performance with {\hard} {\smfttext} supervision improves S2H generalization on image}
\subsection{Improved performance on {\hard}-text can transfer to S2H generalization on image}

\looseness-1
{\cmft} supervision, adapted from {\mixft} from \cref{sec:consecutive_read_exp}, trains the model with an equal mix of {\hard} {\smfttext} supervision and {\simple}{ \mixft} supervision.  

\looseness-1

\begin{tcolorbox}[observation]
{\cmft} supervision shows significantly better image {\simpletohard}, demonstrating an effective transfer of reasoning capability from text to image.
\end{tcolorbox}

With only $3 \times 10^4$ data, {\cmft} quickly improves the model's accuracy on {\hard}-text examples to $\ge95\%$. At the same time, {\cmft} supervision leads to a significant improvement on image {\simpletohard} --- the model can achieve $64\%$, $92\%$ and $35\%$ S2H accuracy on {\hard}-images (respectively, {\tableread}, {\gridnav}, and {\visualanalogy}) after being trained on $12 \times 10^4$ data (\cref{fig:main-results}). We conclude that {\cmft} supervision can effectively transfer the injected reasoning on {\hard}-text to S2H generalization on images.

\subsection{Dual capability of {\cmft}}

\looseness-1
Motivated by the observed benefit of {\imageviatext} supervision from \cref{sec:consecutive_read_exp}, we also measure the image {\simpletohard} of {\cimageviatext} supervision (an equal mix of {\hard} \smfttext{} supervision and {\simple} {\imageviatext} supervision). On {\tableread} and {\visualanalogy}, we observe that {\cimageviatext} supervision outperforms {\cmft} supervision in S2H performance on {\hard}-image by a substantial ($20$-$30\%$p) gap (\cref{fig:compare_itota_cmft}).

\begin{figure}[htbp]
    \centering
    \includegraphics[width=0.97\linewidth]{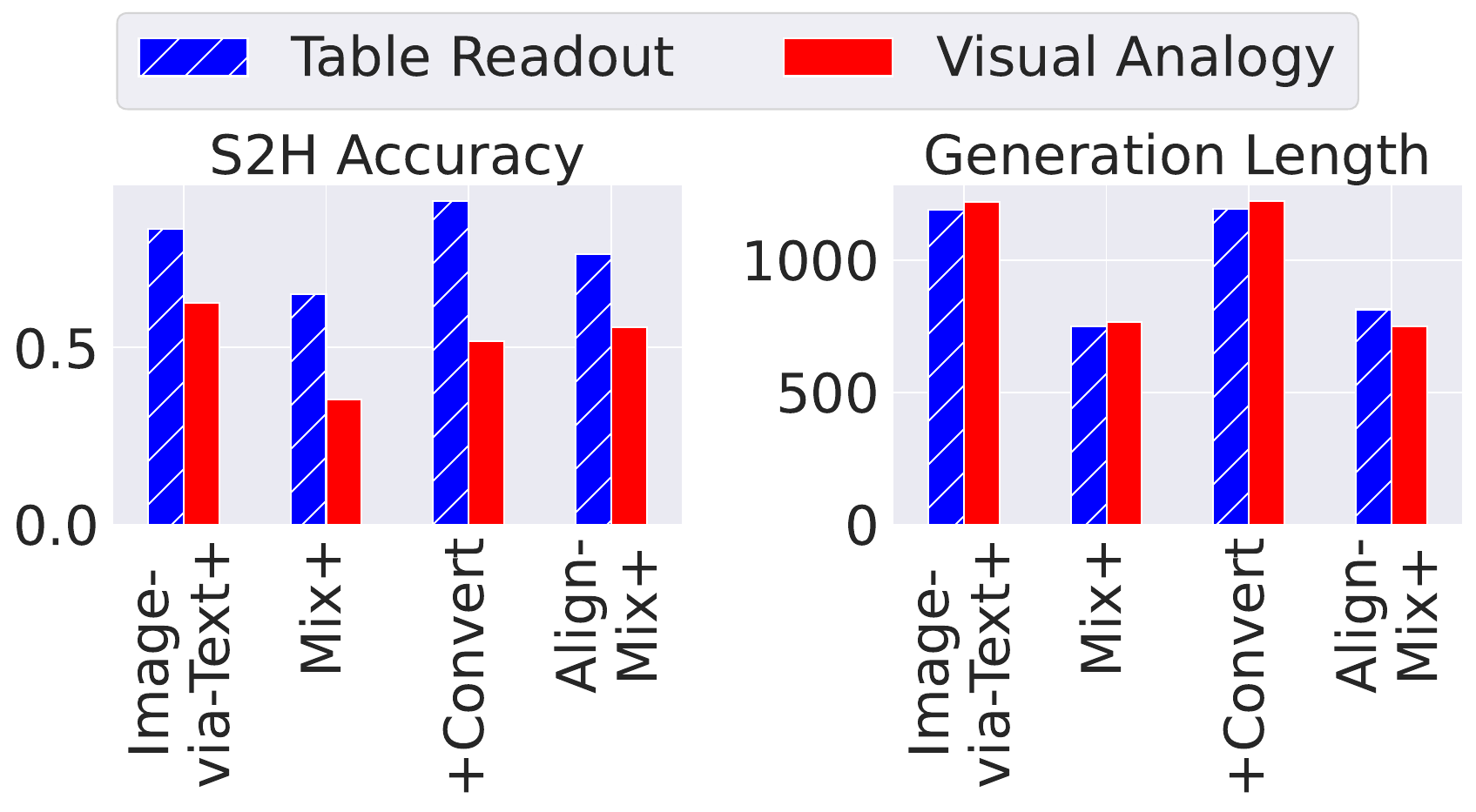}
    \caption{\textbf{{\cimageviatext} on {\tableread} and {\visualanalogy}:} S2H Generalization on image (left) and Generation Length (right) with $12 \times 10^{4}$ training examples. {\cimageviatext} achieves good performance but with higher inference cost. {\cmft} matches the performance of \cimageviatext{} by appending ``\texttt{Convert}'' to the prompt ($\mathsb{\text{+}Convert}$) or by adding an alignment phase ({\alcmft}).}
    \label{fig:compare_itota_cmft}
\end{figure}

\looseness-1
To close this gap, we prompt {\cmft} models with an additional inference time token, ``\texttt{Convert}'', which appears at the start of the {\imageviatext} responses (\cref{sec:sup_types}). We observe that the models respond with an accurate text conversion before generating the reasoning tokens. 

\begin{tcolorbox}[observation]
{\cmft} models exhibit a \textbf{dual capability} in reasoning with or without image-to-text conversion.
\end{tcolorbox}
This is in line with the findings in \citet{su2024dualformer} of the dual learning capability of LLMs in short and long reasoning. Note that when explicitly prompting {\cmft} models to perform image reasoning via text conversion, this still incurs a similar cost in generation length as {\cimageviatext} (\cref{fig:compare_itota_cmft}). We discuss more in \cref{app:ablations_text_conversion}.

\subsection{Benefits of two-phase training }
\label{sec:alcmft}

\looseness-1
Given we previously observe that \imageviatext{} supervision helps with S2H generalization, we add an initial phase that trains the model with \smfttext{} and \imageviatext{} supervision on \simple{} examples. The goal is to precondition the model (via {\simple} {\imageviatext} supervision) to align text and image reasoning on {\simple} examples. Intuitively, the preconditioning must be useful to generalize this knowledge on {\hard} examples later when trained with {\cmft} supervision. We call this two-phase approach {\alcmft}\footnote{For the main experiments, we use $1 \times 10^4$ training examples in the first phase. See \cref{sec:ablate_alcmft} for ablations on the number and composition of data used in the alignment phase.}.

\looseness-1
{\alcmft} significantly boosts {\simpletohard} on image to an accuracy of $76\%$, $96\%$, and $56\%$ on {\hard}-images (respectively {\tableread}, {\gridnav}, and {\visualanalogy}) after training on $12 \times 10^4$ data (\cref{fig:main-results}). {\alcmft} also maintains inference cost (\cref{fig:compare_itota_cmft}). 

\begin{tcolorbox}[observation]
{\alcmft} further improves image {\simpletohard}, while maintaining inference cost.
\end{tcolorbox}
\section{A Study on Loss Dynamics and Gradient Alignment for {\simpletohard}}
\label{sec:grad_align_main}

\looseness-1
Our findings show that {\simpletohard} can be transferred across modalities by simply mixing different types of supervision. This happens without any explicit matching of representations, which motivates us to explore training gradients to obtain insights into how each strategy contributes to {\simpletohard}. Here, we analyze the evaluation loss behavior on {\hard} {\smftimage} and {\hard} {\imageviatext}\footnote{Identical as {\smftimage} and {\imageviatext} except image input $\dataimage \in \inputset_{{\hard}}$.} examples during training. Similar gradient studies have been proposed for measuring influence \citep{koh2017understanding} of training data points on evaluation tasks~\citep{park2023trak,xia2024less,engstrom2024dsdm}.

\subsection{A study on {\consecutivetableread}}
\looseness-1
In \cref{sec:consecutive_read_exp}, we showed that \mixft{} outperforms {\imagetext} supervision in {\simpletohard} on images. The key factor driving this improvement was the inclusion of \imageviatext{} supervision. Here, we show that \mixft{} supervision reduces evaluation loss on {\hard} {\smftimage} examples (therefore improving evaluation accuracy) through a better gradient signal. To do so, we measure the alignment between gradients on {\simple} and {\hard} {\smftimage} examples. 

\looseness-1
Let $\evalitoa(\data)$ denote the \textit{loss on solution given image}, i.e.,
\begin{equation}
    \evalitoa(\data) := \evalloss(\model(\{\dataimage,\datalabel\}), \datalabel))
\end{equation}
where $\datalabel$ contains both $\reasoningtrace(\data)$ and the answer $\reasontask(\data)$. We also denote the \textit{loss on solution given {\hard} image} as
\begin{equation}
    \evalitoahard := \mathbb{E}_{\data \in \inputset_{\hard}} \evalitoa(\data)
    \label{eq:hard-loss}
\end{equation}
If $\mathbf{g}_{\simple}$ and $\mathbf{g}_{\hard}$ denote average gradients on $\inputset_{\simple}$ and $\inputset_{\hard}$ (i.e. $\mathbb{E}_{\data \in \inputset_{\simple}} \nabla \evalitoa(\data)$ and $\mathbb{E}_{\data \in \inputset_{\hard}} \nabla \evalitoa(\data)$ respectively)\footnote{Following \citet{park2023trak}, we apply a random projection on gradients to $4096$ dimension for an efficient storage.}, then we define the \textbf{gradient alignment score} as \footnote{In \cref{sec:adam_update}, we give two alternative measures of gradient alignment. Our takeaways remain the same.}:
\begin{equation}
    \langle \mathbf{g}_{\simple}, \mathbf{g}_{\hard} \rangle / \langle \mathbf{g}_{\hard}, \mathbf{g}_{\hard} \rangle \label{eq:grad_align}
\end{equation}

\looseness-1
Intuitively, the gradient alignment score measures how much the evaluation loss (on {\hard} {\smftimage}) can be reduced by taking gradient updates from the training data ({\simple} {\smftimage}), relative to training on evaluation data directly (see \cref{lem:grad_align_SGD} for a formal statement). In \cref{fig:Utility_ID_OOD_ctr}, we plot this score against the gradient norms on the training data. A stronger gradient alignment at larger values of gradient norm is preferred because the evaluation loss can be reduced more when the training gradients are larger.

\begin{tcolorbox}[observation]
{\mixft} achieves a high gradient alignment score, especially when gradient norms are large. This improved alignment leads to a significant initial drop in the evaluation loss (\textit{loss on solution given {\hard}-image}), which then continues to improve throughout training.
\end{tcolorbox}

\begin{figure}[t]
    \centering
    \includegraphics[width=\linewidth]{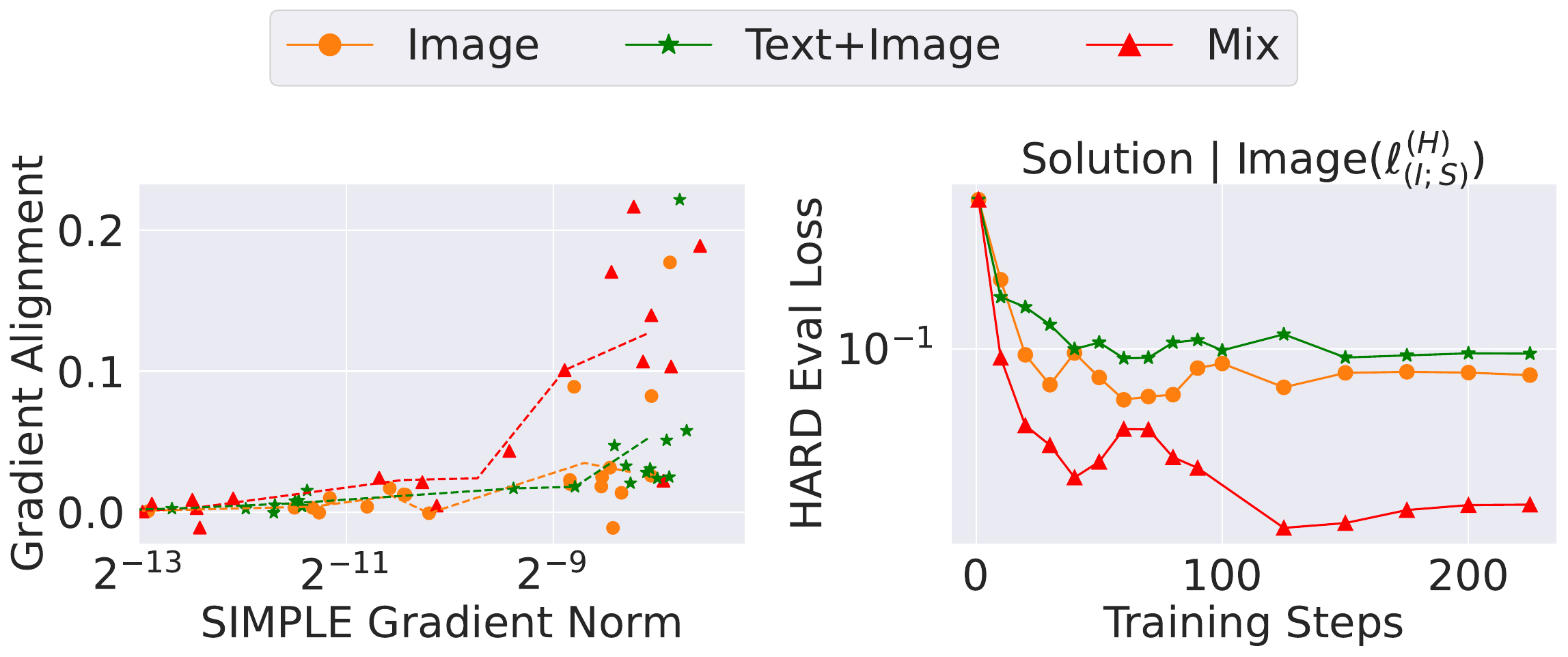}
    \caption{\textbf{Analysis of gradients on {\consecutivetableread}:} (Left) Average Gradient Norm on \simple{} {\smftimage} examples $(\mathbb{E}_{\data \in \inputset_{\simple}} \norm{\nabla \evalitoa (\data)})$ vs. Gradient Alignment Score (\cref{eq:grad_align}) for different training checkpoints; (Right) Average \textit{Loss on solution given {\hard} image} ($\evalitoahard$) during training. Larger gradients for {\mixft} have higher alignment score compared to {\imagetext} and {\smftimage}, showing the importance of {\imageviatext} supervision for generalization.}
    \label{fig:Utility_ID_OOD_ctr}
\end{figure}

\subsection{A study on {\tableread}}
\begin{figure*}[!t]
    \centering
    \includegraphics[width=0.90\linewidth]{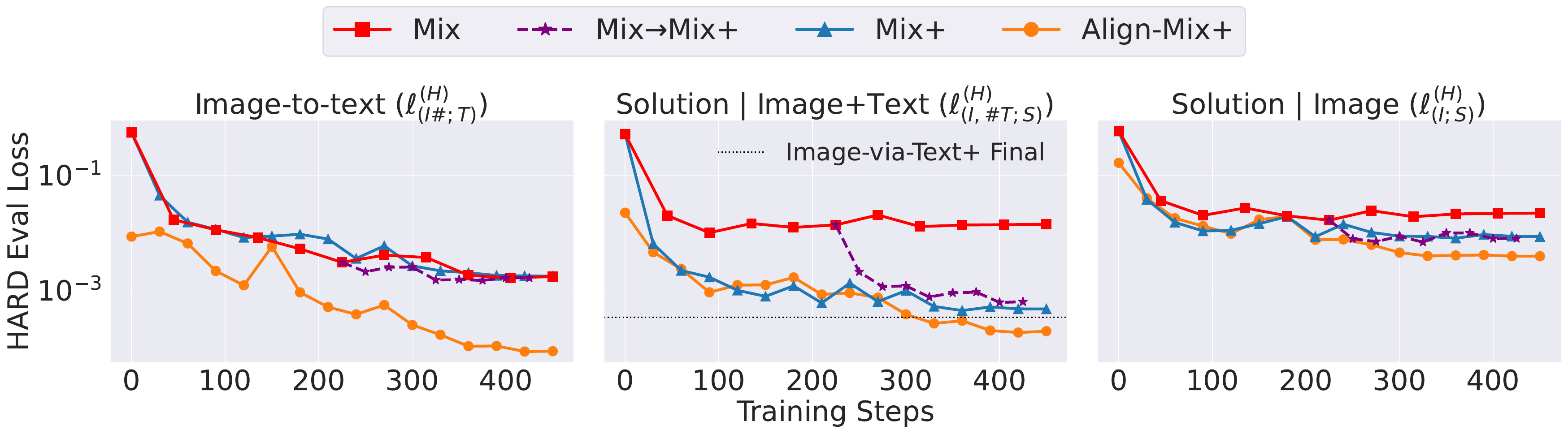}
    \caption{\looseness-1
    \textbf{Analysis of evaluation losses on {\hard} examples on {\tableread}:} (Left) \textit{{\hard} image-to-text conversion loss} ($\evalitothard$ (Eq.\ref{eq:itot})); (Middle) \textit{loss on solution given {\hard} image and text} ($\evalittoahard$ (Eq.\ref{eq:ittoa})); (Right) \textit{loss on solution given {\hard} image} ($\evalitoahard$ (Eq.\ref{eq:hard-loss})). {\mixft} matches {\cmft} in $\evalitothard$, showing that training on {\simple} {\imageviatext} examples is sufficient for {\hard} image-to-text conversion. {\mixft} performs worse in $\evalittoahard$, showing the need for {\hard} {\smfttext} examples for generalization. Taking an intermediate checkpoint of {\mixft} and completing the training with {\cmft} ({\mixft} $\rightarrow$ {\cmft}) leads to evaluation loss values comparable to {\cmft}, suggesting that {\hard} {\smfttext} examples can be introduced later. {\alcmft} starts with smaller $\evalitothard$ and $\evalittoahard$ losses, which helps the model achieve lower $\evalittoahard$ loss than even {\cimageviatext}, that reflects in lower $\evalitoahard$ loss.}
    \label{fig:loss_behavior}
\end{figure*}

In \cref{sec:tr_ar_gn}, we showed that {\cmft} improves {\simpletohard} over {\mixft}, while {\alcmft} can further improve over {\cmft} with an additional alignment training. Here, we study how each included component helps {\simpletohard} across the training strategies.

\paragraph{Insights from the evaluation loss dynamics:} 
We use the following additional notations to report the average loss on specific tokens on a {\hard} {\imageviatext} example and understand which components help the model learn to reason on {\hard}-images via text conversion, and how it translates to a direct solution on {\hard}-image examples.

\begin{itemize}[leftmargin=*]
    \item \textit{{\hard} image-to-text conversion}: Average loss on converted text tokens given the image and the conversion prompt \footnote{$\#$ indicates the additional conversion prompt $P_{convert}$}):
    \begin{equation}
        \evalitothard := \mathbb{E}_{\data \in \inputset_{\hard}} \evalloss(\model(\{\dataimage, P_{convert}, \datatext\}), \datatext). \label{eq:itot} 
    \end{equation}
    
    \item \textit{Solution given {\hard} image and text}: Average loss on solution tokens given the image, the conversion prompt, and the converted text: 
    \begin{equation}
        \evalittoahard := \mathbb{E}_{\data \in \inputset_{\hard}} \evalloss(\model(\{\dataimage, P_{convert}, \datatext, \datalabel\}), \datalabel), 
        \label{eq:ittoa} 
    \end{equation}
    where $\datalabel$ contains both $\reasoningtrace(\data)$ and the answer $\reasontask(\data)$.
\end{itemize}

In \cref{fig:loss_behavior}, we report the above losses for {\mixft}, {\cmft}, and {\alcmft}. Since the model does not see {\hard}-image examples during training, these losses (along with $\evalitoahard$) evaluate the {\simpletohard} on image. We observe:

\begin{enumerate}[leftmargin=*]
    
    \item \textit{{\hard} image-to-text conversion loss} (\cref{eq:itot}) of {\mixft} matches {\cmft}, showing that training on {\simple} {\imageviatext} examples suffices to generalize the conversion subtask to {\hard}-images. 
        
    \item There is a significant gap in the \textit{loss on solution given {\hard} image and text} (\cref{eq:ittoa}) between {\mixft} and {\cmft}. This implies that including {\hard} {\smfttext} is necessary to fully generalize reasoning to {\hard}-images.
    
    As an ablation, we took an intermediate {\mixft} checkpoint and completed the training with {\cmft} supervision\footnote{We preserved the optimizer states and learning rate schedule.}. This transition resulted in negligible changes to \textit{{\hard} image-to-text conversion loss} (\cref{eq:itot}), while \textit{loss on solution given {\hard} image and text} (\cref{eq:ittoa}) and \textit{loss on solution given {\hard}-image} (\cref{eq:hard-loss}) decreased significantly, approaching the values for {\cmft}.
    
    \item Losses on {\hard} {\imageviatext} examples start significantly lower for {\alcmft} after the alignment phase. This shows that training on {\simple} {\imageviatext} examples can return a favorable starting point, even if they aren't sufficient for generalization. It then achieves a better \textit{loss on solution given {\hard} image and text}  (\cref{eq:ittoa}) in the end, which also translates to an improved \textit{loss on solution given {\hard} image} (\cref{eq:hard-loss}).
\end{enumerate}

\begin{figure}[!ht]
    \centering
    \includegraphics[width=0.72\linewidth]{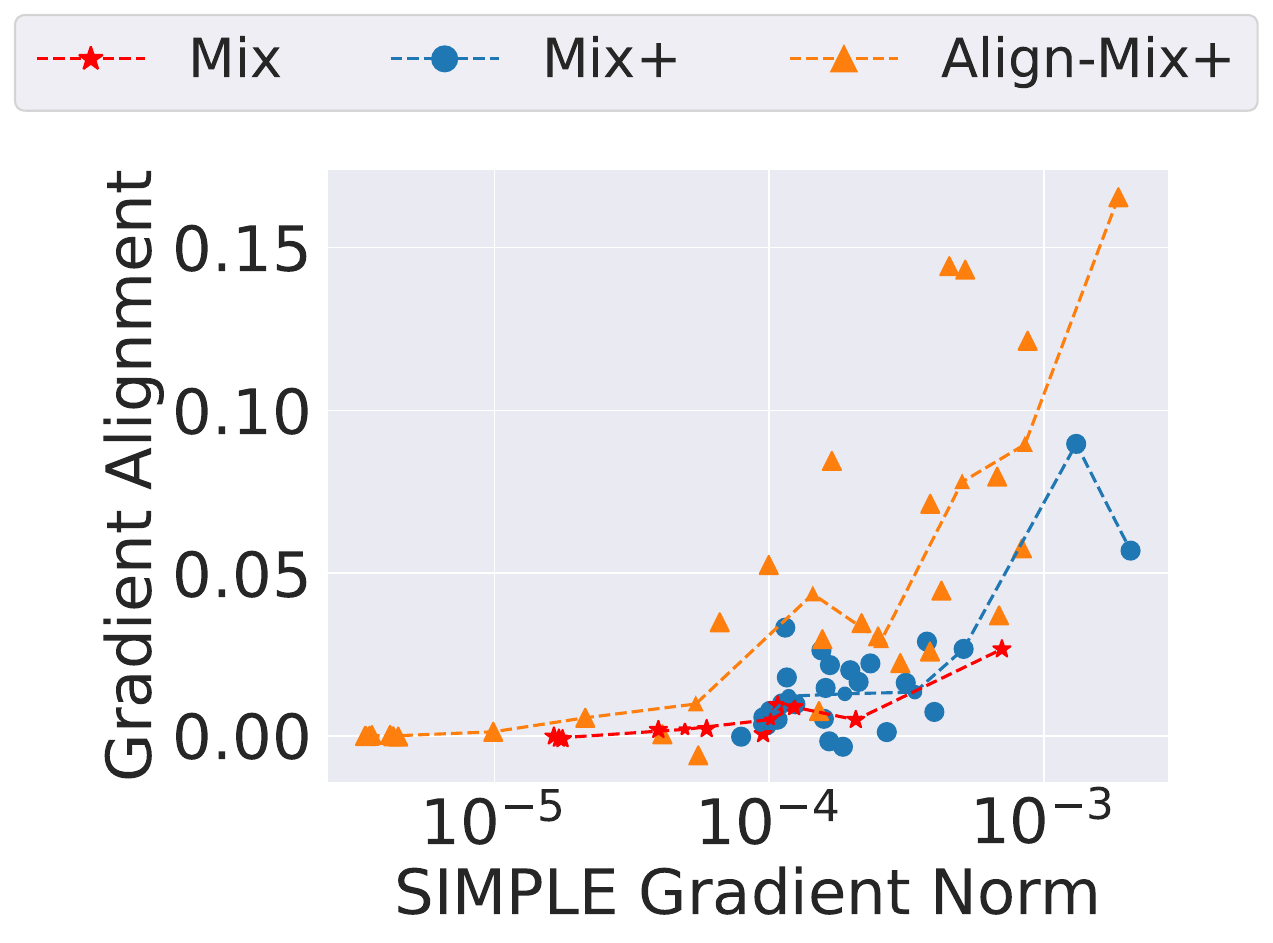}
    \captionof{figure}{\textbf{Analysis of gradients on {\tableread}:} Average Gradient Norm on \simple{} {\smftimage} examples $(\mathbb{E}_{\data \in \inputset_{\simple}} \norm{\nabla \evalitoa (\data)})$ vs. Gradient Alignment Score for different training checkpoints. Larger gradients for {\alcmft} have higher gradient alignment scores. {\cmft} has better gradient alignment scores than {\mixft}.}
    \label{fig:utility_gradient_ItoA_ID_OOD_contour}
\end{figure}

\paragraph{Insights from the gradient alignment score:} We can further quantify the differences in training strategies with the gradient alignment score (\cref{eq:grad_align}) between {\simple} and {\hard} {\smftimage} examples (\cref{fig:utility_gradient_ItoA_ID_OOD_contour}). Intuitively, a higher gradient alignment at each step should accumulate to a better generalization on {\hard}-images. We observe: 
\begin{enumerate}[leftmargin=*]
    \item {\mixft} exhibits lower gradient alignment compared to {\cmft}. Training solely on {\simple} examples fails to provide gradients aligned to {\hard} {\smftimage}. Including {\hard} {\smfttext} examples significantly improves gradient alignment.
    
    \item  On the other hand, {\alcmft} has higher gradient alignment than {\cmft} in earlier training steps, when training gradient norms are large. We give a detailed analysis in \cref{fig:utility_gradient_ItoA_ID_OOD} in \cref{sec:grad_align}. 

\end{enumerate}

\section{Further Ablations}
\label{sec:ablations}

\looseness-1
We perform several ablation studies to identify critical training components that underlie our findings. We push all the details and discussions to the appendix.

\looseness-1
\paragraph{Task interactions in multi-task training:} We compare {\mixft}, {\cmft} and {\alcmft} with an equal mix of $3$ tasks in \cref{sec:tr_ar_gn}. We observe that multi-task training significantly boosts performance on {\tableread} and {\gridnav} but hurts on {\visualanalogy}, which shows the effect of task interactions in our strategies. See \cref{app:ablations_multitask}.

\looseness-1
\paragraph{Transferring reasoning from image to text:} We also experiment with including {\hard} {\smftimage} supervision in training and evaluating on {\hard}-text input, which gives much stronger results (\cref{tab:symbiote} in \cref{sec:symbiote}), 

\looseness-1
\paragraph{Text warm-up pretraining:} We add a text warm-up pretraining {\tw{}} phase before the training of VLM to simulate the effect of a stronger LLM backbone. This pretraining phase completely solves the modality imbalance or further boosts performance of {\alcmft}. See \cref{app:ablations_continual_pretraining}.

\looseness-1
\paragraph{Importance of chain-of-thought:} Completely removing or progressively internalizing CoT \citep{deng2024explicit} fails to achieve image {\simpletohard}, suggesting that CoT is crucial in our strategies. See \cref{app:ablations_cot}.
\section{Discussion, Limitations and Future work} 
\label{sec:discussion}
\looseness-1
We explore the modality imbalance in VLMs by measuring \simpletohard{}. We show that on tasks where VLMs can reliably show generalization on text input after fine-tuning, {\mixft} supervision can induce a similar level of generalization on image input. We then propose $3$ algorithmic tasks, where models trained on {\simple} examples fail to generalize to {\hard} examples in either modality. Mixing {\hard} {\smfttext} examples in training can help the model generalize on {\hard}-image input, revealing \simpletohard{} transfer capabilities of these models. 

\looseness-1
\textbf{Related Works:} Current VLM benchmarks are often solvable without the visual input. To remove such bias, we designed controllable tasks and provided a framework (\simpletohard{}) to quantify and mitigate modality imbalance. While \simpletohard{} has been extensively studied for LLMs, similar investigations remain scarce for VLMs.

Prior strategies to address modality imbalance and cross-modal transfer often rely on matching representations or optimization techniques. However, through gradient alignment studies, we demonstrate that auto-regressive training effectively aligns reasoning across modalities.

For a more detailed discussion, see \cref{sec:related_works}. 

\looseness-1
\textbf{Utility to real-world benchmarks:}
Extending our findings to real-world scenarios is also left for future work. It will require real-world scenarios with precise gradation of \simple{} and \hard{} examples with respect to underlying abstract concepts. Our work suggests that the brittleness of VLMs could be mitigated by training them to create very detailed descriptions of the scene (and this capability could be internalized for faster inference). 

\looseness-1
We note that training even on our synthetically created datasets seems useful for improving the performance of VLMs in real-world settings. Specifically, including our synthetic datasets during pretraining of VLMs yielded significant improvements across different benchmarks (\cref{tab:results_realworld} in \cref{sec:utility}). For example, including {\simple} and {\hard} \smftimage{} supervision examples from all synthetic datasets can improve performance on MMMU \citep{yue2024mmmu} by at least $3\%$p. Similarly, on a chart dataset \citep{wang2024charxiv}, including our synthetic datasets can improve performance by $5.1\%$p on descriptive questions. Therefore, our synthetic datasets involve useful skills that can also help improve VLMs on real-world benchmarks. 

\looseness-1
\textbf{Limitations and possible future directions:} 
We believe {\mixft} or {\cmft} may not be the optimal approach to improve image generalization on tasks where the model exhibits {\simpletohard} in the text modality. Curriculum-based strategies \cite{xie2024doremi, mindermann2022prioritized} that dynamically adjust the data mixture could yield better results. However, our goal is to emphasize the {\hard} generalization gap between text and image inputs, which can be bridged by transferring learning from the dominant modality (text) to the weaker one (image). Therefore, we focus on the effectiveness of our training strategies in transferring knowledge learned on text input to image input.

\looseness-1
In the interest of crispness, we restricted the scope of our study with a small set of prompts and a limited (and synthetic) image distribution. But doing so allowed a clearer and quantitative look at modality imbalance and how it can be bridged.

\looseness-1
Our results highlight that chain-of-thought (CoT) reasoning can play an important role. However, even minor modifications to CoT significantly affect the transferred {\simpletohard} results on image inputs, and mitigating this brittleness through robust training strategies beyond {\cmft} is crucial. Future work could focus on mechanistic insights into our trained models to design more generalizable strategies targeting specific model components.

\section*{Acknowledgements}
SP, AP, CY, and SA acknowledge fundings from NSF, PLI, DARPA, ONR, and OpenAI. CY is additionally supported by the Francis Robbins Upton Fellowship in engineering. We thank Xingyu Zhu, Bingbin Liu, Nikunj Saunshi, Sadhika Malladi, Samy Jelassi, Misha Khodak, Zirui Wang, Mengzhou Xia, Yihe Dong, Haoyu Zhao, Danqi Chen, Tri Dao, and Benjamin Eysenbach for discussions, suggestions, and proof-reading at various stages of the paper.
\section*{Impact Statement}
This paper presents work whose goal is to advance the field of Machine Learning. It primarily is a basic scientific exploration of the capabilities of Vision Language Models. It may lead to the development of better VLMs, but we do not anticipate any negative societal impact.

\clearpage

\bibliography{references}

\begin{thebibliography}{95}
\providecommand{\natexlab}[1]{#1}
\providecommand{\url}[1]{\texttt{#1}}
\expandafter\ifx\csname urlstyle\endcsname\relax
  \providecommand{\doi}[1]{doi: #1}\else
  \providecommand{\doi}{doi: \begingroup \urlstyle{rm}\Url}\fi

\bibitem[Abbe et~al.(2024)Abbe, Bengio, Lotfi, and Rizk]{abbe2024generalization}
Abbe, E., Bengio, S., Lotfi, A., and Rizk, K.
\newblock Generalization on the unseen, logic reasoning and degree curriculum.
\newblock \emph{Journal of Machine Learning Research}, 25\penalty0 (331):\penalty0 1--58, 2024.

\bibitem[Achiam et~al.(2023)Achiam, Adler, Agarwal, Ahmad, Akkaya, Aleman, Almeida, Altenschmidt, Altman, Anadkat, et~al.]{achiam2023gpt}
Achiam, J., Adler, S., Agarwal, S., Ahmad, L., Akkaya, I., Aleman, F.~L., Almeida, D., Altenschmidt, J., Altman, S., Anadkat, S., et~al.
\newblock Gpt-4 technical report.
\newblock \emph{arXiv preprint arXiv:2303.08774}, 2023.

\bibitem[Agrawal et~al.(2016)Agrawal, Batra, and Parikh]{agrawal2016analyzing}
Agrawal, A., Batra, D., and Parikh, D.
\newblock Analyzing the behavior of visual question answering models.
\newblock In Su, J., Duh, K., and Carreras, X. (eds.), \emph{Proceedings of the 2016 Conference on Empirical Methods in Natural Language Processing}, pp.\  1955--1960, Austin, Texas, November 2016. Association for Computational Linguistics.
\newblock \doi{10.18653/v1/D16-1203}.
\newblock URL \url{https://aclanthology.org/D16-1203/}.

\bibitem[Agrawal et~al.(2024)Agrawal, Antoniak, Hanna, Bout, Chaplot, Chudnovsky, Costa, De~Monicault, Garg, Gervet, et~al.]{agrawal2024pixtral}
Agrawal, P., Antoniak, S., Hanna, E.~B., Bout, B., Chaplot, D., Chudnovsky, J., Costa, D., De~Monicault, B., Garg, S., Gervet, T., et~al.
\newblock Pixtral 12b.
\newblock \emph{arXiv preprint arXiv:2410.07073}, 2024.

\bibitem[Anil et~al.(2022)Anil, Wu, Andreassen, Lewkowycz, Misra, Ramasesh, Slone, Gur-Ari, Dyer, and Neyshabur]{anil2022exploring}
Anil, C., Wu, Y., Andreassen, A., Lewkowycz, A., Misra, V., Ramasesh, V., Slone, A., Gur-Ari, G., Dyer, E., and Neyshabur, B.
\newblock Exploring length generalization in large language models.
\newblock \emph{Advances in Neural Information Processing Systems}, 35:\penalty0 38546--38556, 2022.

\bibitem[Anthropic(2024)]{anthropic2024claude35sonnet}
Anthropic.
\newblock Claude 3.5 sonnet, June 2024.
\newblock URL \url{https://www.anthropic.com/claude/sonnet}.
\newblock Artificial intelligence model.

\bibitem[Antol et~al.(2015)Antol, Agrawal, Lu, Mitchell, Batra, Zitnick, and Parikh]{antol2015vqa}
Antol, S., Agrawal, A., Lu, J., Mitchell, M., Batra, D., Zitnick, C.~L., and Parikh, D.
\newblock Vqa: Visual question answering.
\newblock In \emph{Proceedings of the IEEE international conference on computer vision}, pp.\  2425--2433, 2015.

\bibitem[Bai et~al.(2025)Bai, Chen, Liu, Wang, Ge, Song, Dang, Wang, Wang, Tang, Zhong, Zhu, Yang, Li, Wan, Wang, Ding, Fu, Xu, Ye, Zhang, Xie, Cheng, Zhang, Yang, Xu, and Lin]{bai2025qwen25vltechnicalreport}
Bai, S., Chen, K., Liu, X., Wang, J., Ge, W., Song, S., Dang, K., Wang, P., Wang, S., Tang, J., Zhong, H., Zhu, Y., Yang, M., Li, Z., Wan, J., Wang, P., Ding, W., Fu, Z., Xu, Y., Ye, J., Zhang, X., Xie, T., Cheng, Z., Zhang, H., Yang, Z., Xu, H., and Lin, J.
\newblock Qwen2.5-vl technical report, 2025.
\newblock URL \url{https://arxiv.org/abs/2502.13923}.

\bibitem[Barrett et~al.(2018)Barrett, Hill, Santoro, Morcos, and Lillicrap]{barrett2018measuring}
Barrett, D., Hill, F., Santoro, A., Morcos, A., and Lillicrap, T.
\newblock Measuring abstract reasoning in neural networks.
\newblock In \emph{Proceedings of the 35th International Conference on Machine Learning}, pp.\  511--520. PMLR, 2018.

\bibitem[Bhattamishra et~al.(2020)Bhattamishra, Ahuja, and Goyal]{bhattamishra2020ability}
Bhattamishra, S., Ahuja, K., and Goyal, N.
\newblock On the ability and limitations of transformers to recognize formal languages.
\newblock In Webber, B., Cohn, T., He, Y., and Liu, Y. (eds.), \emph{Proceedings of the 2020 Conference on Empirical Methods in Natural Language Processing (EMNLP)}, pp.\  7096--7116, Online, November 2020. Association for Computational Linguistics.
\newblock \doi{10.18653/v1/2020.emnlp-main.576}.
\newblock URL \url{https://aclanthology.org/2020.emnlp-main.576/}.

\bibitem[Burns et~al.(2023)Burns, Izmailov, Kirchner, Baker, Gao, Aschenbrenner, Chen, Ecoffet, Joglekar, Leike, et~al.]{burns2023weak}
Burns, C., Izmailov, P., Kirchner, J.~H., Baker, B., Gao, L., Aschenbrenner, L., Chen, Y., Ecoffet, A., Joglekar, M., Leike, J., et~al.
\newblock Weak-to-strong generalization: Eliciting strong capabilities with weak supervision.
\newblock \emph{arXiv preprint arXiv:2312.09390}, 2023.

\bibitem[Carbune et~al.(2024)Carbune, Mansoor, Liu, Aralikatte, Baechler, Chen, and Sharma]{carbune2024chart}
Carbune, V., Mansoor, H., Liu, F., Aralikatte, R., Baechler, G., Chen, J., and Sharma, A.
\newblock Chart-based reasoning: Transferring capabilities from {LLM}s to {VLM}s.
\newblock In Duh, K., Gomez, H., and Bethard, S. (eds.), \emph{Findings of the Association for Computational Linguistics: NAACL 2024}, pp.\  989--1004, Mexico City, Mexico, June 2024. Association for Computational Linguistics.
\newblock \doi{10.18653/v1/2024.findings-naacl.62}.
\newblock URL \url{https://aclanthology.org/2024.findings-naacl.62/}.

\bibitem[Chen et~al.(2015)Chen, Fang, Lin, Vedantam, Gupta, Doll{\'a}r, and Zitnick]{chen2015microsoft}
Chen, X., Fang, H., Lin, T.-Y., Vedantam, R., Gupta, S., Doll{\'a}r, P., and Zitnick, C.~L.
\newblock Microsoft coco captions: Data collection and evaluation server.
\newblock \emph{arXiv preprint arXiv:1504.00325}, 2015.

\bibitem[Cherian et~al.(2023)Cherian, Peng, Lohit, Smith, and Tenenbaum]{cherian2023deep}
Cherian, A., Peng, K.-C., Lohit, S., Smith, K.~A., and Tenenbaum, J.~B.
\newblock Are deep neural networks smarter than second graders?
\newblock In \emph{Proceedings of the IEEE/CVF Conference on Computer Vision and Pattern Recognition}, pp.\  10834--10844, 2023.

\bibitem[Deng et~al.(2009)Deng, Dong, Socher, Li, Li, and Fei-Fei]{5206848}
Deng, J., Dong, W., Socher, R., Li, L.-J., Li, K., and Fei-Fei, L.
\newblock Imagenet: A large-scale hierarchical image database.
\newblock In \emph{2009 IEEE Conference on Computer Vision and Pattern Recognition}, pp.\  248--255, 2009.
\newblock \doi{10.1109/CVPR.2009.5206848}.

\bibitem[Deng et~al.(2024)Deng, Choi, and Shieber]{deng2024explicit}
Deng, Y., Choi, Y., and Shieber, S.
\newblock From explicit cot to implicit cot: Learning to internalize cot step by step.
\newblock \emph{arXiv preprint arXiv:2405.14838}, 2024.

\bibitem[Duan et~al.(2024)Duan, Yang, Qiao, Fang, Chen, Liu, Dong, Zang, Zhang, Wang, Lin, and Chen]{duan2024vlmevalkit}
Duan, H., Yang, J., Qiao, Y., Fang, X., Chen, L., Liu, Y., Dong, X., Zang, Y., Zhang, P., Wang, J., Lin, D., and Chen, K.
\newblock Vlmevalkit: An open-source toolkit for evaluating large multi-modality models, 2024.

\bibitem[Dubey et~al.(2024)Dubey, Jauhri, Pandey, Kadian, Al-Dahle, Letman, Mathur, Schelten, Yang, Fan, et~al.]{dubey2024llama}
Dubey, A., Jauhri, A., Pandey, A., Kadian, A., Al-Dahle, A., Letman, A., Mathur, A., Schelten, A., Yang, A., Fan, A., et~al.
\newblock The llama 3 herd of models.
\newblock \emph{arXiv preprint arXiv:2407.21783}, 2024.

\bibitem[Dziri et~al.(2024)Dziri, Lu, Sclar, Li, Jiang, Lin, Welleck, West, Bhagavatula, Le~Bras, et~al.]{dziri2024faith}
Dziri, N., Lu, X., Sclar, M., Li, X.~L., Jiang, L., Lin, B.~Y., Welleck, S., West, P., Bhagavatula, C., Le~Bras, R., et~al.
\newblock Faith and fate: Limits of transformers on compositionality.
\newblock \emph{Advances in Neural Information Processing Systems}, 36, 2024.

\bibitem[Engstrom et~al.(2024)Engstrom, Feldmann, and Madry]{engstrom2024dsdm}
Engstrom, L., Feldmann, A., and Madry, A.
\newblock Dsdm: Model-aware dataset selection with datamodels.
\newblock In \emph{Forty-first International Conference on Machine Learning}, 2024.
\newblock URL \url{https://openreview.net/forum?id=GC8HkKeH8s}.

\bibitem[Fan et~al.(2024)Fan, Chen, Liu, Yuan, and Sigal]{fan2024pre}
Fan, W.-C., Chen, Y.-C., Liu, M., Yuan, L., and Sigal, L.
\newblock On pre-training of multimodal language models customized for chart understanding.
\newblock \emph{arXiv preprint arXiv:2407.14506}, 2024.

\bibitem[Fan et~al.(2023)Fan, Xu, Wang, Wang, and Guo]{Fan_2023_CVPR}
Fan, Y., Xu, W., Wang, H., Wang, J., and Guo, S.
\newblock Pmr: Prototypical modal rebalance for multimodal learning.
\newblock In \emph{Proceedings of the IEEE/CVF Conference on Computer Vision and Pattern Recognition (CVPR)}, pp.\  20029--20038, June 2023.

\bibitem[Fan et~al.(2025)Fan, Du, Ramchandran, and Lee]{fan2025looped}
Fan, Y., Du, Y., Ramchandran, K., and Lee, K.
\newblock Looped transformers for length generalization.
\newblock In \emph{The Thirteenth International Conference on Learning Representations}, 2025.
\newblock URL \url{https://openreview.net/forum?id=2edigk8yoU}.

\bibitem[Fu et~al.(2023)Fu, Chen, Shen, Qin, Zhang, Lin, Qiu, Lin, Yang, Zheng, Li, Sun, and Ji]{Fu2023MME}
Fu, C., Chen, P., Shen, Y., Qin, Y., Zhang, M., Lin, X., Qiu, Z., Lin, W., Yang, J., Zheng, X., Li, K., Sun, X., and Ji, R.
\newblock Mme: A comprehensive evaluation benchmark for multimodal large language models.
\newblock \emph{arXiv preprint arXiv:2306.13394}, June 2023.
\newblock URL \url{https://arxiv.org/abs/2306.13394}.

\bibitem[Fu et~al.(2024)Fu, Guo, Khalighinejad, Liu, Dhingra, Yogatama, Jia, and Neiswanger]{fu2024isobench}
Fu, D., Guo, R., Khalighinejad, G., Liu, O., Dhingra, B., Yogatama, D., Jia, R., and Neiswanger, W.
\newblock Isobench: Benchmarking multimodal foundation models on isomorphic representations.
\newblock In \emph{First Conference on Language Modeling}, 2024.
\newblock URL \url{https://openreview.net/forum?id=KZd1EErRJ1}.

\bibitem[Gao et~al.(2024)Gao, Wettig, Yen, and Chen]{gao2024prolong}
Gao, T., Wettig, A., Yen, H., and Chen, D.
\newblock How to train long-context language models (effectively).
\newblock \emph{arXiv preprint arXiv:2410.02660}, 2024.

\bibitem[Ghosal et~al.(2024)Ghosal, Han, Ken, and Poria]{ghosal2024language}
Ghosal, D., Han, V. T.~Y., Ken, C.~Y., and Poria, S.
\newblock Are language models puzzle prodigies? algorithmic puzzles unveil serious challenges in multimodal reasoning.
\newblock \emph{arXiv preprint arXiv:2403.03864}, 2024.

\bibitem[Goyal et~al.(2017)Goyal, Khot, Summers-Stay, Batra, and Parikh]{goyal2017making}
Goyal, Y., Khot, T., Summers-Stay, D., Batra, D., and Parikh, D.
\newblock Making the v in vqa matter: Elevating the role of image understanding in visual question answering.
\newblock In \emph{Proceedings of the IEEE conference on computer vision and pattern recognition}, pp.\  6904--6913, 2017.

\bibitem[Hill et~al.(2019)Hill, Santoro, Barrett, Morcos, and Lillicrap]{hill2019learning}
Hill, F., Santoro, A., Barrett, D., Morcos, A., and Lillicrap, T.
\newblock Learning to make analogies by contrasting abstract relational structure.
\newblock In \emph{International Conference on Learning Representations}, 2019.

\bibitem[Hsieh et~al.(2024)Hsieh, Zhang, Ma, Kembhavi, and Krishna]{hsieh2024sugarcrepe}
Hsieh, C.-Y., Zhang, J., Ma, Z., Kembhavi, A., and Krishna, R.
\newblock Sugarcrepe: Fixing hackable benchmarks for vision-language compositionality.
\newblock \emph{Advances in neural information processing systems}, 36, 2024.

\bibitem[Huang et~al.(2023)Huang, Sun, Xie, Li, and Liu]{huang2023t2i}
Huang, K., Sun, K., Xie, E., Li, Z., and Liu, X.
\newblock T2i-compbench: A comprehensive benchmark for open-world compositional text-to-image generation.
\newblock \emph{Advances in Neural Information Processing Systems}, 36:\penalty0 78723--78747, 2023.

\bibitem[Huang et~al.(2022)Huang, Lin, Zhou, Yang, and Huang]{pmlr-v162-huang22e}
Huang, Y., Lin, J., Zhou, C., Yang, H., and Huang, L.
\newblock Modality competition: What makes joint training of multi-modal network fail in deep learning? ({P}rovably).
\newblock In Chaudhuri, K., Jegelka, S., Song, L., Szepesvari, C., Niu, G., and Sabato, S. (eds.), \emph{Proceedings of the 39th International Conference on Machine Learning}, volume 162 of \emph{Proceedings of Machine Learning Research}, pp.\  9226--9259. PMLR, 17--23 Jul 2022.

\bibitem[Jelassi et~al.(2023)Jelassi, d'Ascoli, Domingo-Enrich, Wu, Li, and Charton]{jelassi2023length}
Jelassi, S., d'Ascoli, S., Domingo-Enrich, C., Wu, Y., Li, Y., and Charton, F.
\newblock Length generalization in arithmetic transformers.
\newblock \emph{arXiv preprint arXiv:2306.15400}, 2023.

\bibitem[Kazemnejad et~al.(2024)Kazemnejad, Padhi, Natesan~Ramamurthy, Das, and Reddy]{kazemnejad2024impact}
Kazemnejad, A., Padhi, I., Natesan~Ramamurthy, K., Das, P., and Reddy, S.
\newblock The impact of positional encoding on length generalization in transformers.
\newblock \emph{Advances in Neural Information Processing Systems}, 36, 2024.

\bibitem[Kil et~al.(2024)Kil, Tavazoee, Kang, and Kim]{kil2024ii}
Kil, J., Tavazoee, F., Kang, D., and Kim, J.-K.
\newblock {II}-{MMR}: Identifying and improving multi-modal multi-hop reasoning in visual question answering.
\newblock In Ku, L.-W., Martins, A., and Srikumar, V. (eds.), \emph{Findings of the Association for Computational Linguistics: ACL 2024}, pp.\  10698--10709, Bangkok, Thailand, August 2024. Association for Computational Linguistics.
\newblock \doi{10.18653/v1/2024.findings-acl.636}.
\newblock URL \url{https://aclanthology.org/2024.findings-acl.636/}.

\bibitem[Kim et~al.(2024)Kim, Lee, Park, and Seo]{kim2024language}
Kim, D., Lee, J., Park, J., and Seo, M.
\newblock How language models extrapolate outside the training data: A case study in textualized gridworld.
\newblock \emph{arXiv preprint arXiv:2406.15275}, 2024.

\bibitem[Kingma \& Ba(2015)Kingma and Ba]{kingma2017adammethodstochasticoptimization}
Kingma, D.~P. and Ba, J.
\newblock Adam: A method for stochastic optimization.
\newblock In \emph{The Third International Conference for Learning Representations}, 2015.

\bibitem[Koh \& Liang(2017)Koh and Liang]{koh2017understanding}
Koh, P.~W. and Liang, P.
\newblock Understanding black-box predictions via influence functions.
\newblock In \emph{International conference on machine learning}, pp.\  1885--1894. PMLR, 2017.

\bibitem[Lee et~al.(2024)Lee, Sreenivasan, Lee, Lee, and Papailiopoulos]{lee2023teaching}
Lee, N., Sreenivasan, K., Lee, J.~D., Lee, K., and Papailiopoulos, D.
\newblock Teaching arithmetic to small transformers.
\newblock In \emph{The Twelfth International Conference on Learning Representations}, 2024.
\newblock URL \url{https://openreview.net/forum?id=dsUB4bst9S}.

\bibitem[Li \& McClelland(2023)Li and McClelland]{li2023representations}
Li, Y. and McClelland, J.
\newblock Representations and computations in transformers that support generalization on structured tasks.
\newblock \emph{Transactions on Machine Learning Research}, 2023.

\bibitem[Li et~al.(2023)Li, Du, Zhou, Wang, Zhao, and Wen]{Li-hallucination-2023}
Li, Y., Du, Y., Zhou, K., Wang, J., Zhao, X.~Z., and Wen, J.-R.
\newblock Evaluating object hallucination in large vision-language models.
\newblock In \emph{The 2023 Conference on Empirical Methods in Natural Language Processing}, 2023.

\bibitem[Liang et~al.(2021)Liang, Wu, Ziyin, Morency, and Salakhutdinov]{liang2021cross}
Liang, P.~P., Wu, P., Ziyin, L., Morency, L.-P., and Salakhutdinov, R.
\newblock Cross-modal generalization: Learning in low resource modalities via meta-alignment.
\newblock In \emph{Proceedings of the 29th ACM International Conference on Multimedia}, pp.\  2680--2689, 2021.

\bibitem[Lin et~al.(2024)Lin, Wang, Cai, Liu, Fu, Tang, Yu, and Kot]{lin2024suppress}
Lin, X., Wang, S., Cai, R., Liu, Y., Fu, Y., Tang, W., Yu, Z., and Kot, A.
\newblock Suppress and rebalance: Towards generalized multi-modal face anti-spoofing.
\newblock In \emph{Proceedings of the IEEE/CVF Conference on Computer Vision and Pattern Recognition}, pp.\  211--221, 2024.

\bibitem[Liu et~al.(2023{\natexlab{a}})Liu, Ash, Goel, Krishnamurthy, and Zhang]{liu2022transformers}
Liu, B., Ash, J.~T., Goel, S., Krishnamurthy, A., and Zhang, C.
\newblock Transformers learn shortcuts to automata.
\newblock In \emph{The Eleventh International Conference on Learning Representations}, 2023{\natexlab{a}}.
\newblock URL \url{https://openreview.net/forum?id=De4FYqjFueZ}.

\bibitem[Liu et~al.(2023{\natexlab{b}})Liu, Li, Li, and Lee]{liu2023LLaVA1.5}
Liu, H., Li, C., Li, Y., and Lee, Y.~J.
\newblock Improved baselines with visual instruction tuning.
\newblock In \emph{NeurIPS 2023 Workshop on Instruction Tuning and Instruction Following}, 2023{\natexlab{b}}.

\bibitem[Liu et~al.(2023{\natexlab{c}})Liu, Li, Wu, and Lee]{liu2023llava}
Liu, H., Li, C., Wu, Q., and Lee, Y.~J.
\newblock Visual instruction tuning.
\newblock In \emph{Thirty-seventh Conference on Neural Information Processing Systems}, 2023{\natexlab{c}}.

\bibitem[Liu et~al.(2024{\natexlab{a}})Liu, Li, Li, Li, Zhang, Shen, and Lee]{liu2024llavanext}
Liu, H., Li, C., Li, Y., Li, B., Zhang, Y., Shen, S., and Lee, Y.~J.
\newblock Llava-next: Improved reasoning, ocr, and world knowledge, January 2024{\natexlab{a}}.

\bibitem[Liu et~al.(2024{\natexlab{b}})Liu, Li, Huang, Yang, Yu, Li, Yin, Liu, Jin, and Bai]{liu2024ocrbenchhiddenmysteryocr}
Liu, Y., Li, Z., Huang, M., Yang, B., Yu, W., Li, C., Yin, X.-C., Liu, C.-L., Jin, L., and Bai, X.
\newblock Ocrbench: on the hidden mystery of ocr in large multimodal models.
\newblock \emph{Science China Information Sciences}, 67\penalty0 (12), December 2024{\natexlab{b}}.
\newblock ISSN 1869-1919.
\newblock \doi{10.1007/s11432-024-4235-6}.
\newblock URL \url{http://dx.doi.org/10.1007/s11432-024-4235-6}.

\bibitem[Liu et~al.(2025)Liu, Duan, Zhang, Li, Zhang, Zhao, Yuan, Wang, He, Liu, et~al.]{liu2025mmbench}
Liu, Y., Duan, H., Zhang, Y., Li, B., Zhang, S., Zhao, W., Yuan, Y., Wang, J., He, C., Liu, Z., et~al.
\newblock Mmbench: Is your multi-modal model an all-around player?
\newblock In \emph{European Conference on Computer Vision}, pp.\  216--233. Springer, 2025.

\bibitem[Liu et~al.(2022)Liu, Mao, Wu, Feichtenhofer, Darrell, and Xie]{liu2022convnet}
Liu, Z., Mao, H., Wu, C.-Y., Feichtenhofer, C., Darrell, T., and Xie, S.
\newblock A convnet for the 2020s.
\newblock In \emph{Proceedings of the IEEE/CVF conference on computer vision and pattern recognition}, pp.\  11976--11986, 2022.

\bibitem[McLeish et~al.(2024)McLeish, Schwarzschild, and Goldstein]{mcleish2024benchmarking}
McLeish, S., Schwarzschild, A., and Goldstein, T.
\newblock Benchmarking chatgpt on algorithmic reasoning.
\newblock \emph{arXiv preprint arXiv:2404.03441}, 2024.

\bibitem[Mindermann et~al.(2022)Mindermann, Brauner, Razzak, Sharma, Kirsch, Xu, H{\"o}ltgen, Gomez, Morisot, Farquhar, et~al.]{mindermann2022prioritized}
Mindermann, S., Brauner, J.~M., Razzak, M.~T., Sharma, M., Kirsch, A., Xu, W., H{\"o}ltgen, B., Gomez, A.~N., Morisot, A., Farquhar, S., et~al.
\newblock Prioritized training on points that are learnable, worth learning, and not yet learnt.
\newblock In \emph{International Conference on Machine Learning}, pp.\  15630--15649. PMLR, 2022.

\bibitem[Monajatipoor et~al.(2023)Monajatipoor, Li, Rouhsedaghat, Yang, and Chang]{monajatipoor-etal-2023-metavl}
Monajatipoor, M., Li, L.~H., Rouhsedaghat, M., Yang, L., and Chang, K.-W.
\newblock {M}eta{VL}: Transferring in-context learning ability from language models to vision-language models.
\newblock In Rogers, A., Boyd-Graber, J., and Okazaki, N. (eds.), \emph{Proceedings of the 61st Annual Meeting of the Association for Computational Linguistics (Volume 2: Short Papers)}, pp.\  495--508, Toronto, Canada, July 2023. Association for Computational Linguistics.
\newblock \doi{10.18653/v1/2023.acl-short.43}.

\bibitem[Nesterov(2018)]{nesterov2018lectures}
Nesterov, Y.
\newblock Lectures on convex optimization.
\newblock \emph{Springer Optimization and Its Applications}, 137, 2018.

\bibitem[Nguyen et~al.(2024)Nguyen, Le, Mai, and Le]{nguyen2024ada2i}
Nguyen, C.-V.~T., Le, T.-S., Mai, A.-T., and Le, D.-T.
\newblock Ada2i: Enhancing modality balance for multimodal conversational emotion recognition.
\newblock In \emph{Proceedings of the 32nd ACM International Conference on Multimedia}, pp.\  9330--9339, 2024.

\bibitem[Park et~al.(2023)Park, Georgiev, Ilyas, Leclerc, and Madry]{park2023trak}
Park, S.~M., Georgiev, K., Ilyas, A., Leclerc, G., and Madry, A.
\newblock Trak: Attributing model behavior at scale.
\newblock In \emph{International Conference on Machine Learning (ICML)}, 2023.

\bibitem[Peng et~al.(2022)Peng, Wei, Deng, Wang, and Hu]{peng2022balanced}
Peng, X., Wei, Y., Deng, A., Wang, D., and Hu, D.
\newblock Balanced multimodal learning via on-the-fly gradient modulation.
\newblock In \emph{Proceedings of the IEEE/CVF conference on computer vision and pattern recognition}, pp.\  8238--8247, 2022.

\bibitem[Radford et~al.(2021)Radford, Kim, Hallacy, Ramesh, Goh, Agarwal, Sastry, Askell, Mishkin, Clark, et~al.]{radford2021learning}
Radford, A., Kim, J.~W., Hallacy, C., Ramesh, A., Goh, G., Agarwal, S., Sastry, G., Askell, A., Mishkin, P., Clark, J., et~al.
\newblock Learning transferable visual models from natural language supervision.
\newblock In \emph{International conference on machine learning}, pp.\  8748--8763. PMLR, 2021.

\bibitem[Rahmanzadehgervi et~al.(2024)Rahmanzadehgervi, Bolton, Taesiri, and Nguyen]{rahmanzadehgervi2024vision}
Rahmanzadehgervi, P., Bolton, L., Taesiri, M.~R., and Nguyen, A.~T.
\newblock Vision language models are blind.
\newblock In \emph{Proceedings of the Asian Conference on Computer Vision}, pp.\  18--34, 2024.

\bibitem[Rasley et~al.(2020)Rasley, Rajbhandari, Ruwase, and He]{rasley2020deepspeed}
Rasley, J., Rajbhandari, S., Ruwase, O., and He, Y.
\newblock Deepspeed: System optimizations enable training deep learning models with over 100 billion parameters.
\newblock In \emph{Proceedings of the 26th ACM SIGKDD International Conference on Knowledge Discovery \& Data Mining}, KDD '20, pp.\  3505–3506, New York, NY, USA, 2020. Association for Computing Machinery.
\newblock ISBN 9781450379984.
\newblock \doi{10.1145/3394486.3406703}.
\newblock URL \url{https://doi.org/10.1145/3394486.3406703}.

\bibitem[Sanford et~al.(2024)Sanford, Fatemi, Hall, Tsitsulin, Kazemi, Halcrow, Perozzi, and Mirrokni]{sanford2024understanding}
Sanford, C., Fatemi, B., Hall, E., Tsitsulin, A., Kazemi, M., Halcrow, J., Perozzi, B., and Mirrokni, V.
\newblock Understanding transformer reasoning capabilities via graph algorithms.
\newblock In \emph{The Thirty-eighth Annual Conference on Neural Information Processing Systems}, 2024.
\newblock URL \url{https://openreview.net/forum?id=AfzbDw6DSp}.

\bibitem[Shi et~al.(2025)Shi, Liu, Wang, Liao, Radhakrishnan, Zhao, Huang, Yin, Sapra, Yacoob, Shi, Catanzaro, Tao, Kautz, Yu, and Liu]{shi2024EAGLE}
Shi, M., Liu, F., Wang, S., Liao, S., Radhakrishnan, S., Zhao, Y., Huang, D.-A., Yin, H., Sapra, K., Yacoob, Y., Shi, H., Catanzaro, B., Tao, A., Kautz, J., Yu, Z., and Liu, G.
\newblock Eagle: Exploring the design space for multimodal {LLM}s with mixture of encoders.
\newblock In \emph{The Thirteenth International Conference on Learning Representations}, 2025.
\newblock URL \url{https://openreview.net/forum?id=Y2RW9EVwhT}.

\bibitem[Singh et~al.(2019)Singh, Natarajan, Shah, Jiang, Chen, Batra, Parikh, and Rohrbach]{singh2019towards}
Singh, A., Natarajan, V., Shah, M., Jiang, Y., Chen, X., Batra, D., Parikh, D., and Rohrbach, M.
\newblock Towards vqa models that can read.
\newblock In \emph{Proceedings of the IEEE/CVF Conference on Computer Vision and Pattern Recognition}, June 2019.

\bibitem[Socher et~al.(2013)Socher, Ganjoo, Manning, and Ng]{socher2013zero}
Socher, R., Ganjoo, M., Manning, C.~D., and Ng, A.
\newblock Zero-shot learning through cross-modal transfer.
\newblock \emph{Advances in neural information processing systems}, 26, 2013.

\bibitem[Su et~al.(2025)Su, Sukhbaatar, Rabbat, Tian, and Zheng]{su2024dualformer}
Su, D., Sukhbaatar, S., Rabbat, M., Tian, Y., and Zheng, Q.
\newblock Dualformer: Controllable fast and slow thinking by learning with randomized reasoning traces.
\newblock In \emph{The Thirteenth International Conference on Learning Representations}, 2025.
\newblock URL \url{https://openreview.net/forum?id=bmbRCRiNDu}.

\bibitem[Sun et~al.(2024)Sun, Yu, Shen, Liu, Yang, Welleck, and Gan]{sun2024easy}
Sun, Z., Yu, L., Shen, Y., Liu, W., Yang, Y., Welleck, S., and Gan, C.
\newblock Easy-to-hard generalization: Scalable alignment beyond human supervision.
\newblock In \emph{The Thirty-eighth Annual Conference on Neural Information Processing Systems}, 2024.
\newblock URL \url{https://openreview.net/forum?id=qwgfh2fTtN}.

\bibitem[Tan \& Bansal(2020)Tan and Bansal]{tan2020vokenization}
Tan, H. and Bansal, M.
\newblock Vokenization: Improving language understanding with contextualized, visual-grounded supervision.
\newblock In Webber, B., Cohn, T., He, Y., and Liu, Y. (eds.), \emph{Proceedings of the 2020 Conference on Empirical Methods in Natural Language Processing (EMNLP)}, pp.\  2066--2080, Online, November 2020. Association for Computational Linguistics.
\newblock \doi{10.18653/v1/2020.emnlp-main.162}.
\newblock URL \url{https://aclanthology.org/2020.emnlp-main.162/}.

\bibitem[Taylor et~al.(2024)Taylor, Cuturrufo, Yathish, Ma, and Wang]{taylor2024large}
Taylor, A.~K., Cuturrufo, A., Yathish, V., Ma, M.~D., and Wang, W.
\newblock Are large-language models graph algorithmic reasoners?
\newblock \emph{arXiv preprint arXiv:2410.22597}, 2024.

\bibitem[Thrush et~al.(2022)Thrush, Jiang, Bartolo, Singh, Williams, Kiela, and Ross]{thrush2022winoground}
Thrush, T., Jiang, R., Bartolo, M., Singh, A., Williams, A., Kiela, D., and Ross, C.
\newblock Winoground: Probing vision and language models for visio-linguistic compositionality.
\newblock In \emph{Proceedings of the IEEE/CVF Conference on Computer Vision and Pattern Recognition}, pp.\  5238--5248, 2022.

\bibitem[Tong et~al.(2024)Tong, II, Wu, Woo, IYER, Akula, Yang, Yang, Middepogu, Wang, Pan, Fergus, LeCun, and Xie]{tong2024cambrian1}
Tong, S., II, E. L.~B., Wu, P., Woo, S., IYER, A.~J., Akula, S.~C., Yang, S., Yang, J., Middepogu, M., Wang, Z., Pan, X., Fergus, R., LeCun, Y., and Xie, S.
\newblock Cambrian-1: A fully open, vision-centric exploration of multimodal {LLM}s.
\newblock In \emph{The Thirty-eighth Annual Conference on Neural Information Processing Systems}, 2024.
\newblock URL \url{https://openreview.net/forum?id=Vi8AepAXGy}.

\bibitem[Wang et~al.(2024{\natexlab{a}})Wang, Feng, He, Tan, Han, and Tsvetkov]{wang2024can}
Wang, H., Feng, S., He, T., Tan, Z., Han, X., and Tsvetkov, Y.
\newblock Can language models solve graph problems in natural language?
\newblock \emph{Advances in Neural Information Processing Systems}, 36, 2024{\natexlab{a}}.

\bibitem[Wang et~al.(2024{\natexlab{b}})Wang, Ming, Shi, Vineet, Wang, Li, and Joshi]{wang2024picture}
Wang, J., Ming, Y., Shi, Z., Vineet, V., Wang, X., Li, Y., and Joshi, N.
\newblock Is a picture worth a thousand words? delving into spatial reasoning for vision language models.
\newblock In \emph{The Thirty-eighth Annual Conference on Neural Information Processing Systems}, 2024{\natexlab{b}}.
\newblock URL \url{https://openreview.net/forum?id=cvaSru8LeO}.

\bibitem[Wang et~al.(2020)Wang, Tran, and Feiszli]{wang2020makes}
Wang, W., Tran, D., and Feiszli, M.
\newblock What makes training multi-modal classification networks hard?
\newblock In \emph{Proceedings of the IEEE/CVF conference on computer vision and pattern recognition}, pp.\  12695--12705, 2020.

\bibitem[Wang et~al.(2024{\natexlab{c}})Wang, Xia, He, Chen, Liu, Zhu, Liang, Wu, Liu, Malladi, Chevalier, Arora, and Chen]{wang2024charxiv}
Wang, Z., Xia, M., He, L., Chen, H., Liu, Y., Zhu, R., Liang, K., Wu, X., Liu, H., Malladi, S., Chevalier, A., Arora, S., and Chen, D.
\newblock Charxiv: Charting gaps in realistic chart understanding in multimodal {LLM}s.
\newblock In \emph{The Thirty-eight Conference on Neural Information Processing Systems Datasets and Benchmarks Track}, 2024{\natexlab{c}}.
\newblock URL \url{https://openreview.net/forum?id=cy8mq7QYae}.

\bibitem[Wei et~al.(2024)Wei, Feng, Wang, and Hu]{Wei_2024_CVPR}
Wei, Y., Feng, R., Wang, Z., and Hu, D.
\newblock Enhancing multimodal cooperation via sample-level modality valuation.
\newblock In \emph{Proceedings of the IEEE/CVF Conference on Computer Vision and Pattern Recognition (CVPR)}, pp.\  27338--27347, June 2024.

\bibitem[Weiss et~al.(2021)Weiss, Goldberg, and Yahav]{weiss2021thinking}
Weiss, G., Goldberg, Y., and Yahav, E.
\newblock Thinking like transformers.
\newblock In \emph{International Conference on Machine Learning}, pp.\  11080--11090. PMLR, 2021.

\bibitem[Wu et~al.(2024{\natexlab{a}})Wu, Zhao, Saxon, Bui, Wang, Zhang, and Chang]{wu2024vsp}
Wu, Q., Zhao, H., Saxon, M., Bui, T., Wang, W.~Y., Zhang, Y., and Chang, S.
\newblock Vsp: Assessing the dual challenges of perception and reasoning in spatial planning tasks for vlms.
\newblock \emph{arXiv preprint arXiv:2407.01863}, 2024{\natexlab{a}}.

\bibitem[Wu et~al.(2024{\natexlab{b}})Wu, Yu, Huang, Russakovsky, and Arora]{wu2024conceptmix}
Wu, X., Yu, D., Huang, Y., Russakovsky, O., and Arora, S.
\newblock Conceptmix: A compositional image generation benchmark with controllable difficulty.
\newblock In \emph{The Thirty-eight Conference on Neural Information Processing Systems Datasets and Benchmarks Track}, 2024{\natexlab{b}}.
\newblock URL \url{https://openreview.net/forum?id=MU2s9wwWLo}.

\bibitem[Xia et~al.(2024)Xia, Malladi, Gururangan, Arora, and Chen]{xia2024less}
Xia, M., Malladi, S., Gururangan, S., Arora, S., and Chen, D.
\newblock {LESS}: Selecting influential data for targeted instruction tuning.
\newblock In \emph{International Conference on Machine Learning (ICML)}, 2024.

\bibitem[Xia et~al.(2023)Xia, Huang, Zhu, and Zhao]{xia2023achieving}
Xia, Y., Huang, H., Zhu, J., and Zhao, Z.
\newblock Achieving cross modal generalization with multimodal unified representation.
\newblock In \emph{Thirty-seventh Conference on Neural Information Processing Systems}, 2023.

\bibitem[Xie et~al.(2024)Xie, Pham, Dong, Du, Liu, Lu, Liang, Le, Ma, and Yu]{xie2024doremi}
Xie, S.~M., Pham, H., Dong, X., Du, N., Liu, H., Lu, Y., Liang, P.~S., Le, Q.~V., Ma, T., and Yu, A.~W.
\newblock Doremi: Optimizing data mixtures speeds up language model pretraining.
\newblock \emph{Advances in Neural Information Processing Systems}, 36, 2024.

\bibitem[Yu et~al.(2024)Yu, Kaur, Gupta, Brown-Cohen, Goyal, and Arora]{yu2023skill}
Yu, D., Kaur, S., Gupta, A., Brown-Cohen, J., Goyal, A., and Arora, S.
\newblock {SKILL}-{MIX}: a flexible and expandable family of evaluations for {AI} models.
\newblock In \emph{The Twelfth International Conference on Learning Representations}, 2024.
\newblock URL \url{https://openreview.net/forum?id=Jf5gplvglq}.

\bibitem[Yue et~al.(2024)Yue, Ni, Zhang, Zheng, Liu, Zhang, Stevens, Jiang, Ren, Sun, et~al.]{yue2024mmmu}
Yue, X., Ni, Y., Zhang, K., Zheng, T., Liu, R., Zhang, G., Stevens, S., Jiang, D., Ren, W., Sun, Y., et~al.
\newblock Mmmu: A massive multi-discipline multimodal understanding and reasoning benchmark for expert agi.
\newblock In \emph{Proceedings of the IEEE/CVF Conference on Computer Vision and Pattern Recognition}, pp.\  9556--9567, 2024.

\bibitem[Yuksekgonul et~al.(2023)Yuksekgonul, Bianchi, Kalluri, Jurafsky, and Zou]{yuksekgonul2023and}
Yuksekgonul, M., Bianchi, F., Kalluri, P., Jurafsky, D., and Zou, J.
\newblock When and why vision-language models behave like bags-of-words, and what to do about it?
\newblock In \emph{The Eleventh International Conference on Learning Representations}, 2023.

\bibitem[Zhang et~al.(2024{\natexlab{a}})Zhang, Huang, Jin, and Lu]{zhang2024vision}
Zhang, J., Huang, J., Jin, S., and Lu, S.
\newblock Vision-language models for vision tasks: A survey.
\newblock \emph{IEEE Transactions on Pattern Analysis and Machine Intelligence}, 2024{\natexlab{a}}.

\bibitem[Zhang et~al.(2024{\natexlab{b}})Zhang, Zhai, Zhao, Zong, Wen, and Zhao]{zhang2024if}
Zhang, L., Zhai, X., Zhao, Z., Zong, Y., Wen, X., and Zhao, B.
\newblock What if the tv was off? examining counterfactual reasoning abilities of multi-modal language models.
\newblock In \emph{Proceedings of the IEEE/CVF Conference on Computer Vision and Pattern Recognition}, pp.\  21853--21862, 2024{\natexlab{b}}.

\bibitem[Zhang et~al.(2024{\natexlab{c}})Zhang, Jiang, Zhang, Lin, Guo, Qiu, Zhou, Lu, Chang, Gao, et~al.]{zhang2024mathverse}
Zhang, R., Jiang, D., Zhang, Y., Lin, H., Guo, Z., Qiu, P., Zhou, A., Lu, P., Chang, K.-W., Gao, P., et~al.
\newblock Mathverse: Does your multi-modal llm truly see the diagrams in visual math problems?
\newblock \emph{arXiv preprint arXiv:2403.14624}, 2024{\natexlab{c}}.

\bibitem[Zhang et~al.(2023)Zhang, Li, Wu, and Shi]{zhang2023lost}
Zhang, X., Li, S., Wu, Z., and Shi, N.
\newblock Lost in translation: When gpt-4v (ision) can't see eye to eye with text. a vision-language-consistency analysis of vllms and beyond.
\newblock \emph{arXiv preprint arXiv:2310.12520}, 2023.

\bibitem[Zhang et~al.(2024{\natexlab{d}})Zhang, Li, Shi, Hauer, Wu, Kondrak, Abdul-Mageed, and Lakshmanan]{zhang2024cross}
Zhang, X., Li, S., Shi, N., Hauer, B., Wu, Z., Kondrak, G., Abdul-Mageed, M., and Lakshmanan, L.~V.
\newblock Cross-modal consistency in multimodal large language models.
\newblock \emph{arXiv preprint arXiv:2411.09273}, 2024{\natexlab{d}}.

\bibitem[Zhang et~al.(2024{\natexlab{e}})Zhang, Yoon, Bansal, and Yao]{Zhang_2024_CVPR}
Zhang, X., Yoon, J., Bansal, M., and Yao, H.
\newblock Multimodal representation learning by alternating unimodal adaptation.
\newblock In \emph{Proceedings of the IEEE/CVF Conference on Computer Vision and Pattern Recognition (CVPR)}, pp.\  27456--27466, June 2024{\natexlab{e}}.

\bibitem[Zhang et~al.(2024{\natexlab{f}})Zhang, Wang, Zhang, Li, Qin, and Zhu]{zhang2024llm4dyg}
Zhang, Z., Wang, X., Zhang, Z., Li, H., Qin, Y., and Zhu, W.
\newblock Llm4dyg: can large language models solve spatial-temporal problems on dynamic graphs?
\newblock In \emph{Proceedings of the 30th ACM SIGKDD Conference on Knowledge Discovery and Data Mining}, pp.\  4350--4361, 2024{\natexlab{f}}.

\bibitem[Zhao et~al.(2024)Zhao, Kaur, Yu, Goyal, and Arora]{zhao2024can}
Zhao, H., Kaur, S., Yu, D., Goyal, A., and Arora, S.
\newblock Can models learn skill composition from examples?
\newblock In \emph{The Thirty-eighth Annual Conference on Neural Information Processing Systems}, 2024.
\newblock URL \url{https://openreview.net/forum?id=1sLdprsbmk}.

\bibitem[Zhao et~al.(2023)Zhao, Gu, Varma, Luo, Huang, Xu, Wright, Shojanazeri, Ott, Shleifer, Desmaison, Balioglu, Damania, Nguyen, Chauhan, Hao, Mathews, and Li]{zhao2023FSDP}
Zhao, Y., Gu, A., Varma, R., Luo, L., Huang, C.-C., Xu, M., Wright, L., Shojanazeri, H., Ott, M., Shleifer, S., Desmaison, A., Balioglu, C., Damania, P., Nguyen, B., Chauhan, G., Hao, Y., Mathews, A., and Li, S.
\newblock Pytorch fsdp: Experiences on scaling fully sharded data parallel, 2023.
\newblock URL \url{https://arxiv.org/abs/2304.11277}.

\bibitem[Zhou et~al.(2024{\natexlab{a}})Zhou, Bradley, Littwin, Razin, Saremi, Susskind, Bengio, and Nakkiran]{zhou2024LengthGeneralization}
Zhou, H., Bradley, A., Littwin, E., Razin, N., Saremi, O., Susskind, J.~M., Bengio, S., and Nakkiran, P.
\newblock What algorithms can transformers learn? a study in length generalization.
\newblock In \emph{The Twelfth International Conference on Learning Representations}, 2024{\natexlab{a}}.

\bibitem[Zhou et~al.(2024{\natexlab{b}})Zhou, Alon, Chen, Wang, Agarwal, and Zhou]{zhou2024transformers}
Zhou, Y., Alon, U., Chen, X., Wang, X., Agarwal, R., and Zhou, D.
\newblock Transformers can achieve length generalization but not robustly.
\newblock \emph{arXiv preprint arXiv:2402.09371}, 2024{\natexlab{b}}.

\end{thebibliography}
\bibliographystyle{icml2025}

\newpage
\appendix
\onecolumn

\addcontentsline{toc}{section}{Appendix} % Add the appendix text to the document TOC
\part{Appendix} % Start the appendix part
\parttoc % Insert the appendix TOC

\section{Appendix Structure}
\label{app:app_struct}

The appendix provides omitted experimental details, additional empirical explorations, and theoretical statements, which we outline below. 

\paragraph{Related works:} In \cref{sec:related_works}, we provide an overview of relevant lines of research in VLM benchmarks and evaluations, modality imbalance, cross-modal transfer of generalization, and simple-to-hard generalization. We highlight the contributions that differentiate our work from the similar ones.

\paragraph{Experimental details:} We provide all details of our synthetic data generation in \cref{app:synthetic}. We present our data generation algorithm for creating training data in \cref{sec:data-gen}, details on {\consecutivetableread}, {\tableread}, {\visualanalogy}, and {\gridnav} in \cref{sec:ctr_app,sec:tr_app,sec:ar_app,sec:gn_app} respectively. We show examples from our training data for each synthetic setting in \cref{fig:table-id,fig:table-ood,fig:ar-domain-transfer-heldout-id,fig:ar-domain-transfer-heldout-ood,fig:grid-id,fig:grid-ood}. We present details on training and evaluation in \cref{sec:training_details,sec:evaluation_details} respectively.

\paragraph{Consistent results on another model family and size:} We replicate some experiments from the main paper with \textit{Qwen2.5-VL-3B-Instruct} and \textit{Qwen2.5-VL-7B-Instruct}. We report the results in \cref{app:qwen}.

\paragraph{Continued discussion from main paper:} We continue the discussion in the main paper in \cref{sec:contd_discussion}. We present results on {\consecutivetableread} after normalizing the number of unique samples used across training strategies (\cref{sec:cmp_atequalTsimp}), present results on {\em Pattern-Heldout Visual Analogy} --- a {\textgeneralizable} version of {\visualanalogy} (\cref{sec:compositional_VA}), compare training strategies on {\nontextgeneralizable} tasks by normalizing the total number of training data used (\cref{sec:cmp_atequalT}), discuss further on transferring reasoning from image to text modality (\cref{sec:symbiote}), discuss further on text warm-up pretraining (\cref{app:ablations_continual_pretraining}), and report the utility of our created synthetic datasets for real-world benchmarks (\cref{sec:utility}).

\paragraph{Continued discussion on gradients:} We continue our discussion on gradient alignment in \cref{sec:grad_align}. We first show that the gradient alignment score connects to the expected drop in evaluation loss with SGD on training gradients (\cref{lem:grad_align_SGD}). We then propose results on additional measures --- gradient cosine similarity and Adam update alignment score (\cref{sec:adam_update}) --- that better capture the Adam gradient updates used for optimization.

\paragraph{Ablation studies:} We conduct extensive ablation studies to measure the effect of each experimental design decision in our training strategies on {\nontextgeneralizable} tasks and report the results in \cref{sec:ablate_nontextgen}. We report the performance on other multimodal models on our synthetic data (\cref{app:other_models}). We study design choices in {\cmft} (\cref{app:ablations_component}), design choices in {\alcmft} (\cref{sec:ablate_alcmft}), design choices in text warm-up pretraining (\cref{sec:ablate_text_warmup}), the effect of the choice of a text representation (\cref{app:text_representation}), the effect of text conversion (\cref{app:ablations_text_conversion}), the role of chain-of-thought (\cref{app:ablations_cot}), the effect of multi-task training (\cref{app:ablations_multitask}), and the effect of repeated training examples (\cref{app:unique_samples}).

\paragraph{Interpretability experiments:} We further conduct interpretability experiments on our trained models. We use gradient attribution to track the focus of the model on different image pixels during chain-of-thought generation (\cref{sec:interp_gradient}). We also report failure modes of models trained on our synthetic data when evaluated on {\hard} examples (\cref{sec:failure_modes}).

\clearpage
\section{Related Works}\label{sec:related_works}
\looseness-1\paragraph{Benchmarks and evaluations for VLMs}
VLMs are evaluated on benchmarks such as visual question answering (VQA) \citep{antol2015vqa}, image captioning \citep{chen2015microsoft}, zero-shot image classification \citep{5206848}, and compositional reasoning \citep{thrush2022winoground, yuksekgonul2023and, hsieh2024sugarcrepe}. However, these benchmarks often suffer from language bias, allowing solutions to use shortcuts with minimal visual information \citep{agrawal2016analyzing, goyal2017making, zhang2024mathverse}. Although recent work \citep{rahmanzadehgervi2024vision, wang2024picture, kil2024ii} proposed new benchmarks that aim to evaluate the spatial understanding and reasoning of VLM, most evaluation tasks are in the form of VQA questions that only require ``single-hop'' reasoning or relatively fewer reasoning steps. To create a controlled setting with well-defined \simple{} and \hard{} tasks, we focus on algorithmic visual reasoning tasks. These tasks allow us to precisely control the number of steps in the step-by-step reasoning process and the level of dynamic interaction between textual and visual inputs. Closely related works have explored graph-based algorithmic reasoning in LLMs \citep{taylor2024large, mcleish2024benchmarking, zhang2024llm4dyg, wang2024can, sanford2024understanding} but such studies remain limited for VLMs.

\paragraph{Modality imbalance}
Studies have shown that models exhibit different learning capabilities and learning speed on multimodal inputs \citep{wang2020makes, nguyen2024ada2i}. The imbalanced contribution of individual modality to the final prediction can result in overreliance on a few dominant, optimized modalities, while underutilizing signals of the weak ones. \citet{peng2022balanced} and \citet{lin2024suppress} attempt to rebalance the convergence speed of all modalities by modulating the learning rate or gradients. \citet{Fan_2023_CVPR} propose a representative embedding to guide the slow-learning modality and regularize the fast-learning one. \citet{Zhang_2024_CVPR} propose an alternating unimodal training to minimize interference between modalities. Despite their success in traditional multimodal joint training, it remains challenging to repeat the same for adapter-based VLMs due to significant differences in architecture and training pipeline. Our work aims to address this issue specifically for VLMs from the perspective of transferring the strong learning behaviors from the dominant modality (text) to the weak one (image).

\paragraph{Generalization transfer between input modes}
Given the high cost of training VLMs from scratch, recent research on adapter-based VLMs has been driven primarily by the idea of leveraging pretrained LLM backbones. The success of this approach is built on the idea of cross-modal generalization, which enables the model to harness information from the auxiliary modality (e.g. text) to improve unimodal task on the primary modality (e.g. image classification). This knowledge transference has been exploited for both small-scale multimodal models \citep{socher2013zero,liang2021cross,tan2020vokenization} and more recent VLMs \citep{monajatipoor-etal-2023-metavl,carbune2024chart,zhang2024vision}. However, existing works often require explicit alignment of the modality, such as learning unified representation using contrastive learning \citep{xia2023achieving}, for models to transfer knowledge across modalities. The cost of curating a large, perfectly aligned multimodal dataset to learn the modality alignment becomes expensive as the model size increases. In our work, we find that transfer of generalization across input modes naturally emerges from auto-regressive training.

\paragraph{{\simpletohard}}
Recent studies have explored simple-to-hard generalization in LLMs, with a focus on length generalization in transformers. These works evaluate models on tasks requiring longer computations than those seen during training, using synthetic datasets like parity, Dyck-1 languages, decimal addition, structural recursion, and finite state automata \citep{anil2022exploring,lee2023teaching,jelassi2023length,li2023representations,kazemnejad2024impact,liu2022transformers,abbe2024generalization,bhattamishra2020ability,zhou2024transformers,fan2025looped}.
\citet{zhou2024LengthGeneralization} connect length generalization to the RASP programming language \citep{weiss2021thinking}, offering a unified perspective. \citet{sun2024easy} recently propose easy-to-hard generalization to measure generalizable verification for math and code datasets. OOD generalization beyond human supervision remains an important open question for the advancement of current AI models \citep{burns2023weak}.

\clearpage
\section{Details on Synthetic Tasks}
\label{app:synthetic}

\begin{table}[!t]
    \caption{
    \textbf{Summary of the {\simple} and {\hard} task setup} for {\tableread}, {\gridnav}, and {\visualanalogy}}
    \label{tab:avr_summary}
    \centering
    \begin{tabular}{c||c||c|c}
        \toprule
        Setting & Attribute & {\simple} & {\hard} \\
        \midrule
        \multirow{4}{*}{\tableread} & Mean Length & $12$  & $35$ \\ & {\# Turns} & $1-4$ & $> 4$ \\ & \multirow{2}{*}{Pattern} & Spiral & {Composition of} \\ & & {/ Sinusoidal} & {Spiral / Sinusoidal} \\
        \hline
        \multirow{3}{*}{\gridnav} & \# DFS steps & $[10, 25]$ & $[26, 60]$ \\
        & {\# Objects} & $\{ 1,2 \}$ & $\{  2,3,4,5 \}$ \\ & {\# Obstacle type} & $\{ 1 \}$ & $\{ 3,4,5 \}$ \\
        \hline
        \multirow{2}{*}{\visualanalogy} & {Example Patterns} & Same & Different \\ & {Query Pattern} & Seen & Held-out \\
        \bottomrule
    \end{tabular}
\end{table}

\subsection{Formal description of data generation}
\label{sec:data-gen}

\begin{figure}[ht]
\centering
\begin{algorithm}[H]
\small
\caption{Data generation pipeline for main experiments}
\label{alg:data_generation}
\begin{algorithmic} 
 \REQUIRE Task $f: \inputset \to \reasonset$, Dataset $\inputset = \inputset_{\simple{}} \cup \inputset_{\hard{}}$, Number of data to generate $N$, Type of supervision $s$.

\IF{$s \in \{$ {\smfttext}, {\smftimage} $\}$}
    \STATE Initialize the number of data per difficulty $N_{\simple{}} = N$, $N_{\hard{}} = 0$ and the number of unique examples $N_{\simple}^{u} = N_{\simple{}}$
\ELSIF{$s \in\{$ {\imagetext{}}, {\imageviatext{}}, {\mixft} $\}$}
    \STATE Initialize the number of data per difficulty $N_{\simple{}} = N$, $N_{\hard{}} = 0$ and the number of unique examples $N_{\simple}^{u} = \frac{N_{\simple{}}}{3}$
\ELSIF{$s \in\{$ {\cimageviatext}, {\cmft} $\}$}
    \STATE Initialize the number of data per difficulty $N_{\simple{}} = \frac{N}{2}$, $N_{\hard{}} = \frac{N}{2}$ and the number of unique examples $N_{\simple}^{u} = \frac{N_{\simple{}}}{3}$
\ELSIF{$s \in\{$ {\al{}} $\}$}
    \STATE Initialize the number of data per difficulty $N_{\simple{}} = N$, $N_{\hard{}} = 0$ and the number of unique examples $N_{\simple}^{u} = \frac{N_{\simple{}}}{2}$
\ELSIF{$s \in\{$ {\tw{}} $\}$}
    \STATE Initialize the number of data per difficulty $N_{\simple{}} = \frac{N}{2}$, $N_{\hard{}} = \frac{N}{2}$ and the number of unique examples $N_{\simple}^{u} = N_{\simple{}}$
\ENDIF
 
\STATE Initialize $\mathcal{S} = \Phi$.
\FOR{$t=1 \to N_{\simple}^u$}
\STATE Sample $\data \sim \inputset_{\simple}$.
\STATE If $s \in \{$ {\smfttext}, {\imagetext}, {\imageviatext}, {\mixft}, {\cimageviatext}, {\cmft}, {\al{}}, {\tw{}}  $\}$, then $\mathcal{S} \leftarrow \mathcal{S} \cup (\{\datatext, \reasoningtrace(\data), f(\data)\})$.
\STATE If $s \in \{$ {\smftimage}, {\imagetext}, {\imageviatext}, {\mixft}, {\cimageviatext}, {\cmft}  $\}$, then $\mathcal{S} \leftarrow \mathcal{S} \cup (\{\dataimage, \reasoningtrace(\data), f(\data)\})$.
\STATE If $s \in \{$ {\imageviatext}, {\mixft}, {\cimageviatext}, {\cmft}, {\al{}}  $\}$, then $\mathcal{S} \leftarrow \mathcal{S} \cup (\{\dataimage, P_{convert}, \datatext, \reasoningtrace(\data), f(\data)\})$.
\ENDFOR

\STATE Determine number of epochs to repeat $e = \frac{N_{\simple{}}}{\left|{\mathcal{S}}\right|}$
\STATE Randomly shuffle $\mathcal{S}$ and repeat it $e$ times (i.e., take the first $e \cdot \left|\mathcal{S}\right|$ elements from repeated copies of $\mathcal{S}$)

\FOR{$t=1 \to N_{\hard}$}
\STATE Sample $\data \sim \inputset_{\hard}$.
\STATE $\mathcal{S} \leftarrow \mathcal{S} \cup (\{\datatext, \reasoningtrace(\data), f(\data)\})$.
\ENDFOR
\STATE Randomly shuffle $\mathcal{S}$ and return $\mathcal{S}$.
\end{algorithmic}
\end{algorithm}
\caption{\textbf{Pseudo-code for generating data mixture:} For ablation studies, the algorithm might be slightly modified.}
\end{figure}

In \cref{alg:data_generation}, we provide the pseudo-code for generating the training data mixture for the main experiments. Below we provide more details in the setup. 

\subsubsection{When training only on {\simple} examples}

For {\consecutivetableread} and for any type of supervision among {\smfttext}, {\smftimage}, {\imagetext{}}, {\imageviatext{}}, and {\mixft}:

\begin{itemize}
    \item For each unique data $\data \in \inputset$ and for each type of supervision --- {\smfttext}, {\smftimage}, and {\imageviatext}, we choose whether to include it in the training data, depending on whether these types of supervision are used for training (\cref{sec:sup_types}). We denote the number of unique data $\data$ used from $\inputset_{\simple}$ as $N_{\simple}^u$.
    
    \item We compare all training strategies with the total number of training data used, given by:
    \begin{align*}
        N_{\simple{}} = \text{Number of epochs} \times N_{\simple}^u \times \text{Number of types of supervision per input}
    \end{align*}
\end{itemize}

For a fair comparison, we keep the number of unique data $N_{\simple}^u$ fixed across {\imagetext{}}, {\imageviatext{}}, and {\mixft}. Then to match $N_{\simple{}}$, we set the number of epochs to $1.5$ for {\imagetext{}} ($50\%$ samples are repeated $2\times$), $3$ for {\imageviatext{}}, and $1$ for {\mixft}. 

\textbf{Note on {\smfttext} and {\smftimage} for {\consecutivetableread}:} Since our result depends heavily on the success of {\smfttext} and the failure of {\smftimage} in {\consecutivetableread}, we carefully tune the number of training epochs to achieve optimal performance. We conduct ablations where instead of setting $N_{\simple{}}^u = N_{\simple{}}$, we also try setting $N_{\simple{}}^u$ equal to $\frac{N_{\simple{}}}{2}$ or $\frac{N_{\simple{}}}{3}$ (respectively, the number of epochs is set at $2, 3$). The results presented in \cref{fig:exact_match_consecutiveread} corresponds to $N_{\simple{}}^u = \frac{N_{\simple{}}}{2}$ for {\smfttext} and $N_{\simple{}}^u = \frac{N_{\simple{}}}{3}$ for {\smftimage}. We discuss further in \cref{sec:cmp_atequalTsimp}.

\subsubsection{When also training on {\hard} examples}

For {\cimageviatext} or {\cmft} on {\nontextgeneralizable} tasks:
\begin{itemize}
    \item We set $N_{\hard}$, the number of data from the {\hard} task, equal to $N_{\simple}$, the number of data from the {\simple} task. 
    \item We generate a mixture of $N_{\simple}$ examples under {\imageviatext{}} or {\mixft}. We include $N_{\hard}$ instances of {\hard} {\smfttext}.
\end{itemize}

\subsubsection{Reasoning alignment (\al{}) or text warm-up pretraining (\tw{})}
When generating data for the reasoning alignment phase (\al{}):
\begin{itemize}
    \item We set $N = 10^4$ and include an equal number of \simple{} {\smfttext} and \simple{} {\imageviatext} examples.
\end{itemize}
When generating data for the text warm-up pretraining phase (\tw{}):
\begin{itemize}
    \item We set $N = 10^4$ and include an equal number of \simple{} {\smfttext} and \hard{} {\smfttext} examples.
\end{itemize}
After training on \tw{} and/or \al{}, we continue with the main phase of supervision (e.g., {\cmft} for {\alcmft}).

\subsection{{\consecutivetableread}}
\label{sec:ctr_app}

Given a table with $n_r$ rows and $n_c$ columns, a start cell $(r_s, c_s)$ and an end cell $(r_e, c_e)$, the model is tasked to read all numbers between the start cell and end cell following the given rules.
\label{sec:warmup_data_details_consecutiveread}
\begin{itemize}[leftmargin=*]

    \item If $r_s < r_e$, move left-to-right within each row: 
    \begin{equation*}
        (r_s, c_s), (r_s, c_s + 1), \cdots, (r_s, n_c), (r_s + 1, 1), (r_s +1, 2), \cdots, (r_e, 1), (r_e, 2), \cdots, (r_e, c_e)
    \end{equation*}
    \item  If $r_s > r_e$, move right-to-left within each row: 
    \begin{equation*}
        (r_s, c_s), (r_s, c_s - 1), \cdots, (r_s, 1), (r_s - 1, n_c), (r_s - 1, n_c - 1), \cdots, (r_e, n_c), (r_e, n_c - 1), \cdots, (r_e, c_e)
    \end{equation*}
    \item If $r_s = r_e$, move from $(r_s, c_s)$ to $(r_e, c_e)$.
\end{itemize}

See example images in \cref{fig:continual_table_readout_example_images}.

\subsection{{\tableread}}
\label{sec:tr_app}

Given a table with $n_r$ rows and $n_c$ columns (where $n_r, n_c \in [8, 12]$), a start cell $(r_s, c_s)$, an end cell $(r_e, c_e)$, and a path of cells $P$ connecting the two cells (without any loops), the task is to read the numbers on the path starting from the start cell and ending at the end cell. Each path is continuous and is a concatenation of linear segments, where consecutive segments are separated by 90 degree turns. On the {\simple} task, each path contains $1-4$ linear segments, following a spiral or sinusoidal pattern, and has an average length of $12$. On the {\hard} task, each path contains $>4$ linear segments, following a compositional spiral or sinusoidal pattern, and has an average length of $35$. See example images in \cref{fig:table-id,fig:table-ood} and an example pseudo-code to create the spiral or sinusoidal patterns in \cref{alg:spiral_path,alg:sinusoidal_path}.

\begin{figure}[ht]
\centering
\begin{minipage}{0.48\textwidth}
    \begin{algorithm}[H]
    \small
    \caption{Spiral Path Generation that changes  directions as right$\to$down$\to$left$\to$up$\to$right$\to\cdots$}\label{alg:spiral_path}
    \begin{algorithmic} 
     \REQUIRE Table with $n_r$ rows and $n_c$ columns, start cell, $k$ linear segments
     \STATE \textbullet~Initial $n_{seg} = 0$
    \STATE \textbullet~Initialize current-cell coordinates as start cell coordinates
    \STATE \textbullet~Initialize current-direction to ``right''
    \STATE \textbullet~Initialize Path$=\Phi$.
    \STATE \textbullet~Direction-Change $=\{\text{``right''}: \text{``down''}, \text{``down''}: \text{``left''}, \text{``left''}: \text{``up''}, \text{``up''}: \text{``right''}\}$
    \STATE \textbullet~Coordinate-Update $=\{\text{``right''}: (0, 1), \text{``down''}: (1, 0), \text{``left''}: (0, -1), \text{``up''}: (-1, 0)\}$
    \WHILE{$n_{seg} \ne k$}
    \STATE \textbullet~Add current cell to Path.
    \STATE \textbullet~Compute temporary-cell by adding coordinate update vector for current-direction from Coordinate-Update to current-cell.
    \STATE \textbullet~If temporary-cell is out of bounds, update current-direction using Direction-Change and increment $n_{seg}$.
    \STATE \textbullet~Update current-cell by adding coordinate update vector for current-direction from Coordinate-Update to current-cell.
    \ENDWHILE
    \STATE Return Path
    \end{algorithmic}
    \end{algorithm}
\end{minipage}\hfill
\begin{minipage}{0.48\textwidth}
    \begin{algorithm}[H]
    \small
    \caption{Sinusoidal Path Generation that changes directions as right$\to$down$\to$left$\to$up$\to$right$\to\cdots$, where down and up movements contain only $2$ cells}\label{alg:sinusoidal_path}
    \begin{algorithmic} 
     \REQUIRE Table with $n_r$ rows and $n_c$ columns, start cell, $k$ linear segments
     \STATE \textbullet~Initial $n_{seg} = 0$
    \STATE \textbullet~Initialize current-cell coordinates as start cell coordinates
    \STATE \textbullet~Initialize current-direction to ``right''
    \STATE \textbullet~Initialize Path$=\Phi$.
    \STATE \textbullet~Direction-Change $=\{\text{``right''}: \text{``left''}, \text{``left''}: \text{``right''}\}$
    \STATE \textbullet~Coordinate-Update $=\{\text{``right''}: (0, 1),  \text{``left''}: (0, -1)\}$
    \WHILE{$n_{seg} \ne k$}
    \STATE \textbullet~Add current cell to Path.
    \STATE \textbullet~Compute temporary-cell by adding coordinate update vector for current-direction from Coordinate-Update to current-cell.
    \IF {temporary-cell is out of bounds}
    \STATE \textbullet~If $n_{seg}=k-1$, break
    \STATE Loop twice
    \STATE \textbullet~$\quad\quad$ Increment column coordinate by 1 in current-cell
    \STATE \textbullet~$\quad\quad$ Add current-cell to Path.
    \STATE \textbullet~Update current-direction using Direction-Change 
    \STATE \textbullet~Increment $n_{seg}$ by $2$.
    \ENDIF
    \STATE \textbullet~Update current-cell by adding coordinate update vector for current-direction from Coordinate-Update to current-cell.
    \ENDWHILE
    \STATE \textbullet~Return Path
    \end{algorithmic}
    \end{algorithm}
\end{minipage}
\caption{\textbf{Pseudo-code for generating spiral and sinusoidal paths on {\tableread}:} For simplicity, we present a single variant of each pattern. By permuting the Direction-Change map, the presented variants can be modified to include other direction patterns.}
\end{figure}

\subsection{{\gridnav}}
\label{sec:gn_app}
Given a grid with $n_r$ rows and $n_c$ columns (where $n_r, n_c \in [8, 12]$), a start cell $(r_s, c_s)$, an end cell $(r_e, c_e)$, and a set of objects and obstacles placed at various positions within the grid, the task is to find a path from the start cell to the end cell that collects all specified objects while avoiding all obstacles. 

For each generated grid, we randomly select several objects from a set of $30$ possibilities: heart, crown, flag, star, flower, umbrella, plane, phone, spark, diamond, queen, hammer, club, gear, arrow, sun, bishop, note, coffee, anchor, cloud, pawn, castle, horse, infinity, moon, null, approx, integral, product, and sum. Each chosen object is represented as an Unicode character, as shown in \cref{fig:unicode_grid}. Obstacles are chosen from the following five symbols: dot, cross, square, triangle, and plus. The names and representations of all these symbols—both objects and obstacles—have been verified using GPT-4o.

\begin{figure}[t]
    \centering
    \includegraphics[width=0.4\linewidth]{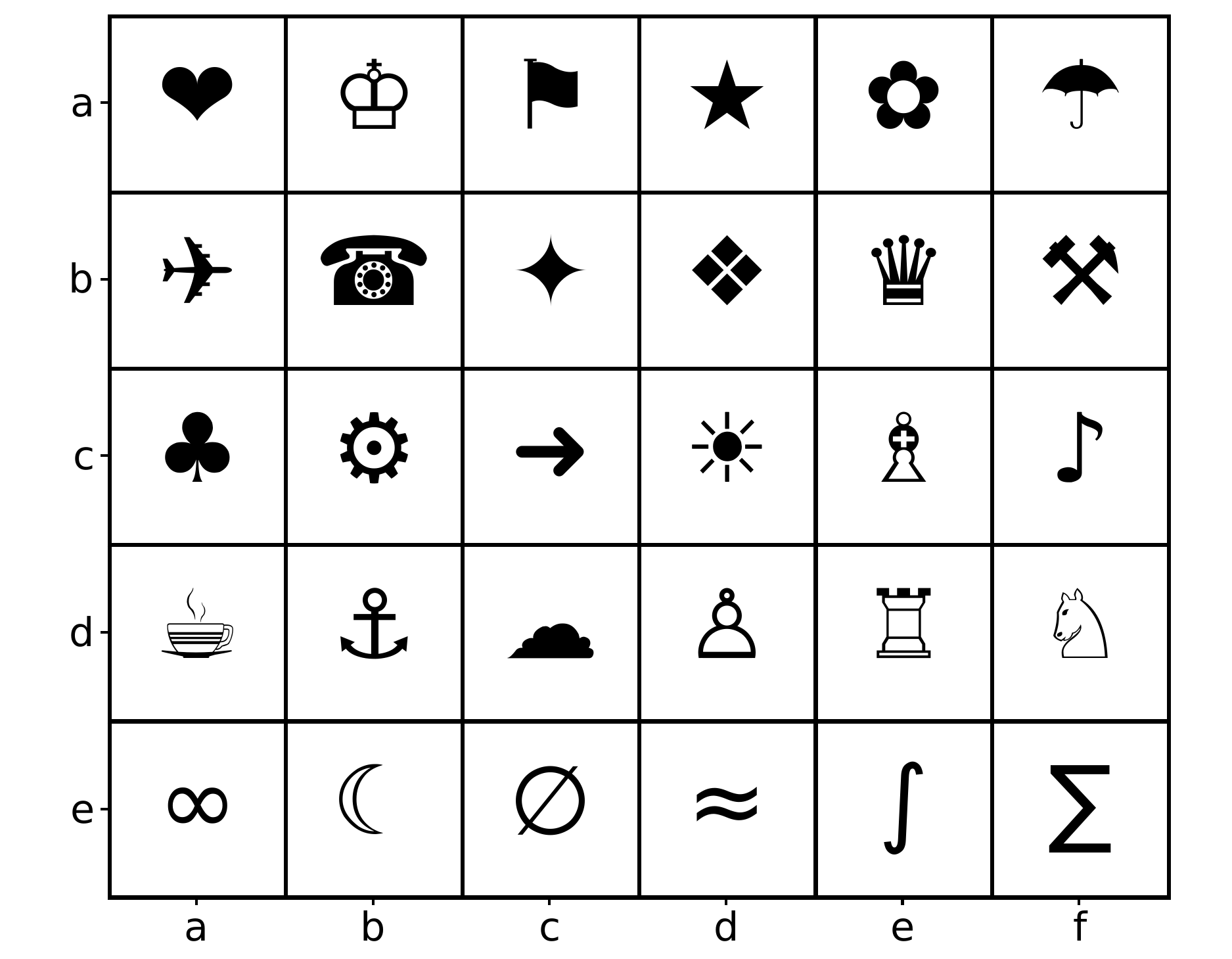}
    \vspace{-3mm}
    \caption{\textbf{Details on {\gridnav}:} Unicode characters used for specifying each object.}
    \label{fig:unicode_grid}
\end{figure}

The {\simple} task requires the model to collect $k\in[1,2]$ objects spread across the grid, while avoiding a single kind of obstacle. The {\hard} task requires the model to collect $k\in[2, 5]$ objects spread across the grid, while avoiding a composition of $o\in[3, 5]$ obstacles. The {\simple} task requires $t\in[10,25]$ DFS steps, while the {\hard} task requires $t\in [25,60]$ DFS steps. 

See example images in \cref{fig:grid-id,fig:grid-ood}.

\subsection{{\visualanalogy}}
\label{sec:ar_app}

We create a multimodal visual analogy dataset based on the Procedurally Generated Matrices (PGM) data proposed in \citet{barrett2018measuring} and \citet{hill2019learning}. Each instance consists of $2$ examples of three images, a query of two images, and four answer options. Each instance has a latent logical relation $r \in \{\texttt{XOR},\texttt{OR},\texttt{AND},\texttt{Progression}\}$ that will be applied to both the examples and the query. There are also three latent domains $d_1, d_2, d_\text{query}$ (for each example and the query, respectively), chosen from $\{\texttt{line\_type}$, $\texttt{line\_color}$, $\texttt{shape type}$, $\texttt{shape\_color}$, $\texttt{shape\_size}$, $\texttt{shape\_quantity}$, $\texttt{shape\_position}\}$. For each example $i$, the value of the domain $d_i$ in the third image follows from applying the relation $r$ to the values in the first two images. The task is to choose one of the four options so that there exists a domain $d_\text{query}$ where applying the relation $r$ along $d_\text{query}$ in the first two images of the query leads to the chosen option. 

Note that following \citet{hill2019learning}, we exclude all spurious correlations of the examples and query such that they follow exactly one pattern $(d,r)$. Furthermore, we create three nontrivial confounding options such that each of them, when combined with the query images, is consistent with exactly one pattern $(d_\text{option$_i$},r_\text{option$_i$})$ where $r_\text{option$_i$}\ne r_\text{query}$.

We also reserve a held-out set of combinations $\mathcal{S}=\{(d, r)\}$ that does not appear in the training images. On the \simple{} task, $d_1 = d_2 = d_\text{query}$ and the query pattern $(d_\text{query}, r_\text{query})$ is never chosen from the held-out set. On the \hard{} task, $d_1, d_2, d_\text{query}$ are distinct and both $(d_i, r_i)$ and $(d_\text{query}, r_\text{query})$ are always chosen from the held-out set $\mathcal{S}$. 

See example images in \cref{fig:ar-domain-transfer-heldout-id,fig:ar-domain-transfer-heldout-ood} and the complete list of all possible attribute values in \cref{tab:ar_values}.

\begin{table}[!t]
    \caption{\textbf{List of all possible attribute values for each domain in {\visualanalogy}:}, We reproduce \citet{hill2019learning} with slight modifications. The diverse combination of the attribute values results in high complexity of this task, testing various both OOD and compositional generalizability of the model to a great extent.}
    \label{tab:ar_values}
    \centering
    \begin{tabular}{l || c}
        \toprule
         \multirow{3}{*}{\texttt{line type}} &  $\{\text{falling diagonal line, rising diagonal line, horizontal line, vertical line,}$ \\ & diamond lines, circular line, V-shape facing up, V-shape facing left \\ & $\text{V-shape facing down, V-shape facing right}\}$ \\
         \hline 
         \texttt{line color} & $\{0~\text{(black)}, 90~\text{(dark grey)}, 135~\text{(grey)}, 189~\text{(light grey)}\}$ \\
         \hline 
         \texttt{shape type} & $\{\text{circle, rectangle, triangle, pentagon, hexagon}\}$ \\
         \hline
         \texttt{shape color} & $\{0~\text{(black)}, 90~\text{(dark grey)}, 135~\text{(grey)}, 189~\text{(light grey)}, 255~\text{(white)}\}$ \\
         \hline
         \texttt{shape size} & $\{20, 27, 34, 41\}$ \\
         \hline
         \texttt{shape quantity} & $\{0,1,2,3,4,5,6,7,8,9\}$ \\
         \hline
         \texttt{shape position} & $\{(0,0), (0,1), (0,2), (1,0), (1,1), (1,2), (2,0), (2,1), (2,2)\}$ \\
        \bottomrule
    \end{tabular}
\end{table}

\subsection{Issues during synthetic data creation}
\label{sec:gen-issue}
Here, we outline the primary issues that we faced while creating the synthetic datasets, which might be of value to the general community.

\subsubsection{\consecutivetableread{}, \tableread{}}
The primary issue that we faced during creation of these datasets were as follows:
\begin{itemize}
    \item \textbf{Resolution issues:} For images, we found that representing numbers as their English names (e.g. 9 represented as NINE) improved the OCR performance substantially. When represented as numerics, the model often confused between pairs (7, 9), and (0, 8). These issues were largely mitigated by replacing numerics with English names.
    \item \textbf{Color:} The model's S2H generalization can vary drastically depending on the color used to highlight the cells. On \hard{} (15-20) images, the performance of the model trained with \smftimage{} supervision can vary from $30\%$ to $70\%$ depending on which color (e.g. {\em purple} or {\em yellow}) was used.
    \item \textbf{CoT Trace:} Our original CoT Trace simply outlined the numbers on the path, without any mention of the row number, column number, row name, and column name of the cells in the highlighted path. This resulted in poor performance of the model when trained with images. We then switched to a more verbose CoT, where the model was provided with the above details at each step of traversing the highlighted path, and the model's performance substantially improved.
\end{itemize}

\textbf{For \consecutivetableread{}, we find that the verbose CoT trace shows S2H generalization, and not the final solution that the model reports.} Hence, we report our evaluation performance for \consecutivetableread{} on the CoT trace.

\subsubsection{\gridnav{}}
The major issue that we faced in \gridnav{} was the design of chain-of-thought reasoning steps to represent the Depth First Search trace. At multiple points, we found that current VLMs are fragile to read image inputs, and our CoT trace needed to be very explicit to train the model effectively on \simple{} examples.

\paragraph{An initial version of \gridnav{}:} In our first version, we designed an extremely simple dataset, where the grids only had a source cell, a destination cell, and a few cells marked by {\em red} color that represented {\em obstacles}. 

\begin{itemize}
    \item \textbf{Models failed to train on image-input without verbose details in CoT:} Our initial CoT would only provide the following at each DFS step: "[current cell]: [proposed next action]" without iterating through all invalid actions considered before proposing this action. e.g., a 3-step DFS step would look as follows:
    \begin{itemize}
        \item (1, 1): right
        \item (1, 2): down
        \item (2, 2): backtrack        
    \end{itemize}
where we don't explain why we need to "backtrack" at (2, 2). This made the model learn the following:
\begin{enumerate}
    \item answer formatting
    \item knowing how to retrieve the current location (row, col index) and the destination location
    \item knowing which action is preferred (the one that minimizes the distance towards destination)
\end{enumerate}
but the model never picked up on why we sometimes backtrack or sometimes take an action that is not the most preferred. At generation, it would ignore all obstacles and try to take the most preferred action.

\textbf{On the other hand, we observed that the model could still recognize the reasoning for ``backtracking'' on text input and could get $100\%$ accuracy on \simple{} text for \smfttext{} supervision, and also $100\%$ accuracy on \simple{} images for \mixft{} supervision. } Thus, for cases where the model couldn't train with image-input but could train with text input, \mixft{} was useful to train the model even for improving accuracy on in-domain examples. However, this setting was slightly different from our {\simpletohard} view, and so we decided to make the CoT more verbose.

\item \textbf{In later attempts, we switched to a more verbose CoT:} We iterate through all possible actions at each state, giving reasons why that action is valid / invalid. e.g. a 3-step DFS trace that starts from cell (1, 1) will look as follows
\begin{itemize}
    \item Current cell: (1, 1): right would lead to (1, 2) which is available and not visited yet, so we can move right.
    \item Current cell: (1, 2): down would lead to (2, 2) which is available and not visited yet, so we can move down
    \item Current cell: (2, 2): right would lead to (2, 3) but it has an obstacle; down would lead to (3, 2) but it has an obstacle; left would lead to (2, 1) but it has an obstacle; we have no more action left, so backtrack
\end{itemize}
The model now gets almost perfect {\simpletohard} on both text / image no matter which supervision we give. So we couldn't really compare the performance of different types of supervision. This was because once the model learns how to iterate through different actions and determine its validity, length generalization was trivial.

\end{itemize}

Thus, we switched to our current version of \gridnav{}, where the task additionally involved spatial reasoning of different combinations of objects and obstacles spread across the grid.

\subsubsection{\visualanalogy{}}
Here, the main challenge is to recreate the Procedurally Generated Matrices (PGM) dataset first introduced in \citet{hill2019learning} and \citet{barrett2018measuring}, as the data generation code is not publicly available. Therefore, we try our best to recreate the data set with slight adaptations. Specifically, we have $10$ variations in the attribute values for \texttt{line\_type}, \texttt{shape\_quantity}, and \texttt{shape\_position} as in the original paper. For the rest of attributes \texttt{line\_color}, \texttt{shape\_type}, and \texttt{shape\_size}, we only include $\le 5$ variations of attribute values. Meanwhile, as the original papers do not list all the attribute values used in the original data generation, nor was the source code publicly available, we decide upon the list of possible attribute values based on the consideration that they are clearly differentiable from a human perspective.

The original papers claim that solving PGM puzzles is a challenging vision task. While we acknowledge that our recreated version of the data reduces the complexity compared to the original version, we note that our adaptations do not qualitatively change the challenging nature of this task. As mentioned in \citet{barrett2018measuring}, the challenge of effective knowledge composition comes mainly from the necessity to represent abstract logical rules in discrete symbolic explanations. They show that training with auxiliary information of {\em meta-targets} vectors that encode the relation, object, and attribute types as a binary string significantly helps abstract reasoning performance, and in particular, in terms of compositional generalization. Our text representations are inspired by the construction of the {\em meta-targets} vectors with many tweaks to fit into the context length of the model. We observe that by including the discrete representation of knowledge in the form of {\imageviatext} supervision, {\mixft} and {\cmft} show a much better S2H generalization on image input, which aligns with previous observations in \citet{barrett2018measuring}.

\clearpage
\section{Training Details}
\label{sec:training_details}

We first prepare \textit{Eagle-X2-Llama3-8B}, a variation of \textit{Eagle-X5-8B} \citep{shi2024EAGLE}. We choose Llama3-8B-Instruct \citep{dubey2024llama} as the LLM backbone for its good reasoning capability. We choose CLIP-448 \citep{radford2021learning} and ConvNeXt \citep{liu2022convnet} as the visual encoders because previous works show that combining the two leads to a significant improvement, whereas any additional visual encoder leads to marginal improvement \citep{shi2024EAGLE}. 

At the beginning of the project, the codebase released by \citet{shi2024EAGLE} was incomplete. To incorporate the Llama3-8B model architecture and the tokenizer, we adapt the codebase from \citet{tong2024cambrian1}.

We use the same 595k pretraining data from \citet{liu2023LLaVA1.5} and 1.8M finetuning (visual instruction tuning) data from \citet{shi2024EAGLE}. We use Deepspeed ZeRO Stage 2 \citep{rasley2020deepspeed} for a Distributed Data Parallel (DDP) training on 8 GPUs on a HPC Cluster. We use the AdamW optimizer with no weight decay (i.e., equivalent to Adam), a learning rate schedule with a linear warmup of 0.03 and cosine decay to zero. We truncate the trail of any text that exceeds the maximum number of text tokens (2048). During pretraining, only the adapter is trained, whereas in all other stages of training, all weights in the model are unfrozen.

With this \textit{Eagle-X2-Llama3-8B} as the base model, we then continuously finetune it on different data mixtures across our synthetic tasks. In \cref{tab:training_details}, we report some key hyperparameters. 
 
\begin{table}[!h]
    \caption{\textbf{Hyperparameter settings:} For all values not reported here, we use the same values as in \citet{shi2024EAGLE}. }
    \label{tab:training_details}
    \centering
    \begin{tabular}{l | c | c | c | c | c }
        \toprule
          & Batch Size & LR & Epochs & Total \# Data & Max \# Text Tokens \\
        \midrule
        Pretraining & 256 & 1e-3 & 1 & 595k & 2048 \\
        \midrule
        Finetuning & 128 & 2e-5 & 1 & 1,809k & 2048 \\
        \midrule
        Finetuning on Task & 128 & 2e-5 & \multicolumn{2}{|c|}{experiment-specific} & 2048 \\
        \bottomrule
    \end{tabular}
\end{table}

\clearpage
\section{Evaluation Details}
\label{sec:evaluation_details}

We extend the VLMEvalKit \citep{duan2024vlmevalkit} to evaluate the finetuned \textit{Eagle-X2-Llama3-8B} on held-out data. For generation, we apply greedy decoding and generate up to 2048 tokens. 

\subsection{Evaluation on {\consecutivetableread} and {\tableread}}
$\reasoningtrace(\data)$ visits each cell sequentially on the path, by giving the row and column index, row and column names, and the value in the cell (see \cref{fig:table-id,fig:table-ood}). The final answer $\reasontask(\data)$ gives the list of numbers again, and also sum of the numbers. We evaluate by simply checking whether the list of numbers are correct. Furthermore, because this list of numbers can be extracted from both the final answer and also the CoT, we report the best performance out of the two. On {\consecutivetableread}, we find that we get the best performance on {\hard} examples by extracting the numbers from CoT. On the other hand, for {\tableread}, there isn't much difference between extracting numbers from CoT and extracting them from the final answer.

\subsection{Evaluation on {\gridnav}}
$\reasoningtrace(\data)$ records the sequence of visited cells during a depth-first search (DFS) from the start to the end cell. At each visited cell, the trace includes a full description of neighboring cells and whether they are available for the next step. The DFS algorithm always prefers directions that minimize the distance towards the nearest uncollected object, or the destination (if all objects are collected). If no directions are possible, we backtrack to the most previously visited cell. The final answer $\reasontask(\data)$ is a simplified sequence of directions (left, right, up, down) that connect the start and destination cells, where all backtrack movements are removed from the stack (see \cref{fig:grid-id,fig:grid-ood}). We evaluate by simulating the movements in the sequence returned by the model and checking if we arrive at the destination after collecting all objects and avoiding obstacles.

\subsection{Evaluation on {\visualanalogy}}
\label{app:eval_visual_analogy}
$\reasoningtrace(\data)$ enumerates all the values of the tasks-relevant attributes for each panel with the conclusion of whether there exists a logical pattern among those values for each attribute domain in the examples. The trace includes a summary sentence of what (domain, relation) pattern the two examples demonstrate. After that, the trace performs the same enumeration process for the query panels. It then looks at the options and checks whether it is consistent with the desired relation given the attribute values in the query panels. The final answer $\reasontask(\data)$ identifies the pattern in the form of (domain, relation) (e.g. \texttt{(line type, XOR)}) for all examples and the query combined with each option, as well as the final answer of the correct option. The evaluation checks whether the identified patterns and the final answer are correct.
\clearpage
\section{Consistent Results on Another Model Family and Size}
\label{app:qwen}

\subsection{Training Details}
We take \textit{Qwen2.5-VL-3B-Instruct} and \textit{7B-Instruct} \citep{bai2025qwen25vltechnicalreport} as the base model and finetune it with on different data mixtures across our synthetic tasks.

We use the \texttt{SFTTrainer} class in the \texttt{trl} package. We employ FSDP \citep{zhao2023FSDP} for training on 8 GPUs on a HPC Cluster. We use the AdamW optimizer with no weight decay (i.e., equivalent to Adam), a learning rate schedule with a linear warmup of 0.03 and cosine decay to zero.

Due to a deficiency in the \texttt{trl} package, we slightly modify \cref{alg:data_generation} to ensure that each gradient computation (before gradient accumulation) includes a training example with $\dataimage$ unless we train exclusively on $\datatext$ (while also maintaining randomness in the data). Instead of concatenating and randomly shuffling the entire dataset, we first shuffle the examples within each supervision, then interleave the individually shuffled data. For example, to construct {\cmft}, we first shuffle {\smfttext}, {\smftimage}, {\imageviatext}, and {\hard}-{\smfttext} individually, then construct the final dataset by repeatedly taking the next examples from ({\smfttext}, {\smftimage}, {\imageviatext}, {\hard}-{\smfttext}, {\hard}-{\smfttext}, {\hard}-{\smfttext}) respectively.

In \cref{tab:qwen_training_details}, we report some key hyperparameters. Note that \citet{bai2025qwen25vltechnicalreport} do not report the hyperparameters for their internal training, so we used the hyperparameters for \textit{Eagle-X2-Llama-8B} as closely as possible. However, we noticed that for \textit{Qwen2.5-VL-7B}, training on {\gridnav} or {\visualanalogy} with a learning rate of 2e-5 often broke the model (e.g., model starts outputting Chinese tokens), so we had to adjust the learning rate to 5e-6 or 2e-6. 
 
\begin{table}[!h]
    \caption{\textbf{Hyperparameter settings for Qwen2.5-VL.}}
    \label{tab:qwen_training_details}
    \centering
    \begin{tabular}{c | c | c | c | c | c }
        \toprule
        Model Size & Task & Batch Size & LR & Epochs & Total \# Data \\
        \midrule
        3B & All & 128 & 2e-5 & \multicolumn{2}{|c}{experiment-specific} \\
        \midrule
        7B & {\consecutivetableread} & 128 & 2e-5 & \multicolumn{2}{|c}{experiment-specific} \\
        7B & {\tableread} & 128 & 2e-5 & \multicolumn{2}{|c}{experiment-specific} \\
        7B & {\gridnav} & 128 & 2e-6, 5e-6 & \multicolumn{2}{|c}{experiment-specific} \\
        7B & {\visualanalogy} & 128 & 2e-6, 5e-6 & \multicolumn{2}{|c}{experiment-specific} \\
        \bottomrule
    \end{tabular}
\end{table}

\subsection{Evaluation Details}
We extend the VLMEvalKit \citep{duan2024vlmevalkit} to evaluate the finetuned \textit{Qwen2.5-VL-3B-Instruct} and \textit{7B-Instruct} on the same held-out data as used for the main experiments. For generation, we apply the default setting for the \textit{Qwen2.5-VL} family (top $p$=0.001 and temperature=0.01) and generate up to 2048 tokens. We set the maximum number of pixels to be $1280 \times 28 \times 28$.

\subsection{Results}
In \cref{tab:qwen_3b_results,tab:qwen_7b_results}, we report the S2H-generalization on image for most supervision types we consider. 

For {\consecutivetableread}, we find that \textit{Qwen2.5-VL} (both 3B and 7B models) completely fail to solve {\hard}-text examples even when {\hard} {\smfttext} is a part of the training data (e.g., {\cmft}). For this reason, we relax the definition of hardness and instead train with \textsc{medium}-text (if applicable) and evaluate on \textsc{medium}-text and \textsc{medium}-image (see \cref{sec:consecutive_read_exp} for the definitions of \textsc{medium} and {\hard}). Even then, \textit{Qwen2.5-VL-3B-Instruct} fail to solve \textsc{medium}-text examples even when it is explicitly trained with \textsc{medium} {\smfttext}. Therefore, none of our proposed methods can improve S2H-generalization on image. However, we find that even though \textit{Qwen2.5-VL-7B-Instruct} does not S2H-generalize on text (which is understandable since different models can S2H-generalize on different tasks), our proposed supervision types for non-S2H generalizing tasks ({\cmft} and {\alcmft}) successfully improve the S2H-generalization on image. 

For the other 3 tasks ({\tableread}, {\gridnav}, and {\visualanalogy}), we generally observe a consistent result from the main text: 1) the models do not S2H-generalize on either text or image; 2) {\cmft} improves S2H-generalization on image by transferring the injected reasoning on {\hard}-text; 3) {\alcmft} further improves this generalization. Note that for {\gridnav}, and {\visualanalogy} on \textit{Qwen2.5-VL-7B-Instruct}, we report the best result between the two learning rates (2e-6, 5e-6).

\clearpage
\begin{table}[h!]
    \caption{\textbf{Results for Qwen2.5-VL-3B-Instruct:} For {\smfttext} supervision, we evaluate on {\hard}-text, but for all other supervision types, we evaluate on {\hard}-image. For {\consecutivetableread}, we train with \textsc{medium} {\smfttext} (if applicable) and evaluate on \textsc{medium}-image. }
    \label{tab:qwen_3b_results}
    \centering
    \begin{tabular}{l|r|rr|rr|rr}
        \toprule
        & {\consecutivetableread} & \multicolumn{2}{c|}{{\tableread}} & \multicolumn{2}{c|}{{\gridnav}} & \multicolumn{2}{c}{\visualanalogy} \\
        \textbf{Supervision} & 30k & 30k & 60k & 30k & 60k & 30k & 60k \\
        \midrule
        {\smfttext} (eval on {\hard}-text) & 3 & 8 & 11 & 0 & 15 & 0 & 0 \\
        {\smftimage} & 0 & 11 & 10 & 22 & 22 & 0 & 0 \\
        {\imagetext} & 1 & 7 & 6 & 0 & 14 & 0 & 0 \\
        {\imageviatext} & 1 & 12 & 8 & 13 & 14 & 0 & 0 \\
        {\mixft} & 1 & 11 & 10 & 14 & 16 & 0 & 1 \\
        {\cimageviatext} & 0 & 81 & 90 & 67 & 58 & 48 & 48 \\
        {\cmft} & 4 & 78 & 86 & 77 & 91 & 20 & 27 \\
        {\alcmft} & - & 66 & 91 & 80 & 91 & 38 & 42 \\
        \bottomrule
    \end{tabular}
\end{table}

\begin{table}[h!]
    \caption{\textbf{Results for Qwen2.5-VL-7B-Instruct:} For {\smfttext} supervision, we evaluate on {\hard}-text, but for all other supervision types, we evaluate on {\hard}-image. For {\consecutivetableread}, we train with \textsc{medium} {\smfttext} (if applicable) and evaluate on \textsc{medium}-image. }
    \label{tab:qwen_7b_results}
    \centering
    \begin{tabular}{l|r|rr|rr|rr}
        \toprule
        & {\consecutivetableread} & \multicolumn{2}{c|}{{\tableread}} & \multicolumn{2}{c|}{{\gridnav}} & \multicolumn{2}{c}{\visualanalogy} \\
        \textbf{Supervision} & 30k & 30k & 60k & 30k & 60k & 30k & 60k \\
        \midrule
        {\smfttext} (eval on {\hard}-text) & 1 & 22 & 2 & 15 & 18 & 1 & 0 \\
        {\smftimage} & 0 & 18 & 17 & 14 & 29 & 0 & 0 \\
        {\imagetext} & 4 & 8 & 5 & 6 & 11 & 0 & 0 \\
        {\imageviatext} & 36 & 9 & 13 & 13 & 18 & 0 & 0 \\
        {\mixft} & 52 & 8 & 17 & 15 & 12 & 0 & 0 \\
        {\cimageviatext} & 73 & 82 & 88 & 75 & 67 & 41 & 44 \\
        {\cmft} & 72 & 13 & 66 & 69 & 85 & 12 & 17 \\
        {\alcmft} & - & 93 & 92 & 36 & 58 & 25 & 34 \\
        \bottomrule
    \end{tabular}
\end{table}

\clearpage
\section{Continued Discussion From Main Paper}
\label{sec:contd_discussion}

\subsection{Comparisons at equal unique samples for {\consecutivetableread}}\label{sec:cmp_atequalTsimp}
In \cref{fig:exact_match_consecutiveread}, we compare {\smfttext}, {\smftimage}, and {\mixft} under the same $N_{\simple}$, the total number of training data. Note that {\mixft} is trained for only a single epoch, while the reported results for {\smfttext} and {\smftimage} are based on 2 and 3 epochs of training, respectively. We make these choices because, for {\smfttext}, the {\simpletohard} performance peaks at 2 epochs and then declines sharply, whereas for {\smftimage}, the {\simpletohard} performance sees a slight improvement between 2 and 3 epochs. As an illustrative example, \cref{fig:multi-epoch_training} shows the performance of {\smfttext} and {\smftimage} when $N_{\simple}^u$, the number of unique samples, is fixed at $4 \times 10^4$. Consequently, in \cref{fig:equiflop_compare_CTR_uniquedata}, we revisit the results of \cref{fig:exact_match_consecutiveread}, this time explicitly indicating the number of unique samples $N_{\simple}^u$ used.

\begin{figure*}[!ht]
    \centering
    \includegraphics[width=0.7\linewidth]{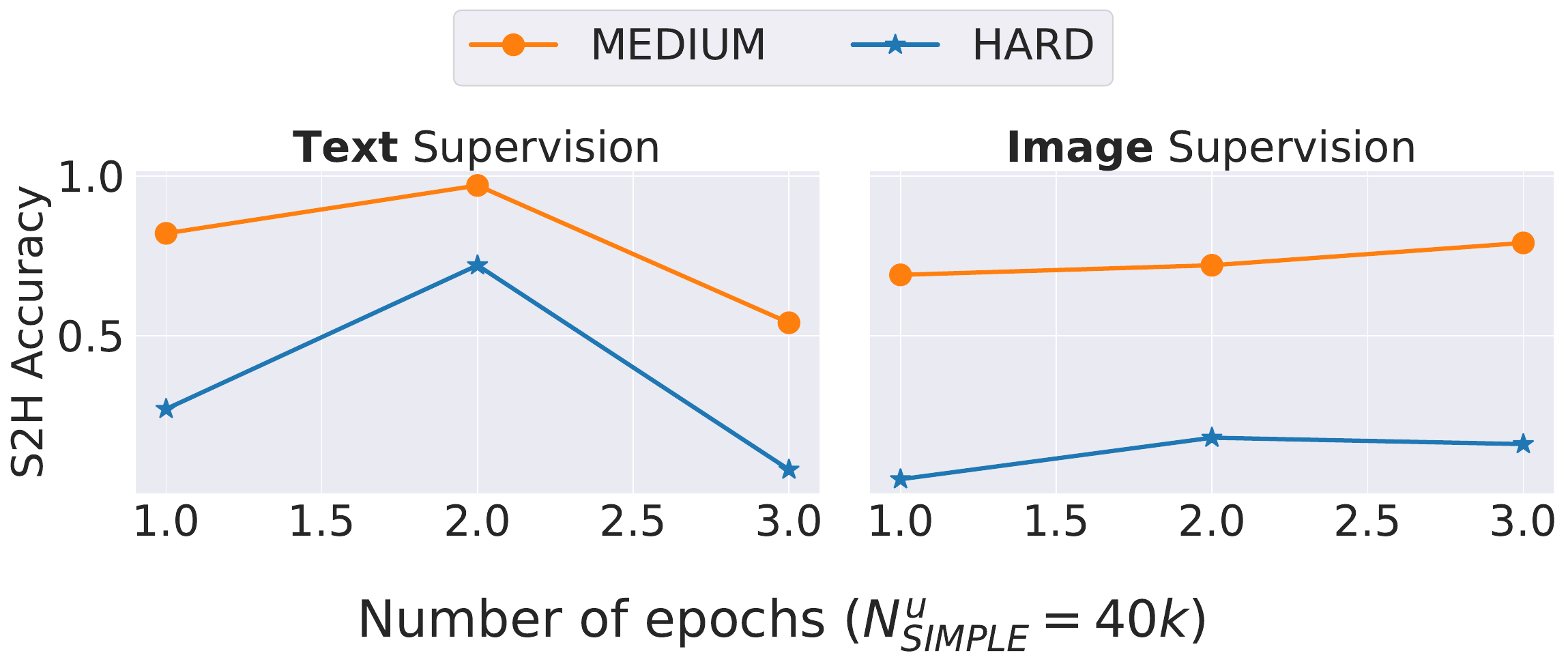}
    \caption{\textbf{Ablation on the number of epochs on {\consecutivetableread}:} We measure the {\simpletohard} performance of {\smfttext} on {\hard}-text and {\smftimage} on {\hard}-image with multi-epoch training, when $N_{\simple}^u$ is fixed as $4 \times 10^4$. We observe that the generalization performance of {\smfttext} supervision peaks at $2$ epoch training, after which it drastically drops, while the generalization performance of {\smftimage} supervision increases slightly between $2$ and $3$ epochs of training.}
    \label{fig:multi-epoch_training}
\end{figure*}
\begin{figure}[t]
    \centering
    \includegraphics[width=0.6\linewidth]{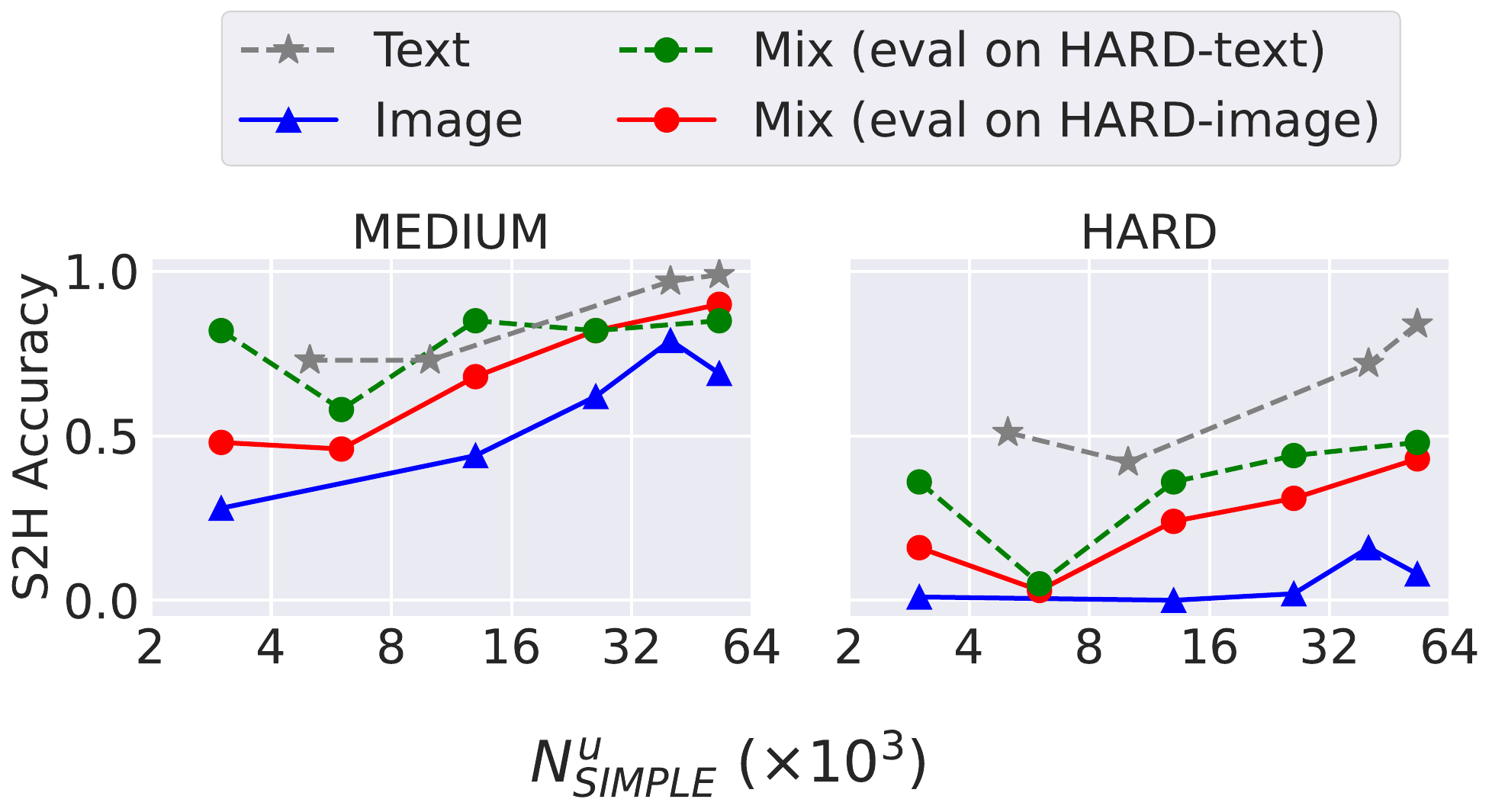}
    \caption{\textbf{Results on {\consecutivetableread} based on the number of unique samples $N_{\simple}^u$:} Our observations from \cref{sec:consecutive_read_exp} hold true even when different types of supervision are compared at the same value $N_{\simple}^u$, instead of  $N_{\simple}$.}
    \label{fig:equiflop_compare_CTR_uniquedata}
\end{figure}

\subsection{Additional setting for an {\textgeneralizable} task} \label{sec:compositional_VA}
Here, we consider {\em Pattern-Heldout Visual Analogy} --- a {\textgeneralizable} version of {\visualanalogy} --- by defining an alternative version of {\hard} examples. We keep the definition of \simple{} examples from \cref{sec:tr_ar_gn,sec:ar_app}, but modify \hard{} instances to only measure analogical reasoning on held-out reasoning patterns, without requiring the domain to be different across the in-context examples. 

That is, let $d_1, d_2, d_\text{query}$ denote the latent domains of the examples and the query, $r$ denote the latent logical relation to be applied on the latent domains, and $\mathcal{S}$ denote a held-out set of combinations $(d, r)$. The \simple{} task contains puzzles where $d_1 = d_2 = d_\text{query}$ and $(d_\text{query}, r_\text{query}) \not \in \mathcal{S}$, whereas the \hard{} task contains puzzles where $d_1 = d_2 = d_\text{query}$ and $(d_\text{query}, r_\text{query}) \in \mathcal{S}$. Note that in {\visualanalogy}, we had additionally required $d_1, d_2, d_\text{query}$ to be distinct for {\hard} puzzles.

In \cref{tab:compositional_ar_results}, we compare the {\simpletohard} performance of \smfttext{}, \smftimage{}, and \mixft{} supervision on {\em Pattern-Heldout Visual Analogy}. The model learns the task more easily on text than on image: while the image {\simpletohard} for \smftimage{} supervision is bounded by $32\%$, the text {\simpletohard} for \smfttext{} supervision can reach $49\%$ when trained on $24 \times 10^4$ data. 

On the other hand, \mixft{} supervision can transfer the {\simpletohard} from text to image and improve the performance on {\hard}-image ($41\%$ with $12 \times 10^4$ training data).

\begin{table}[!ht]
    \caption{\textbf{Results on {\em Pattern-Heldout Visual Analogy}:} {\simpletohard} for \smfttext{}, \smftimage{}, and \mixft{} supervision are reported on {\hard}-text and {\hard}-image examples after varying the number of training data in each strategy. {\simpletohard} on {\hard}-images under \smftimage{} supervision peaks at $36\%$, while for {\hard}-text examples under \smfttext{} supervision, it reaches $45.6\%$ after $24 \times 10^4$ training examples. Leveraging the better performance on {\hard}-text, \mixft{} supervision improves {\simpletohard} on {\hard}-images to $41\%$ with $12 \times 10^4$ examples.}
    \label{tab:compositional_ar_results}
    \centering
    \begin{tabular}{c|cccc|cccc}
        \toprule
        & \multicolumn{4}{c|}{\textbf{S2H accuracy on {\hard}-text}} & \multicolumn{4}{c}{\textbf{S2H accuracy on {\hard}-image}} \\
        \cline{2-9} 
        & \multicolumn{4}{c|}{Number of training data} & \multicolumn{4}{c}{Number of training data} \\
        \textbf{Supervision} & $30$k & $60$k & $120$k & $240$k & $30$k & $60$k & $120$k & $240$k \\
        \midrule
        \smfttext{}   & - & $37.2$ & $32.6$ & $45.6$ & $0.0$ & $0.0$ & $0.0$ & $0.0$ \\
        \smftimage{}  & $0.0$ & $0.0$ & $0.0$ & $0.0$ & - & $27.0$ & $35.6$ & $34.0$ \\
        \mixft{}      & $31.2$ & $42.4$ & $49.0$ & $39.8$ & $24.6$ & $35.0$ & $41.0$ & $39.6$ \\
        \bottomrule
    \end{tabular}
\end{table}

\subsection{Comparison at equal FLOPs for {\nontextgeneralizable} tasks}\label{sec:cmp_atequalT}
{\alcmft} uses an additional phase over {\cmft}, where training sequences from {\simple} split are utilized. In \cref{fig:main-results}, however, we compare {\cmft} and {\alcmft} only in terms of the amount of training data used in the final phase. This raises a potential concern that {\alcmft} might only appear stronger because it involves more total training FLOPs. To address this, \cref{fig:equiflop_compare} presents a revised comparison, plotting {\alcmft} against {\cmft} in terms of the total training data employed across all stages. Under these conditions, {\alcmft} still consistently outperforms {\cmft}.

\begin{figure}[!ht]
    \centering
    \includegraphics[width=\linewidth]{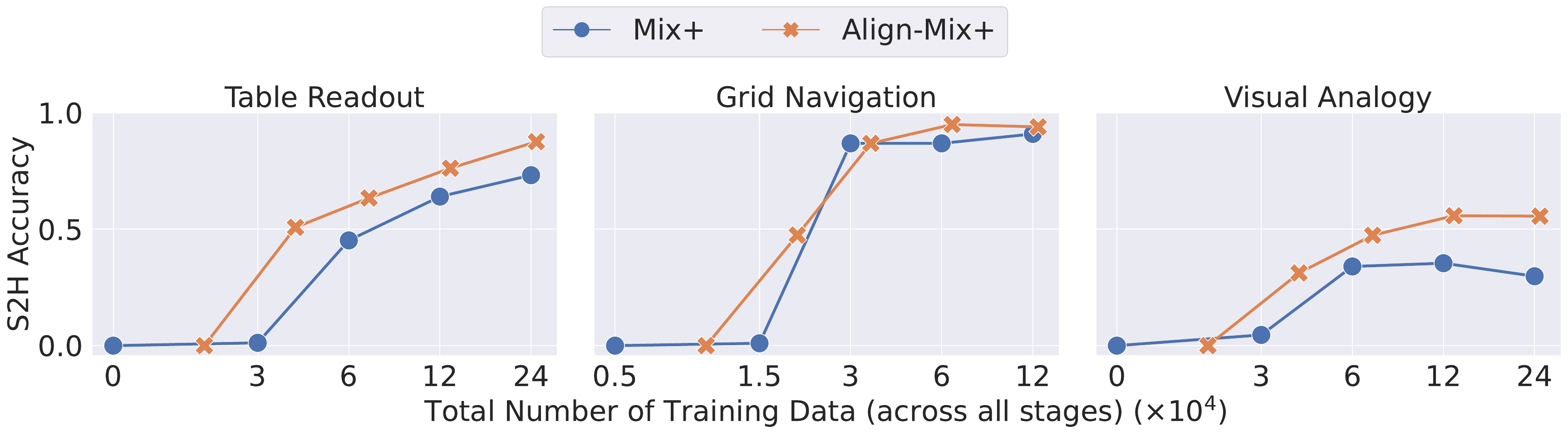}
    \caption{\textbf{Results on {\nontextgeneralizable} tasks based on the total number of training data (\cref{fig:compare_itota_cmft} but including the amount of data in the alignment stage):} {\alcmft} still outperforms {\cmft} when compared at the same amount of total training data.}
    \label{fig:equiflop_compare}
\end{figure}

\subsection{Transferring reasoning from image to text}
\label{sec:symbiote}
In the main experiments, we tested whether {\simpletohard} can transfer from text inputs to image inputs. In \cref{tab:symbiote}, we observe that the transfer can happen in the opposite direction as well. After $24 \times 10^4$ training samples that now includes training data from \hard{} {\smftimage} instead of \hard{} {\smfttext}, a modified version of {\cmft} achieves \simpletohard{} accuracy of $86.0\%$ on {\hard}-text input on {\tableread} and $85.6\%$ on {\hard}-text input on {\visualanalogy}. As a comparison, when trained with the same number of data, {\cmft} shows \simpletohard{} accuracy of $73.2\%$ on {\hard}-image input on {\tableread} and $35.4\%$ on {\hard}-image input on {\visualanalogy}.

\begin{table}[t]
    \caption{\textbf{Ablation on transferring reasoning from image to text:} We modify {\cmft} to include {\hard} {\smftimage} examples in training, instead of {\hard}  {\smfttext} examples, while keeping the same {\simple} {\mixft} supervision. Evaluation is now performed on {\hard}-text input. We observe that improving generalization performance on {\hard}-image input strongly transfers to {\hard}-text input. }
    \label{tab:symbiote}
    \centering
    \begin{tabular}{l|cc|cc}
        \toprule
        & \multicolumn{2}{c}{ {\tableread}} &   \multicolumn{2}{c}{{\visualanalogy}} \\
        \midrule
         Number of  & {\hard}-image & {\hard}-text & {\hard}-image & {\hard}-text \\
         training examples & (Included in training) & (Excluded in training)& (Included in training)& (Excluded in training) \\
         \midrule 
         $3\times 10^4$ & 72.0 & 34.0 & 86.8 & 81.4 \\
         $6\times 10^4$ & 98.2 & 70.4 & 97.8 & 80.0 \\
         $12\times 10^4$ & 99.4 & 76.4 & 94.6 & 86.4 \\
         $24\times 10^4$ & 99.8 & 86.0 & 99.6 & 85.6 \\
         \bottomrule
    \end{tabular}
\end{table}

\begin{figure*}[!t]
    \centering
    \begin{subfigure}[t]{0.98\linewidth}
        \centering
        \includegraphics[width=1.\linewidth]{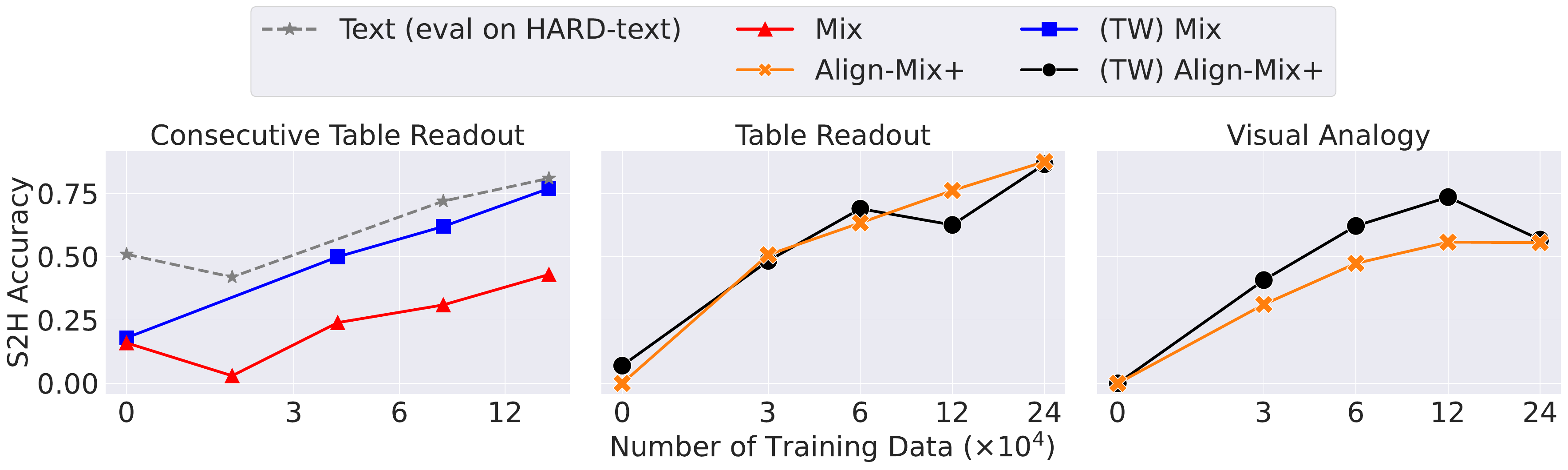}
        \end{subfigure}
    \caption{\textbf{Effect of text warm-up pretraining:} We report the S2H generalization on image with/without text warm-up before {\cmft} ({\consecutivetableread}) or {\alcmft} ({\tableread} and {\visualanalogy}). S2H generalization on text from {\smfttext} supervision serves as a reference (in gray dashed line). Text warm-up enhances image {\simpletohard} across tasks. {\contmixft} closes the text-image generalization gap on the {\hard} task for {\consecutivetableread}, while {\contalcmft} outperforms {\alcmft} for {\visualanalogy}.}
    \label{fig:ct_cmp_loss}
\end{figure*}

\subsection{Text warm-up pretraining}
\label{app:ablations_continual_pretraining}
In \cref{sec:tr_ar_gn}, we observe that {\mixft} fails to improve image generalization when the LLM backbone does not show strong generalization on text modality. Furthermore, models trained with {\cmft} show significantly better image generalization when the text reasoning capability of the LLM backbone is strengthened by fine-tuning on {\hard} {\smfttext} examples. Hence, one may expect that the reasoning capability of the LLM backbone on \hard{} examples is a crucial factor for the reasoning capability to transfer to the image inputs. In this section, we investigate the effect of text generalization of the LLM backbone. Specifically, we simulate different levels of text reasoning ability by including a pretraining stage of the model on {\simple} and {\hard} {\smfttext} examples. We call this text-only training (during which only the LLM backbone is updated) before full finetuning of the VLM model \emph{text warm-up pretraining} \tw{}. This stage of training only uses a small set of $10^4$ text examples (equal mix of {\simple} and {\hard} for non S2H-generalizable tasks, just {\simple} examples for \consecutivetableread{}). Our results are shown in \cref{fig:ct_cmp_loss}.

We observe that the additional TW training further boosts the image generalization. In particular, {\contmixft} closes the modality imbalance, reflected by the image-text generalization gap on the {\hard} (25-30) task for {\consecutivetableread}. On {\visualanalogy}, {\contalcmft} outperforms {\alcmft} by $15\%$p with $12 \times 10^4$ training data. These results suggest that future stronger LLM backbone can further close the generalization gap between text and image modalities using our proposed strategy.

\subsection{Utility of our synthetic datasets for existing evaluation benchmarks} 
\label{sec:utility}

Existing evaluation benchmarks test the ability of VLMs to perform OCR, chart interpretation, image reasoning, and caption generation. However, they primarily test the ability to generate short answers on a given question. On the other hand, our created tasks evaluate long form reasoning generation from models. As such, our fine-tuned models quickly forget to return short responses during training and struggle on existing benchmarks. 

Nonetheless, we assess the utility of our synthetic datasets to existing benchmarks by including them during visual instruction tuning. That is,  we prepare two different versions of an alternate base model \textit{Eagle+Synthetic-X2-Llama-8B}: 1) where 30 / 60 / 120k of our {\smftimage} training mixture (equal mix of \simple{} and \hard{}) has been mixed in with the 1.8M finetuning data; or where 240k of our {\cmft} training mixture (80k for each synthetic task) has been mixed in with the 1.8M finetuning data. 

Results in \cref{tab:results_realworld} demonstrate consistent improvements across tasks such as OCR, chart interpretation, and multimodal understanding. However, a decline in performance is also observed on binary classification (yes/no response) benchmarks, such as MME. These findings indicate that the proposed synthetic datasets can be valuable for future research. Further investigation is necessary to determine how long reasoning datasets like our proposed tasks can be best leveraged to enhance general reasoning capabilities (e.g. \citet{gao2024prolong}).

\begin{table}[!h]
    \caption{\textbf{Utility of our synthetic data:} We compare the benchmark results of \textit{Eagle-X2-Llama-8B}, solely instruction tuned on Eagle-1.8M dataset, and \textit{Eagle+Synthetic-X2-Llama-8B}, instruction tuned on a mixture of Eagle-1.8M and our synthetic data mixture. Including our data can improve model's performance on OCR and chart reasoning benchmarks, but may hurt performance on benchmarks where models need to output a yes/no answer (marked with *) or a short phrase (marked with **). $\dagger$: performance reported on validation set. }
    \label{tab:results_realworld}
    \centering
    \begin{tabular}{l|c|ccc|c}
        \toprule
        & \multicolumn{5}{c}{Visual Instruction Tuning Dataset}  \\
        \midrule
        \multirow{3}{*}{\shortstack[l]{Evaluation \\ Benchmark}} & \multirow{3}{*}{Eagle-1.8M} & \multicolumn{3}{c|}{Eagle-1.8M} & Eagle-1.8M \\
         & & \multicolumn{3}{c|}{+ \simple{} {\smftimage} and \hard{} {\smftimage}} & + {\cmft} mixture  \\
         & & (30k) & (60k) & (120k) & (240k) \\
        \midrule
        MMMU$\dagger$ & 35.4 & 38.2 & \textbf{38.8} & 38.7 &  36.3 \\
        MME* &  \textbf{1529} & 1242 & 1377 & 1376 &  1364\\
        MMBench & 67.6 & \textbf{69.2} & 68.4 & 67.5 &  \textbf{69.2} \\
        \midrule
        POPE* & 86.6 & 88.7 & \textbf{88.9} & 87.6 &  87.5  \\
        TextVQA**  & 66.8 & 66.5 & \textbf{66.9} & 66.8 &  65.8\\
        OCR(Bench) & 47.3 & \textbf{50.9} & 50.4 & 47.0 &  48.2 \\
        \midrule
        ChartQA & 69.6 & \textbf{71.6} & 69.8 & 70.5 & 69.4 \\
        CharXiv-Reasoning$\dagger$ & 16.8 & 16.4 & 16.5 & 17.0  &  \textbf{17.2} \\
        CharXiv-Descriptive$\dagger$ & 30.7 & 28.3 & \textbf{35.8} & 31.1 & 34.4 \\
        \bottomrule
    \end{tabular}
\end{table}

\textbf{Reported benchmarks:} Here is a summary of the reported evaluation benchmarks.
\begin{itemize}
    \item MMMU \citep{yue2024mmmu}: Evaluates on multi-discipline tasks measuring college-level subject knowledge and reasoning. 
    \item MME \citep{Fu2023MME}: Evaluates both perception and cognition abilities across 14 subtasks with yes/no answers.
    \item  MMBench \citep{liu2025mmbench}: Evaluates on VQA, which includes both multiple-choice and free-form answers.
    \item  POPE \citep{Li-hallucination-2023}: Evaluates object hallucination with yes/no answers.
    \item TextVQA \citep{singh2019towards}: Evaluates understanding and reading text within images with short-phrase answers.
    \item  OCR(Bench) \citep{liu2024ocrbenchhiddenmysteryocr}: Evaluates on Character Recognition (OCR) capabilities across 29 datasets covering text / handwritten mathematical expression recognition, key information extraction, and scene text / document VQA.
    \item CharXiv \citep{wang2024charxiv}: Evaluates on chart understanding based on 2323 charts from arXiv papers, paired with descriptive and reasoning questions, covering 8 major academic subjects.
\end{itemize}

\clearpage
\section{Continued Discussion on Gradient Alignment}
\label{sec:grad_align}

\textbf{Augmentation in notation:} Suppose the current model parameters are given by $\theta$. We will slightly augment our notation to include the model's current parameters. At model parameter $\theta$, we will use $\evalitoa(\data; \theta)$ to denote the loss on {\smftimage} example for data $\data$ and loss $\evalitoahard(\theta) = \mathbb{E}_{\data \in \inputset_{\hard}} \evalitoa(\data; \theta)$. Then, $\mathbf{g}_{\simple} (\theta)$ and $\mathbf{g}_{\hard}(\theta) $ denote average gradients on $\inputset_{\simple}$ and $\inputset_{\hard}$, i.e. 
\begin{equation*}
    \mathbf{g}_{\simple} (\theta) = \mathbb{E}_{\data \in \inputset_{\simple}} \nabla \evalitoa(\data; \theta),  \quad \mathbf{g}_{\hard}(\theta) = \mathbb{E}_{\data \in \inputset_{\hard}} \nabla \evalitoa(\data; \theta)
\end{equation*}

Recall that the gradient alignment score from \cref{eq:grad_align} is given by
\begin{equation}
    \langle \mathbf{g}_{\simple}(\theta), \mathbf{g}_{\hard}(\theta) \rangle / \langle \mathbf{g}_{\hard}(\theta), \mathbf{g}_{\hard}(\theta) \rangle .\label{eq:grad_align_app}
\end{equation}
In the following theorem, we show that the gradient alignment score quantifies the amount of loss that we can decrease in expectation on \hard{} {\smftimage} examples by taking gradients on \simple{} {\smftimage} examples, relative to taking gradients on \hard{} {\smftimage} examples.

\begin{theorem}\label{lem:grad_align_SGD}
    Suppose for a model $\model$ with parameter $\theta$, loss $\evalitoa$ is Lipschitz and has bounded gradient norm on $\inputset$ around parameters $\theta$, with $\norm{\mathbf{g}_{\hard} (\theta)} \ne 0$.
    The following holds true for expected drop in $\evalitoahard$ with SGD when using a random training sample from the {\simple} task, compared to using a random training sample from the {\hard} task:
    \begin{align*}
        \lim_{\eta \to 0} \frac{ \mathbb{E}_{ \substack{\data \in \inputset_{\simple} \\ \mathbf{g} := \nabla \evalitoa (\data; \theta)} } \left[ \evalitoahard( \theta - \eta \mathbf{g} )  - \evalitoahard( \theta) \right] }{ \mathbb{E}_{ \substack{\data \in \inputset_{\hard} \\ \Tilde{\mathbf{g}} := \nabla \evalitoa (\data; \theta)} } \left[ \evalitoahard( \theta - \eta \Tilde{\mathbf{g}} )  - \evalitoahard( \theta) \right]} = \langle \mathbf{g}_{\hard}(\theta) , \mathbf{g}_{\simple} (\theta)\rangle / \langle \mathbf{g}_{\hard}(\theta) , \mathbf{g}_{\hard}(\theta) \rangle.
    \end{align*} 
\end{theorem}
The proof follows from standard convergence analysis of gradient descent algorithm \citep{nesterov2018lectures}.

\subsection{Proof of \cref{lem:grad_align_SGD}}
\label{proof:grad_align_SGD}

\begin{proof}
    Say $\mathbf{g} = \nabla \evalitoa (\data; \theta)$. By Taylor's theorem, we have the following for a small enough learning rate $\eta$,
    \begin{equation*}
        \evalitoahard( \theta - \eta \mathbf{g} )  - \evalitoahard( \theta) = -\eta \langle \nabla \evalitoahard(\theta), \mathbf{g} \rangle + \eta^2 \mathbf{g}^\intercal \left( \nabla^2 \evalitoahard(\theta -\eta_0 \mathbf{g}) \right) \mathbf{g}
    \end{equation*}
    for some $\eta_0 \in [0, \eta]$. We first note that $\nabla \evalitoahard(\theta) = \mathbf{g_{\hard}}(\theta)$. Next, since the loss is assumed to be Lipschitz,
    \begin{equation*}
        \left\vert \mathbf{g}^\intercal \left( \nabla^2 \evalitoahard(\theta -\eta_0 \mathbf{g}) \right) \mathbf{g} \right\vert \leq L \norm{\mathbf{g}}^2
    \end{equation*}
    where $L$ is the Lipschitz constant for the loss. Since the gradient norms are also assumed to be bounded, we have
    \begin{equation*}
        \evalitoahard( \theta - \eta \mathbf{g} )  - \evalitoahard( \theta) = -\eta \langle \mathbf{g_{\hard}}(\theta) , \mathbf{g} \rangle + \mathcal{O}(\eta^2),
    \end{equation*}
    First assume $\data \in \inputset_{\simple}$. By taking expectation over $\data$,
    \begin{align*}
        \mathbb{E}_{ \substack{\data \in \inputset_{\simple} \\ \mathbf{g} := \nabla \evalitoa (\data; \theta)} } \left[ \evalitoahard( \theta - \eta \mathbf{g} )  - \evalitoahard( \theta) \right] 
        &= -\eta \left\langle \mathbf{g_{\hard}}(\theta) , \mathbb{E}_{\data \in \inputset_{\simple}} \nabla \evalitoa (\data; \theta) \right\rangle + \mathcal{O}(\eta^2) \\
        &= -\eta \left\langle \mathbf{g_{\hard}}(\theta) ,\mathbf{g_{\simple}}(\theta) \right\rangle + \mathcal{O}(\eta^2)
    \end{align*}
    Similarly, assume $\Tilde{\mathbf{g}} = \nabla \evalitoa (\data; \theta)$ where $\data \in \inputset_{\hard}$. By taking expectation over $\data$,
    \begin{align*}
        \mathbb{E}_{ \substack{\data \in \inputset_{\hard} \\ \Tilde{\mathbf{g}} := \nabla \evalitoa (\data; \theta)} } \left[ \evalitoahard( \theta - \eta \Tilde{\mathbf{g}} )  - \evalitoahard( \theta) \right] 
        &= -\eta \left\langle \mathbf{g_{\hard}}(\theta) , \mathbb{E}_{\data \in \inputset_{\hard}} \nabla \evalitoa (\data; \theta) \right\rangle + \mathcal{O}(\eta^2) \\
        &= -\eta \left\langle \mathbf{g_{\hard}}(\theta) ,\mathbf{g_{\hard}}(\theta) \right\rangle + \mathcal{O}(\eta^2)
    \end{align*}
    Therefore, we have
    \begin{equation*}
        \frac{ \mathbb{E}_{ \substack{\data \in \inputset_{\simple} \\ \mathbf{g} := \nabla \evalitoa (\data; \theta)} } \left[ \evalitoahard( \theta - \eta \mathbf{g} )  - \evalitoahard( \theta) \right] }{ \mathbb{E}_{ \substack{\data \in \inputset_{\hard} \\ \Tilde{\mathbf{g}} := \nabla \evalitoa (\data; \theta)} } \left[ \evalitoahard( \theta - \eta \Tilde{\mathbf{g}} )  - \evalitoahard( \theta) \right]} = \frac{-\eta \left\langle \mathbf{g_{\hard}}(\theta) ,\mathbf{g_{\simple}}(\theta) \right\rangle + \mathcal{O}(\eta^2)}{-\eta \left\langle \mathbf{g_{\hard}}(\theta) ,\mathbf{g_{\hard}}(\theta) \right\rangle + \mathcal{O}(\eta^2)} = \frac{\left\langle \mathbf{g_{\hard}}(\theta) ,\mathbf{g_{\simple}}(\theta) \right\rangle + \mathcal{O}(\eta)}{\left\langle \mathbf{g_{\hard}}(\theta) ,\mathbf{g_{\hard}}(\theta) \right\rangle + \mathcal{O}(\eta)}
    \end{equation*}
    Note that $\mathbf{g_{\hard}}(\theta)$ and $\mathbf{g_{\simple}}(\theta)$ do not depend on the value of $\eta$. Furthermore, by assumption, $\norm{\mathbf{g_{\hard}}(\theta)} \ne 0$. We conclude by taking $\eta \rightarrow 0$ on both sides of the equation above.
\end{proof}

\subsection{Additional measure 1: gradient cosine similarity}
\label{sec:grad_cosine}
We additionally define the \textbf{gradient cosine similarity score} as the cosine similarity of gradients from $\inputset_{\simple}$ and $\inputset_{\hard}$:

\begin{equation}
    \text{Gradient Cosine Similarity:} \quad\quad \frac{\langle \mathbf{g}_{\simple}(\theta), \mathbf{g}_{\hard}(\theta)
     \rangle }{\sqrt{\langle \mathbf{g}_{\hard}(\theta), \mathbf{g}_{\hard}(\theta) \rangle \cdot \langle \mathbf{g}_{\simple}(\theta), \mathbf{g}_{\simple}(\theta) \rangle}}
     \label{eq:grad_cosine_similarity}
\end{equation}
Note that this measure ignores the norm of the gradients on $\inputset_{\simple}$ that the model uses during training. Hence, this measure is not an entirely faithful measure on the alignment of the training updates to the loss on {\hard} {\smftimage} examples. \cref{fig:utility_gradient_ItoA_ID_OOD_cosine} shows the gradient cosine similarity score across training strategies for {\tableread}, which follows a similar pattern as the gradient alignment score in \cref{fig:utility_gradient_ItoA_ID_OOD}.

\begin{figure}[!t]
    \centering
    \includegraphics[width=0.65\linewidth]{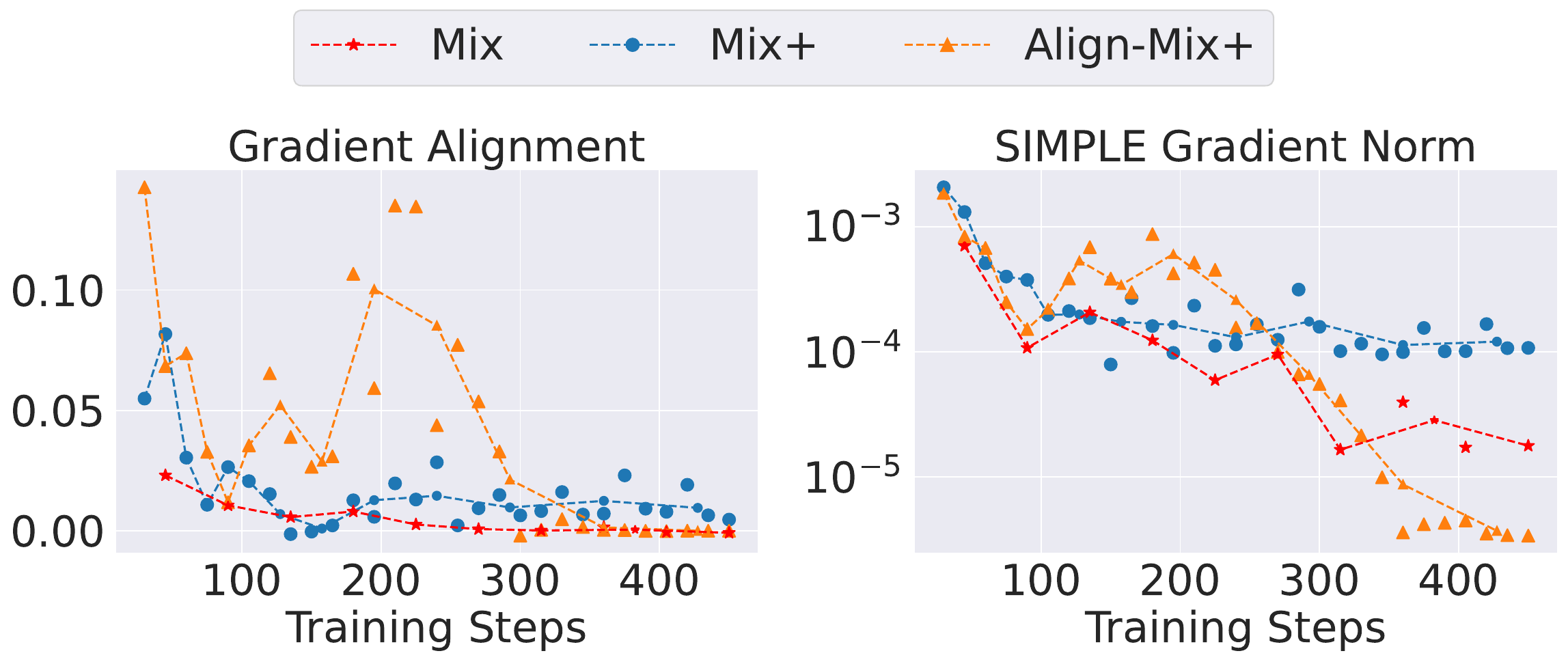}
    \captionof{figure}{\textbf{Analysis of gradients on {\tableread} (additional plots for \cref{fig:utility_gradient_ItoA_ID_OOD_contour}):} (Left) Gradient Alignment Score (\cref{eq:grad_align}); (Right) Average Gradient Norm on {\simple} {\smftimage} examples $\mathbb{E}_{\data \in \inputset_{\simple}} \norm{\nabla {\evalitoa} (\data)}$. {\alcmft} has higher gradient alignment score in the initial phases of training, where it also has higher gradient norm. {\cmft} shows higher gradient alignment score than {\mixft} during the course of training.}
    \label{fig:utility_gradient_ItoA_ID_OOD}
\end{figure}

\begin{figure}[!t]
    \centering
    \includegraphics[width=\linewidth]{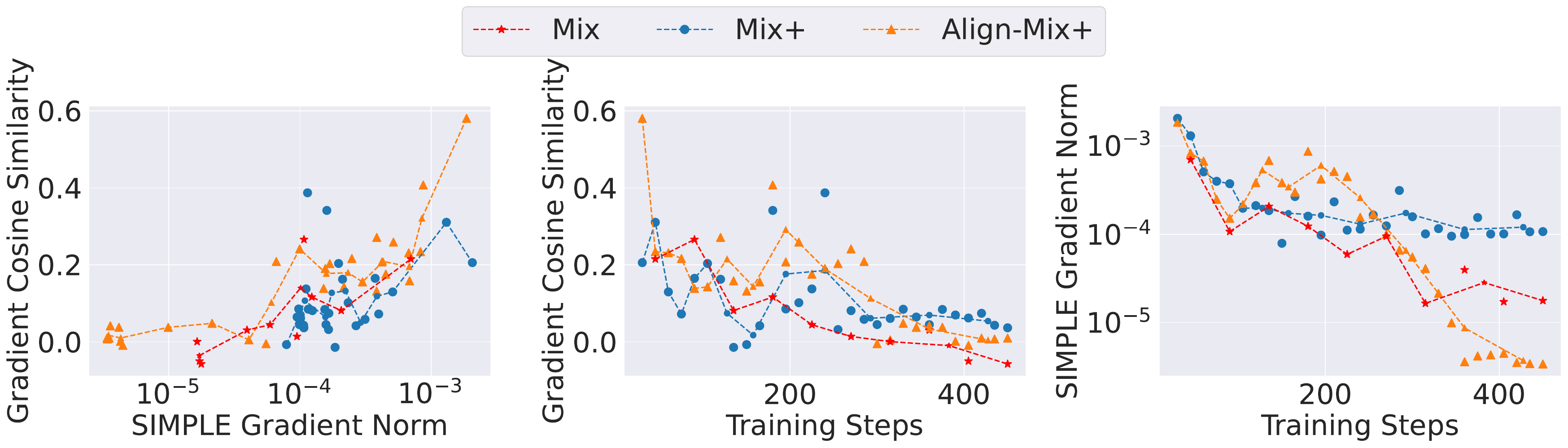}
    \captionof{figure}{\textbf{Analysis of gradients on {\tableread} (replacing gradient alignment score from \cref{fig:utility_gradient_ItoA_ID_OOD_contour,fig:utility_gradient_ItoA_ID_OOD} with gradient cosine similarity):} (Left) Average Gradient Norm on \simple{} {\smftimage} examples $(\mathbb{E}_{\data \in \inputset_{\simple}} \norm{\nabla \evalitoa (\data)})$ vs. Gradient Cosine Similarity (\cref{eq:grad_cosine_similarity}) for different training checkpoints; (Middle) Gradient Cosine Similarity; (Right) Average Gradient Norm. Similar results hold.}
    \label{fig:utility_gradient_ItoA_ID_OOD_cosine}
\end{figure}%

\subsection{Additional measure 2: Adam update alignment}
\label{sec:adam_update}
The gradient alignment score we defined earlier does not account for the fact that we use Adam optimizer \citep{kingma2017adammethodstochasticoptimization} during our experiments. 

\textbf{Brief definition of Adam:} The Adam optimizer maintains two additional states, each representing the running average of the gradients and their squares during training. If $\mathbf{m}_t$ and $\mathbf{v}_t$ denote the two states, then the update rule at training step $t$ with a gradient $\mathbf{g}_t$ and learning rate $\eta$ is given by
\begin{align*}
    &\theta \gets \theta - \eta h(\mathbf{g}_t), \quad \quad \text{ where } h(\mathbf{g}_t) = \frac{(1-\beta_1) \mathbf{g}_t + \beta_1 \mathbf{m}_{t-1}}{ \sqrt{(1-\beta_2) \mathbf{g}_t \odot \mathbf{g}_t + \beta_2 \mathbf{v}_{t-1} } + \epsilon } \\
    & \mathbf{m}_{t} \gets (1-\beta_1) \mathbf{g}_t + \beta_1 \mathbf{m}_{t-1}, \quad \quad \mathbf{v}_{t} \gets (1-\beta_2) \mathbf{g}_t \odot \mathbf{g}_t + \beta_2 \mathbf{v}_{t-1}
\end{align*}
Here, $(\beta_1, \beta_2, \epsilon)$ are hyperparameters for the Adam optimizer and are set at $(0.9, 0.999, 10^{-8})$. 

\textbf{Adam Update Alignment:} A true measure of alignment between {\simple} and {\hard} training would be to compare $h(\cdot)$, the update vector under the Adam optimizer. However, that requires saving the Adam optimizer states throughout training. For storage efficiency purposes\footnote{Remark: retrieving the actual Adam optimizer states requires an additional storage of 138GB per checkpoint.}, we propose an alternate approximate measure called the \textbf{Adam update alignment score}. We compute the following two quantities for model $\model$ with parameters $\theta$:
\begin{align*}
    \mathbf{m}(\theta) &:= \mathbb{E}_{\data \in \inputset_{\simple}}  \left[ \nabla  \evalitoa(\data; \theta) \right] \\
    \mathbf{v}(\theta) &:= \mathbb{E}_{\data \in \inputset_{\simple}}  \left[ \nabla \evalitoa(\data; \theta) \odot \nabla \evalitoa(\data; \theta) \right]
\end{align*}
$\mathbf{m}$ and $\mathbf{v}$ are proxy measures for the Adam optimizer states. Then, we measure the alignment between gradients for $\inputset_{\hard}$ and $\inputset_{\simple}$ as
\begin{align}
    \text{Adam Update Alignment Score:} 
    \quad\quad  &\frac{ \mathbb{E}_{ \substack{\data \in \inputset_{\simple} \\ \mathbf{g} := \nabla \evalitoa (\data; \theta)} }  \langle h(\mathbf{g}),  \mathbf{g}_{\hard}(\theta) \rangle }{   \mathbb{E}_{  \substack{\data \in \inputset_{\hard} \\ \Tilde{\mathbf{g}} := \nabla \evalitoa (\data; \theta)}  } \langle h(\Tilde{\mathbf{g}}),  \mathbf{g}_{\hard}(\theta) \rangle } \label{eq:grad_adam_align} \\&
    \text{where } h(\mathbf{g}) = \frac{(1-\beta_1) \mathbf{g} + \beta_1 \mathbf{m}(\theta)}{ \sqrt{(1-\beta_2) \mathbf{g} \odot \mathbf{g} + \beta_2 \mathbf{v}(\theta) } + \epsilon } \text{ for any vector } \mathbf{g} \nonumber
\end{align}
Intuitively, this measures how much the loss $\evalitoahard$ can be reduced in expectation by taking a gradient update step with Adam using {\simple} {\smftimage} examples, compared to taking a gradient update step with {\hard} {\smftimage} examples, while maintaining the current Adam optimizer states. This can be formalized in the following theorem.

\begin{theorem}
   Suppose for a model $\model$ with parameter $\theta$, loss $\evalitoa$ is Lipschitz and has bounded gradient norm on $\inputset$ around parameters $\theta$, with $\norm{\mathbf{g}_{\hard} (\theta)} \ne 0$. Consider a modified Adam update with learning rate $\eta$ with an arbitrary gradient $\mathbf{g}$, as follows:
    \begin{align*}
        \theta &\gets \theta - \eta h(\mathbf{g})\\ 
        &\text{where } h(\mathbf{g}) = \frac{(1-\beta_1) \mathbf{g} + \beta_1 \mathbf{m}(\theta)}{ \sqrt{(1-\beta_2) \mathbf{g} \odot \mathbf{g} + \beta_2 \mathbf{v}(\theta) } + \epsilon }.
    \end{align*}

    The following holds true for expected drop in $\evalitoahard$ with modified Adam update when using a random training sample from the {\simple} task, compared to using a random training sample from the {\hard} task:
    \begin{align*}
        \lim_{\eta \to 0} \frac{ \mathbb{E}_{\substack{\data \in \inputset_{\simple} \\ \mathbf{g} := \nabla \evalitoa (\data; \theta)}} \left[ \evalitoahard( \theta - \eta h(\mathbf{g}))  - \evalitoahard( \theta) \right] }{ \mathbb{E}_{ \substack{\data \in \inputset_{\hard} \\ \Tilde{\mathbf{g}} := \nabla \evalitoa (\data; \theta)} } \left[ \evalitoahard( \theta - \eta h( \Tilde{\mathbf{g}} )) - \evalitoahard( \theta) \right]} = \frac{ \mathbb{E}_{\substack{\data \in \inputset_{\simple} \\ \mathbf{g} := \nabla \evalitoa (\data; \theta)}}  \langle h(\mathbf{g}),  \mathbf{g}_{\hard}(\theta) \rangle }{   \mathbb{E}_{
    \substack{\data \in \inputset_{\hard} \\ \Tilde{\mathbf{g}} := \nabla \evalitoa (\data; \theta)}  } \langle h( \Tilde{\mathbf{g}} ),  \mathbf{g}_{\hard}(\theta) \rangle }
    \end{align*} 
\end{theorem}

The proof is similar to that of \cref{lem:grad_align_SGD}.

\subsection{Experimental results} 
In \cref{fig:Utility_ID_OOD_ctr_adamupdate}, we present the analysis of gradients for different types of supervision on {\consecutivetableread}. Similar to the behavior of gradient alignment score in \cref{fig:Utility_ID_OOD_ctr}, we observe that when measured against norm of gradients on {\simple} {\smftimage} examples, \mixft\ achieves a higher Adam update alignment score than both {\imagetext} and {\smftimage}. This shows that {\imageviatext} supervision improves the alignment between \simple\ and \hard\ {\smftimage} gradients, when taking Adam gradient updates into account.

Similarly, for {\tableread} in \cref{fig:Utility_ID_OOD_Align_norm_alt_adamupdate}, {\cmft} has a larger Adam update alignment during training. {\alcmft} further improves the Adam update alignment score when gradient norms are large during training.

\begin{figure}[!ht]
    \centering
    \includegraphics[width=0.7\linewidth]{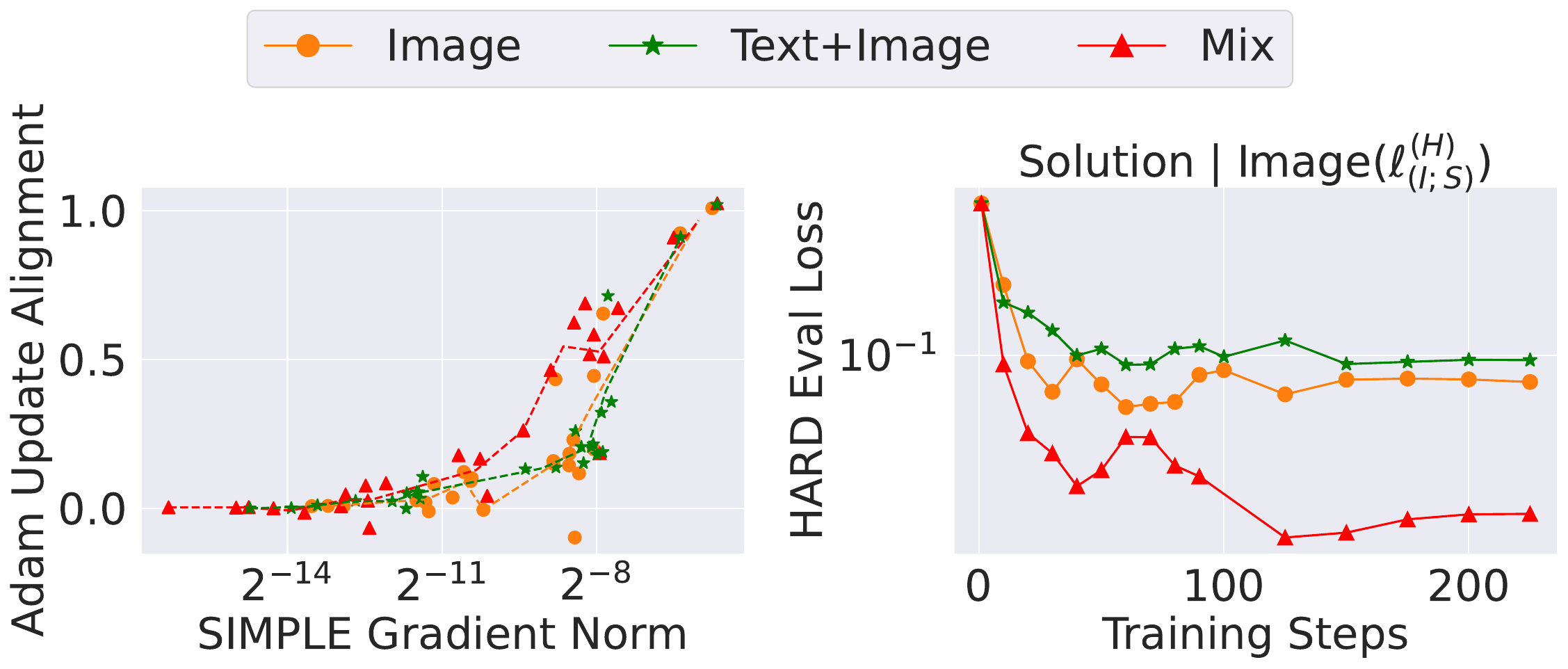}
    \caption{\textbf{Analysis of gradients on {\consecutivetableread} (replacing gradient alignment score from \cref{fig:Utility_ID_OOD_ctr} with Adam update alignment score):} (Left) Average Gradient Norm  on \simple{} {\smftimage} examples $(\mathbb{E}_{\data \in \inputset_{\simple}} \norm{\nabla \evalitoa (\data)})$ vs. Adam Update Alignment Score (\cref{eq:grad_adam_align}) for different training checkpoints; (Right) Average \textit{Loss on solution given {\hard} image} ($\evalitoahard$) during training. Similar results hold.}
    \label{fig:Utility_ID_OOD_ctr_adamupdate}
\end{figure}

\begin{figure}[!ht]
    \centering
    \includegraphics[width=\linewidth]{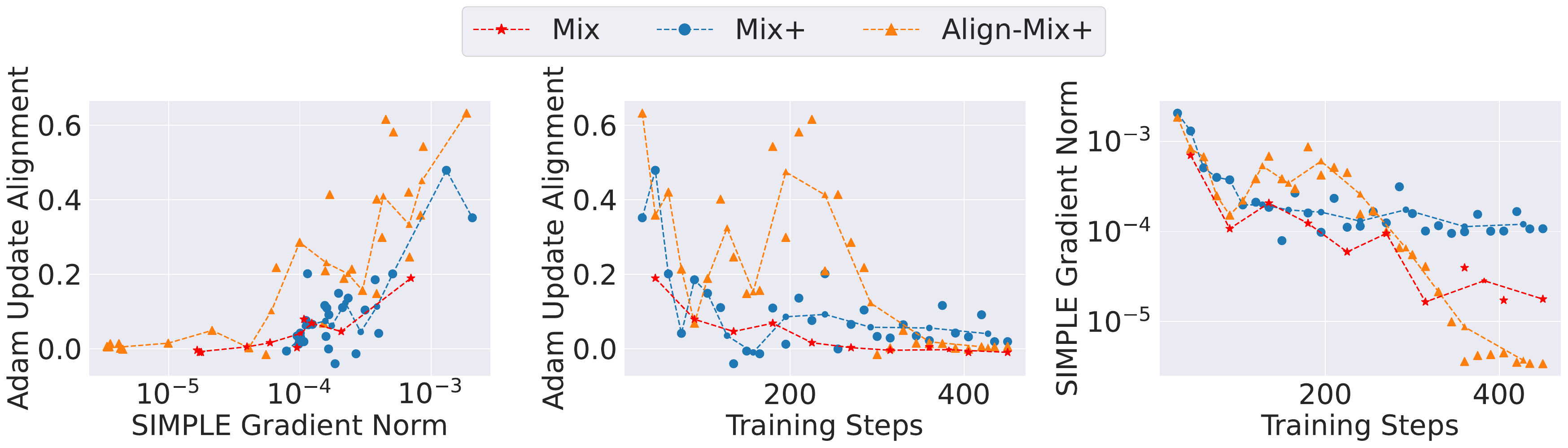}
    \caption{\textbf{Analysis of gradients on {\tableread} (replacing gradient alignment score from \cref{fig:utility_gradient_ItoA_ID_OOD_contour,fig:utility_gradient_ItoA_ID_OOD} with Adam update alignment score):} (Left) Average Gradient Norm on {\simple} {\smftimage} examples $(\mathbb{E}_{\data \in \inputset_{\simple}} \norm{\nabla \evalitoa (\data)})$ vs. Adam Update Alignment Score (\cref{eq:grad_adam_align}) for different training checkpoints; (Middle) Adam Update Alignment Score; (Right) Average Gradient Norm. Similar results hold.}
    \label{fig:Utility_ID_OOD_Align_norm_alt_adamupdate}
\end{figure}
\begin{figure*}[!ht]
    \centering
    \includegraphics[width=0.99\linewidth]{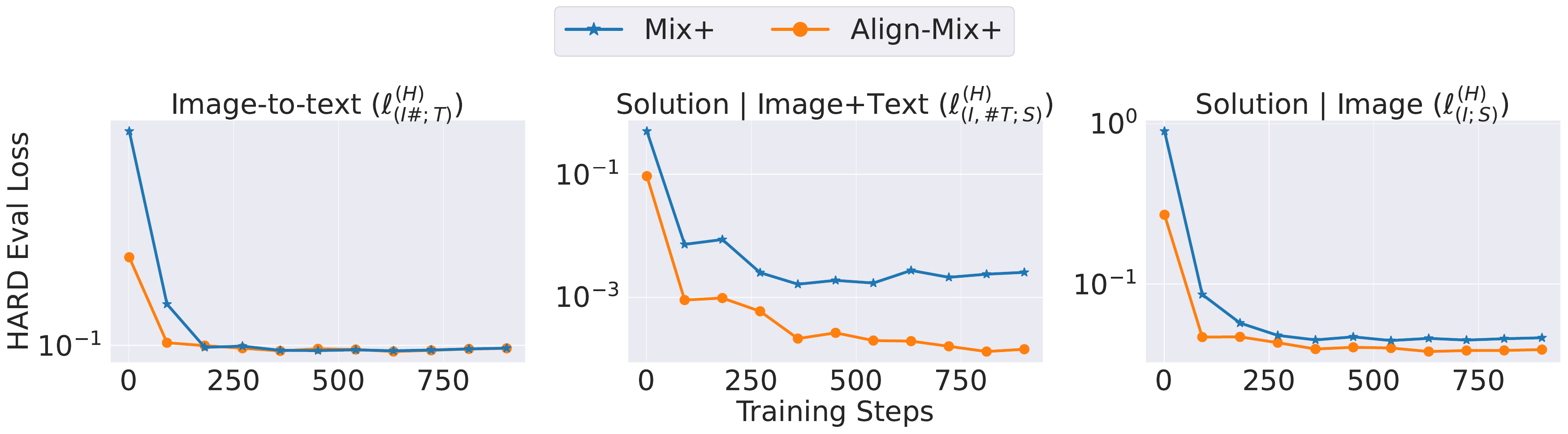}
    \caption{\textbf{Analysis of evaluation losses (repeating \cref{fig:loss_behavior} for {\visualanalogy}):} (Left) \textit{{\hard} image-to-text conversion loss} ($\evalitothard$ (Eq.\ref{eq:itot})); (Middle) \textit{loss on solution given {\hard} image and text} ($\evalittoahard$ (Eq.\ref{eq:ittoa})); (Right) \textit{loss on solution given {\hard} image} ($\evalitoahard$ (Eq.\ref{eq:hard-loss})). Similar results hold.}
    \label{fig:loss_behavior_VA}
\end{figure*}

\clearpage
\section{Additional Ablations}
\label{sec:ablate_nontextgen}

\subsection{Performance of other multimodal models}
\label{app:other_models}
In \cref{tab:llm_apis_close_source}, we present the performance of three closed source and two open source multimodal models on our three {\nontextgeneralizable} tasks. Since we do not train these models, the format of the outputs is more flexible. For convenience, we propose alternative metrics to extract and evaluate on the models' predictions. For \tableread{}, we instead evaluate with the final answer (sum of the sequence of numbers)\footnote{The closed source models have access to tool-use, so in theory, this should be an equivalent, if not more lenient, metric.}. For \gridnav, we evaluate with the same metric as in the main part of the paper --- whether the proposed path can move from the start cell to the end without running into obstacles. For \visualanalogy{}, we evaluate with just the final option, which is a more lenient metric than the one suggested in the main part of the paper. Note that a random choice baseline should get 25\%. 

\begin{table}[!h]
    \caption{\textbf{Performance of other multimodal models:} For convenience, we evaluate the models on a slightly different metric.}
    \label{tab:llm_apis_close_source}
    \centering
    \begin{tabular}{l||rr||rr||rr}
        \toprule
         & \multicolumn{2}{|c||}{\tableread} & \multicolumn{2}{c||}{\gridnav} & \multicolumn{2}{c}{\visualanalogy} \\
        \midrule
        Models & \simple & \hard & \simple & \hard & \simple & \hard \\
        \midrule
        Claude-3.5 Sonnet & 30.0 & 0.0 & 0.0 & 0.0 & 35.0 & 29.8 \\
        GPT-4o & 19.0 & 0.0 & 0.0 & 0.0 & 19.8 & 18.4 \\
        OpenAI o1 & 29.0 & - & 0.0 & - & 30.6 & - \\
        Llama3.2-11B-Vision-Instruct & 4.0 & 0.0 & 0.0 & 0.0 & 16.2 & 17.8 \\
        Pixtral-12B \citep{agrawal2024pixtral} & 9.0 & 0.2 & 0.0 & 0.0 & 24.6 & 21.2 \\
        \bottomrule
    \end{tabular}
\end{table}

\subsection{Ablation of the {\cmft} supervision}
\label{app:ablations_component}
The dataset composition of the {\cmft} supervision consists of three types of supervision in the \simple{} task: {\smfttext}, {\smftimage}, and {\imageviatext}. In this section, we ablate on the importance of each component of the data mixture in the training of {\cmft}, {\contcmft}, and {\contalcmft}. In \cref{fig:ablations_component_ar}, we report the {\simpletohard} performance on image when the \simple{} {\mixft} supervision is replaced with \textit{a varying data composition}. 

In single-stage training (no text warm-up or alignment), {\imageviatext} is the key component of success, as evidenced by the strong performance of {\cimageviatext} supervision. As noted in \cref{sec:tr_ar_gn}, {\cmft} can match the performance by explicitly prompting the resulting model to convert the image first, which comes at a cost of around $1.7$x generated tokens at inference time. 

In multi-stage training (either \tw{} or \tw{ \al{}}), the benefits of {\cmft} are more significant. Specifically, among all other types of supervision with text warm-up, {\contcmft} is able to outperform the others by at least $1.7$x, while retaining efficient inference costs, unlike \tw{ \cimageviatext{}}. Among all types of supervision with text warm-up and alignment, {\contalcmft} achieves the highest performance. 

\begin{figure}[!ht]
    \centering
    \includegraphics[width=1.\linewidth]{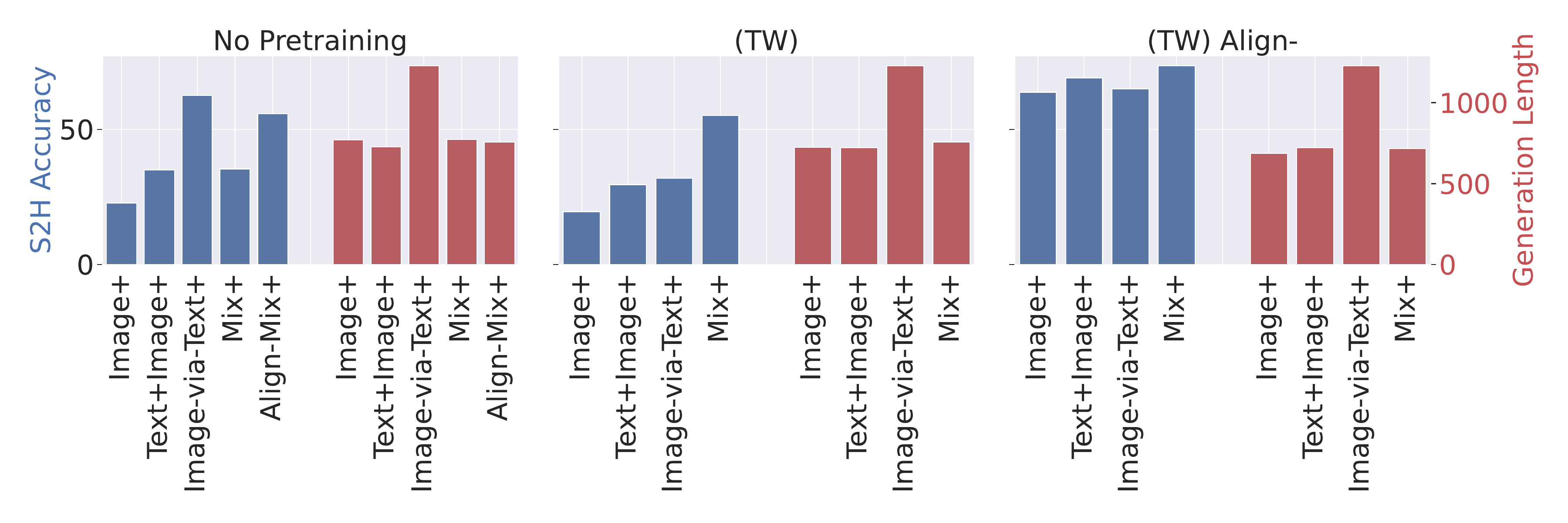}
    \caption{\textbf{Ablation on the \simple{} data composition of {\cmft} on {\visualanalogy}:} Instead of \mixft{}, we use different types of \simple{} supervision in  (Left) {\cmft}; (Middle) {\contcmft}; (Right) {\contalcmft}. The main phase (\cmft{}) uses $12\times 10^4$ training data.}
    \label{fig:ablations_component_ar}
\end{figure}

\subsection{Ablation of the reasoning alignment phase (\al{})}
\label{sec:ablate_alcmft}

We perform two ablations for the first phase of the {\alcmft} supervision. In \cref{fig:ablate_ALdata}, we report the {\simpletohard} generalization performance on image of models trained with \textit{a varying amount of data} in the first phase of {\alcmft}, with the amount of {\cmft} data fixed in the second phase. We don't observe a monotonic improvement in performance when increasing the amount of data in the first phase. In \cref{tab:composition_AL}, we report the {\simpletohard} generalization performance on image of models trained with \textit{a varying data composition} in the first phase of {\alcmft}. Our choice of {\smfttext} and {\imageviatext} from the main section gives the best performance on average on \tableread{} and \visualanalogy{}.

\begin{figure}[!ht]
    \centering
    \begin{minipage}[b]{0.4\textwidth}
        \centering
        \includegraphics[width=\linewidth]{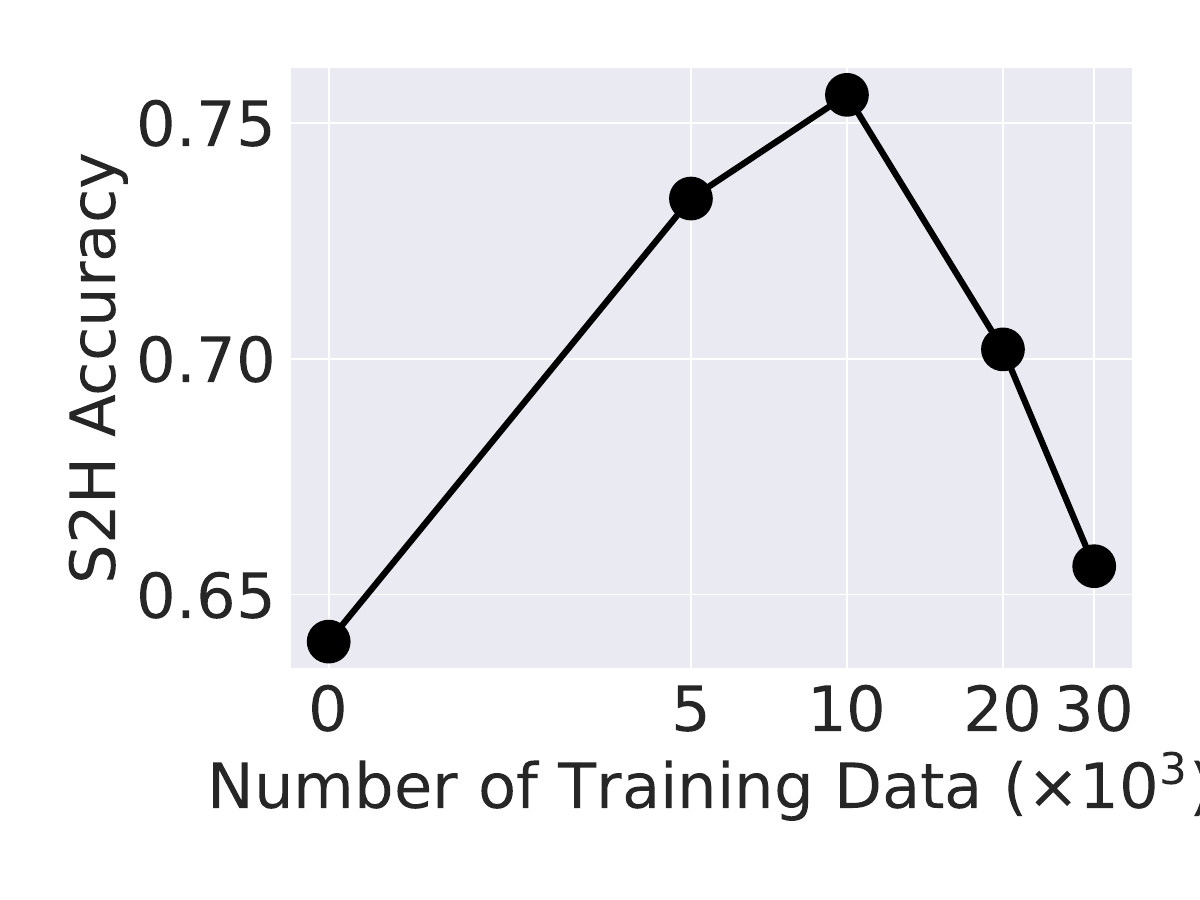}
        \caption{\textbf{Ablation on the amount of data for the alignment phase of {\alcmft} on {\tableread}:} Second phase ({\cmft}) uses $12 \times 10^4$ training data. We don't observe a monotonic improvement in generalization performance with increasing number of training samples in the first phase.}
        \label{fig:ablate_ALdata}
    \end{minipage}%
    \hfill
    \begin{minipage}[b]{0.55\textwidth}
        \centering
        \scalebox{0.8}{
        \begin{tabular}{ccc}
            \hline
            \multirow{2}{*}{Corresponding Name}  & {\simple} data composition & Accuracy  \\
             & in phase 1 ({\alcmft}) & (after phase 2) \\
            \hline
            \multicolumn{3}{c}{{\tableread}} \\
            \hline
            {\al{}} & {\smfttext}, {\imageviatext} & 0.76 \\
            {\imagetext{}} & {\smfttext}, {\smftimage} & 0.52 \\
            {\imageviatext{}}& {\imageviatext} & 0.77 \\
            {\mixft} & {\smfttext}, {\smftimage}, {\imageviatext} & 0.74 \\
            \hline
            \multicolumn{3}{c}{{\visualanalogy}} \\
            \hline
            {\al{}} & {\smfttext}, {\imageviatext} & 0.66 \\
            {\imagetext{}} & {\smfttext}, {\smftimage} & 0.19 \\
            {\imageviatext{}} & {\imageviatext} & 0.51 \\
            {\mixft} & {\smfttext}, {\smftimage}, {\imageviatext} & 0.46 \\
            \hline
        \end{tabular}
        }
        \captionof{table}{\textbf{Ablation on the {\simple} data composition for the alignment phase of {\alcmft}:} Amount of data for the alignment phase is fixed at $10^4$. Second phase ({\cmft}) uses $12 \times 10^4$ training data. Performance is reported on a validation set with $100$ {\hard}-image examples. Our composition of {\smfttext} and {\imageviatext} on the {\simple} task performs best on average on {\tableread} and {\visualanalogy}.}
        \label{tab:composition_AL}
    \end{minipage}
\end{figure}

\subsection{Ablation of the text warm-up pretraining phase \tw{}:}
\label{sec:ablate_text_warmup}
We ablate on the effect of the training data size during the text warm-up. In particular, we are interested in whether models with better reasoning capability on text can achieve better image generalization. To do so, we vary the number of training data used for text warm-up between $\{ 1, 2, 3 \} \times 10^4$ and plot the performance of the warmed-up LLM on {\hard}-text examples against the performance of the final trained model on {\hard}-image examples. We report the performance on \visualanalogy{} in \cref{fig:ablations_multistage}. We observe that model's text performance improves with more training data being used for the text warm-up as expected. However, there is no clear linear correlation between the text capability of the model checkpoint after the warm-up training stage and image {\simpletohard} of the final model. Specifically, a model with $3 \times 10^4$ warm-up performs the best for the {\contcmft} supervision, while a $10^4$ warm-up works the best with the {\contalcmft} supervision. Meanwhile, we observe that {\contalcmft} supervision can universally achieve better {\simpletohard} on image than the {\contcmft} across all data scales. We conclude that an improved text capability by itself is insufficient to guarantee good transfer to image modality. We expect future VLMs with both stronger LLM backbone and better modality alignment can further leverage the text performance and transfer it to images.

\begin{figure}[!ht]
    \centering
    \includegraphics[width=0.6\linewidth]{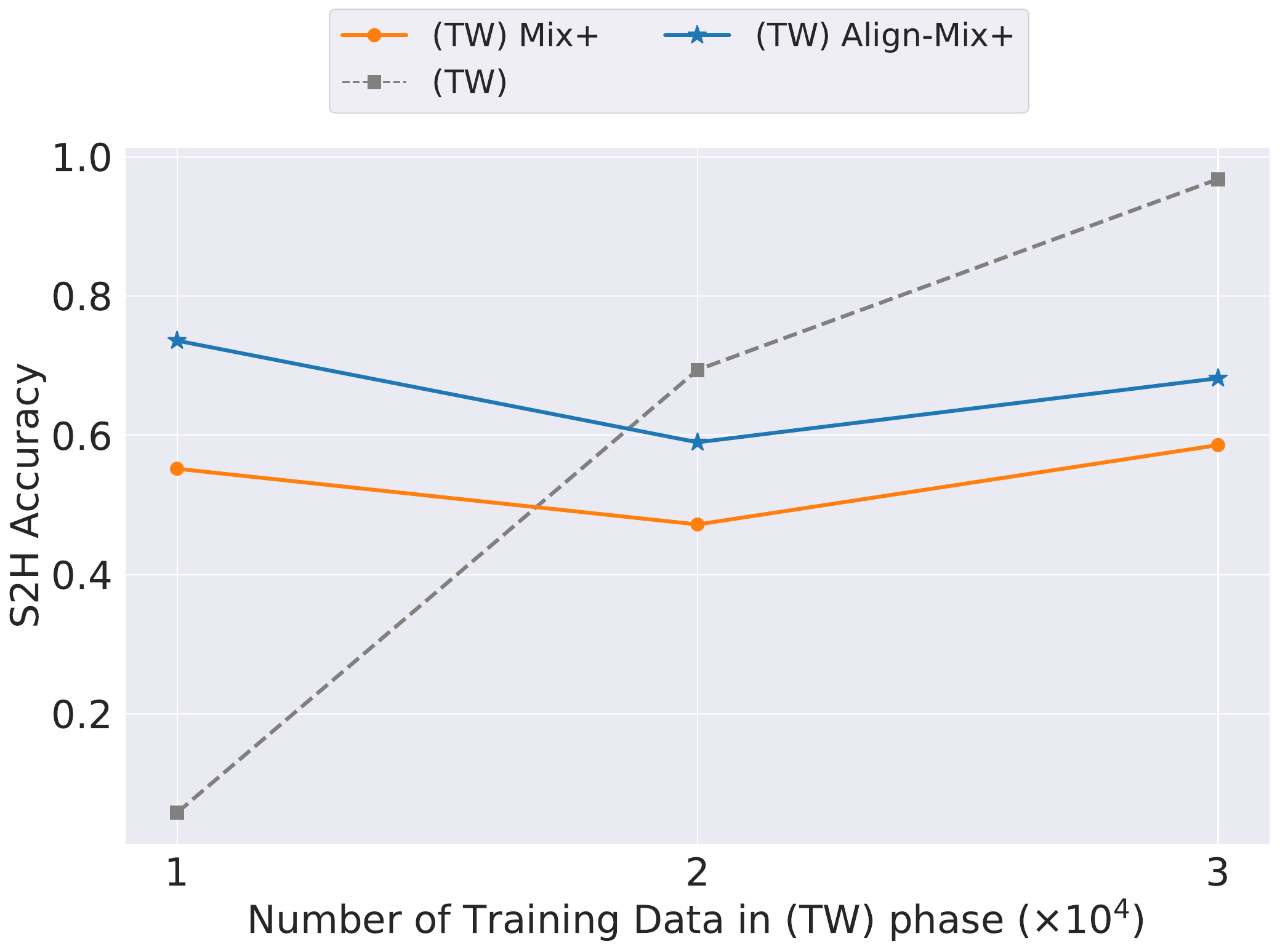}
    \caption{\textbf{Ablation on the amount of data for the text warm-up phase on \visualanalogy{}:} Second phase (\cmft) uses $12 \times 10^4$ training data. Using more data for the warm-up stage results in a stronger LLM backbone with better \hard{}-text performance (gray dashed line), but does not necessarily lead to better image {\simpletohard} of the final model trained with our proposed strategy. This suggests that a stronger text capability is not the only factor that induces {\simpletohard} on image.}
    \label{fig:ablations_multistage}
\end{figure}

\subsection{Requirement of text representation}
\label{app:text_representation}
One potential limitation of our proposed training strategies is the requirement of a text representation corresponding to the image. In {\consecutivetableread}, {\tableread}, and {\gridnav}, we use the LaTeX code of the table or grid, which is considered to be perfectly aligned with the image. In reality, it may be challenging to find an exactly equivalent text description or representation of a real-world image, as many minute visual features cannot be captured by language. We show that our proposed training strategy does not require perfect alignment between the text and the image representation to work. For {\visualanalogy} experiments, the text description of the puzzle in the image is lossy: it only enumerates unique values of all task-relevant attributes without encoding the object to which each corresponds, so one cannot recover the original image given the description (see examples in \cref{fig:ar-domain-transfer-heldout-id,fig:ar-domain-transfer-heldout-ood}). Models trained with our proposed training strategies ({\cmft}, {\alcmft}, {\contalcmft}) all demonstrate significant improvements in image generalization (\cref{fig:compare_itota_cmft}), testifying that our methods work with lossy text representation.

\paragraph{Lossless text representation for \visualanalogy{}:} We additionally conduct experiments where the text representation of the puzzle in the image is a lossless representation. We represent the panels in the puzzle as a code defining each object as a set of attributes. Each geometric object is represented by the values its $5$ attributes:  $\{\texttt{shape type}$, $\texttt{shape\_color}$, $\texttt{shape\_size}$, $\texttt{shape\_quantity}$, $\texttt{shape\_position}\}$, while lines are defined by their $2$ attributes: $\{\texttt{line\_type}$, $\texttt{line\_color}\}$. In order to fit to the context length of the VLM, we describe each object in shorthand notations. For example, for a panel in the puxxle that contains a circle and 2 rectangles, with attribute values $\{45 \text{ (gray-scale)}, 42  \text{ (pixels)}, 1, \text{ top-left}\}$ and  $\{\{0, 90\}, \{21, 21\}, 2, \text{ top-right, bottom-left}\}$, we will represent the panel as 
\begin{equation*}
    \text{CIR-45-42-TL;RECT-0-21-TR;RECT-90-21-BL}
\end{equation*}
We give all details on how to parse the shorthand codes in the prompt. On the other hand, for the same example, the (Lossy) text representation would have been
\begin{align*}
    &\text{type: circle, rectangle} \\
    &\text{color: 0,90} \\
    &\text{size: 21,42} \\
    &\text{quantity: 1,2} \\
    &\text{position: top-left, top-right, bottom-left}
\end{align*}
This substantially reduces the context length on average on our training dataset, and further removes the necessity of parsing a code. However, this isn't an exact representation of the image of the puzzle.

\paragraph{Performance on lossy and lossless \visualanalogy{} tasks:} In \cref{fig:lossless_vs_lossy_AR}, we compare {\cmft}, {\alcmft}, and {\contalcmft} for lossless and lossy \visualanalogy{} tasks at $12 \times 10^4$ training examples in the final phase (\cmft). Our observations reveal that a lossless text representation enhances {\simpletohard} performance on images for {\cmft}. However, for {\alcmft} and {\contalcmft}, the lossy text representation leads to better {\simpletohard} performance on images. This discrepancy could be attributed to the complexity of the shorthand code in the lossless text representation, which requires additional parsing. We did not investigate this phenomenon further.

\begin{figure}[t]
    \centering
    \includegraphics[width=\linewidth]{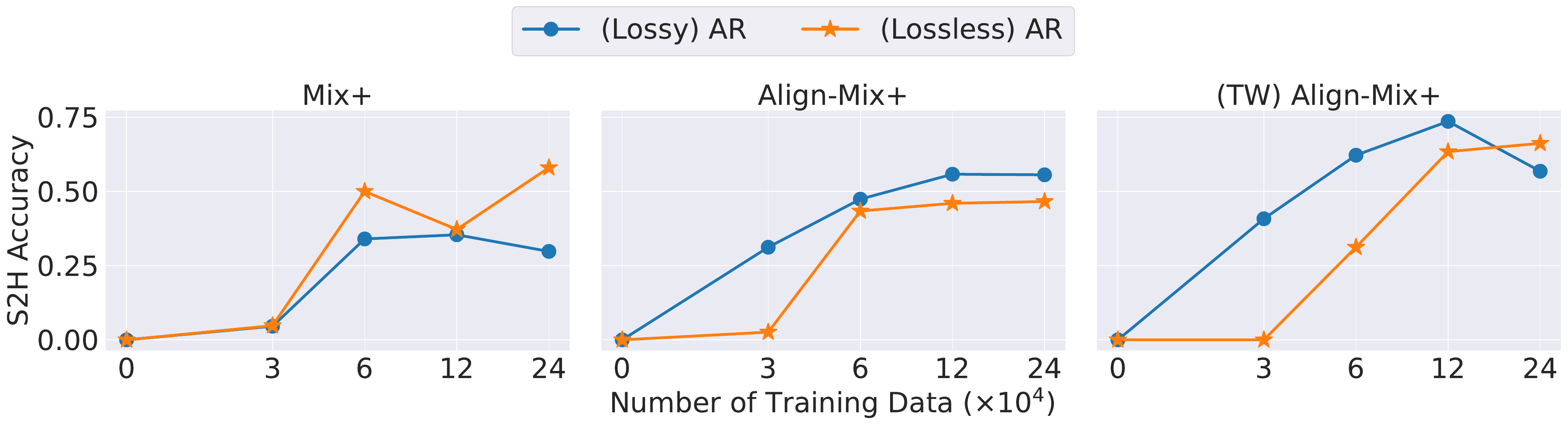}
    \caption{\textbf{Ablation on lossy vs. lossless {\visualanalogy}:} We measure the image {\simpletohard} of different types of supervision for two different versions of text representation for \visualanalogy{}. Models can perform better on Lossless {\visualanalogy} with {\cmft}. However, the trend can change with {\alcmft} and {\contalcmft}.}
    \label{fig:lossless_vs_lossy_AR}
\end{figure}

\subsection{Explicit and implicit text conversion}
\label{app:ablations_text_conversion}
In \cref{app:ablations_component}, we find that explicit text conversion ({\imageviatext}) is the key component in the data composition of the {\cmft} supervision. At inference time, however, models trained with {\cmft} reason directly on {\hard} images, without explicit text conversion. In \cref{tab:ablations_convert_helps}, we observe that the trained models can still perform reasoning with explicit text conversion and that the conversion ability helps it reason.

\paragraph{{\cmft} models can convert image to text when prompted.} If only an image input $\dataimage$ is provided, {\cmft} models will \textit{always} directly predict $\reasoningtrace(\data)$ and $f(\data)$, never converting image to text (under greedy decoding). However, since the prompts used in the {\smftimage} and {\imageviatext} examples are the same, we can induce explicit text conversion in the final trained model by additionally providing the first word ``\texttt{Convert}'' of $P_{convert}$. We find that all trained model are \emph{always} able to continue with explicit text conversion --- they will generate the rest of $P_{convert}$ and an attempted conversion $\datatext$ before $\reasoningtrace(\data)$. The conversion accuracy is around $50\%$ on {\visualanalogy} and is almost $100\%$ on {\tableread}.

\paragraph{Explicit text conversion generally helps the model to reason on image data.} 
Noticeably, the {\cmft} ($240$k) model improves {\simpletohard} accuracy from $73.2\%$ to $96.6\%$ with almost perfect text conversion accuracy of $99.2\%$ on {\tableread}. On {\visualanalogy}, the {\cmft} ($120$k) model improves {\simpletohard} accuracy from $35.4\%$ to $51.8\%$  with a text conversion accuracy of $47\%$. The benefit of explicit text conversion gradually diminishes with multi-stage training strategies.

We also observe a slight drop in performance with prompted text conversion for models trained with {\contalcmft} on {\visualanalogy}, which corresponds to a minor decline in reasoning performance with explicit text conversion. This suggests that the text warm-up training and alignment phase enable the model to close the gap between direct reasoning and reasoning with explicit text conversion, where the model learns to rely more equally on both text and image modalities, and doesn't require explicit text conversion for improved generalization performance.

\paragraph{Models are robust against potential errors in the prompted text conversion.} For models that are prompted to perform text conversion, we examine any negative side effects of this step. When the model does not correctly convert the image to its text format, we investigate whether to what extent the model's reasoning can be affected by the additional noises introduced by the text conversion step. Interestingly, we find that our final trained models are generally robust to such noises. On {\visualanalogy},  we find that the models trained with {\cmft}, {\contcmft}, and {\contalcmft} are still able to arrive at the correct reasoning solutions with accuracy $44.3\%$, $35.4\%$, and $63.1\%$ respectively on evaluation examples where the trained models make a mistake in text conversion.

\begin{table}[!t]
    \caption{\textbf{Ablation on explicitly prompting for text conversion:} When models are additionally prompted with ``\texttt{Convert},'' they exhibit the retained ability of text conversion. The conversion accuracy is near perfect on {\tableread}. The {\simpletohard} performance with an additional prompt ``\texttt{Convert}'' (Prompted) improves from direct inference (Direct). The improvement margin diminishes with stronger Direct performance. All evaluations are on $500$ {\hard}-image examples.}
    \label{tab:ablations_convert_helps}
    \centering
    \begin{tabular}{l|l|c c|c}
        \toprule
        Task & Supervision (Number of Training Data) & Direct & Prompted & Conversion acc \\
        \midrule
        \multirow{3}{*}{{\tableread}} & {{\cmft} ($240$k)} & {$73.2$} & {$96.6$} & {$99.2$} \\
        & {\alcmft} ($240$k) & {$87.6$} & {$98.0$} & {$100.0$}\\
        & {\contalcmft} ($240$k) & {$86.2$} & {$97.8$} & {$99.4$}\\
        \hline
        \multirow{3}{*}{{\visualanalogy}} & {{\cmft} ($120$k)} & {$35.4$} & {$51.8$} & {$47.0$} \\
        & {{\contcmft} ($120$k)} & {$55.2$} & {$62.8$} & {$49.0$} \\
        & {{\contalcmft} ($120$k)} & {$73.6$} & {$70.2$} & {$49.6$} \\
        \bottomrule
    \end{tabular}
\end{table}

\subsection{Explicit and implicit CoT}
\label{app:ablations_cot}
We use chain-of-thought (CoT) as a technique to boost the model's reasoning ability in all our experiments. In this section, we explore the role of CoT in our proposed strategies, as well as the possibility of transferring the reasoning capability from text to image modality without CoT. In \cref{tab:ablations_cot}, we report our observations on \visualanalogy{}. We note that similar observations hold for \consecutivetableread{} and \tableread{}.

\subsubsection{Removing CoT completely}
We first consider completely removing CoT from {\cmft} and observe the drop in performance measured by image {\simpletohard}. We experiment with {\cmft}, {\contcmft}, and {\contalcmft} supervision, in which we completely remove CoT from the last phase of training which has the {\cmft} supervision, while preserving the full CoT in the text warm-up \tw{} and/or reasoning alignment (\al{}) phases. 

\paragraph{Model does not learn when CoT is completely removed:} 
When CoT is completely removed from {\cmft}, performance drops to almost $0\%$ for all three types of supervision. We manually inspect the model's output and find that the generated reasoning on {\hard}-image inputs is identical to the expected behavior for \simple{} instances, which indicates that the reasoning capability on \hard{} instances failed completely to transfer from the text to image modality.

\subsubsection{Progressively internalizing CoT throughout training}
The failure above can be expected: for {\cmft} supervision, CoT may serve as a crucial technique to elicit good reasoning behaviors while for {\contcmft} and {\contalcmft}, the transition from training with full CoT to training without CoT can be too drastic for the model to adapt. Therefore, we consider a milder approach that trains the model to internalize reasoning by progressively removing CoT from the training \citep{deng2024explicit}. We train on the first $30\%$ of $12 \times 10^4$ {\cmft} examples with full CoT, the next $40\%$ of examples with progressively less CoT\footnote{split into $101$ subsets of equal length, each training with $100\%, 99\%, \cdots, 0\%$ of total characters in the CoT.}, and the last $30\%$ of examples with no CoT.

\paragraph{Internalizing CoT during the {\cmft} phase also fails:}
In this scenario, we also observe that the model completely fails on image {\simpletohard}, getting almost $0\%$ {\simpletohard} on \hard{}-image examples for all three types of supervision strategies. 

\subsubsection{Internalizing CoT during text warm-up before removing CoT completely}
We also try a variant for the multi-phase approaches, where we internalize the CoT on the text input during a text warm-up (\tw{}) stage and continue with {\cmft} with CoT completely removed.

\paragraph{CoT can be internalized on text inputs:} We internalize the CoT on the text input during a slightly modified text warm-up phase of {\contcmft}. Specifically, with $10^4$ training data that consists of an equal mix of {\simple} {\smfttext}, {\hard} {\smfttext} supervision, and Eagle instruction tuning data (randomly sampled from 1.8M examples \citep{shi2024EAGLE}), we train on the first $30\%$ examples with full CoT, the next $40\%$ examples with progressively less CoT, and the last $30\%$ examples without CoT as in previous experiments. After the warm-up phase of training, the model can achieve $97.8\%$ accuracy on {\hard}-text examples, which shows the model's ability to internalize reasoning on text inputs.

\paragraph{Explicit CoT is ``necessary'' for the internalized reasoning to transfer to image:} We then continue with the {\cmft} supervision with all CoT removed. The final trained model completely fails with $0\%$ accuracy on the {\hard}-image examples. Similarly, examining model outputs reveals that the reasoning capability on \hard{} instances failed completely to transfer from the text to the image modality. Therefore, we conclude that CoT is ``necessary'' for the cross-modal transfer of knowledge to happen in our setting.

All results testify to our claim that CoT is important in our proposed training strategies. As the techniques used to internalize or remove the CoT dependency in our experiments are very preliminary, we are not eliminating the possibility of internalizing CoT in our setting. We note that to do so may require more careful, post hoc approaches, which we leave to future work.

\begin{table}[!t]
    \caption{\textbf{Ablation on removing or internalizing CoT on {\visualanalogy}:} Preliminary attempts to completely or progressively remove CoT during the {\cmft} phase fails to generalize to \hard{} images, which shows the importance of CoT in our proposed strategies. \textit{full, none, internalizing} CoT refer to including full CoT, completely removing CoT, and progressively removing CoT respectively. `-' means the corresponding phase was not included during training. Unless specified, all evaluations are reported on {\hard}-image examples.}
    \label{tab:ablations_cot}
    \centering
    \begin{tabular}{c | c | c | c | c}
        \toprule
        \multirow{2}{*}{Type of supervision} & \multicolumn{3}{| c |}{Type of CoT} & \multirow{2}{*}{S2H accuracy (\%)} \\ 
        \cline{2-4} & {\tw{}} ($10$k) & {\al{}} ($10$k) & {\cmft} ($120$k) & \\
        \midrule
        {\cmft} & - & - & none & $0.6$ \\
        {\contcmft} & full & - & none & $0.0$ \\
        {\contalcmft} & full & full & none & $3.6$ \\
        \midrule
        {\cmft} & - & - & internalizing & $0.0$ \\
        {\contcmft} & full & - & internalizing & $0.0$ \\
        {\contalcmft} & full & full & internalizing & $3.4$ \\
        \midrule
        {\contcmft} & internalizing & - & - & $97.8$ ({\hard}-text) \\
        {\contcmft} & internalizing & - & none & $0.0$ \\
        \bottomrule
    \end{tabular}
\end{table}

\subsection{Multi-task training: jointly training on all three {\nontextgeneralizable} tasks}
\label{app:ablations_multitask}

In the main experiments, we have trained on each {\nontextgeneralizable} task separately. In this section, we explore the ablation where we combine and randomly shuffle the training data for {\tableread}, {\gridnav}, and {\visualanalogy}. In \cref{fig:multi_task}, we compare the image {\simpletohard} performance when jointly training on all $3$ tasks against training on each task separately. 

Similar to training on each task individually, the average {\simpletohard} on image across all $3$ tasks is strongest for {\cimageviatext}, followed by {\alcmft} and {\cmft}. When analyzing the effect of multi-task training on each task, we observe that it benefits the model's performance on {\tableread} and {\gridnav} but hurts performance on {\visualanalogy}. This is likely because {\tableread} and {\gridnav} are similar in nature. They are both represented by LaTeX code in the text modality, require the model to identify the current location in a table / grid, and reason about neighboring cells. On the other hand, the skills required for {\visualanalogy} are quite distinct. This suggests that the interactions between tasks during a multi-task training can also affect how much reasoning can transfer across modalities.

\begin{figure*}[t]
    \centering
    \begin{subfigure}[t]{1.0\linewidth}
        \centering
        \includegraphics[width=1.\linewidth]{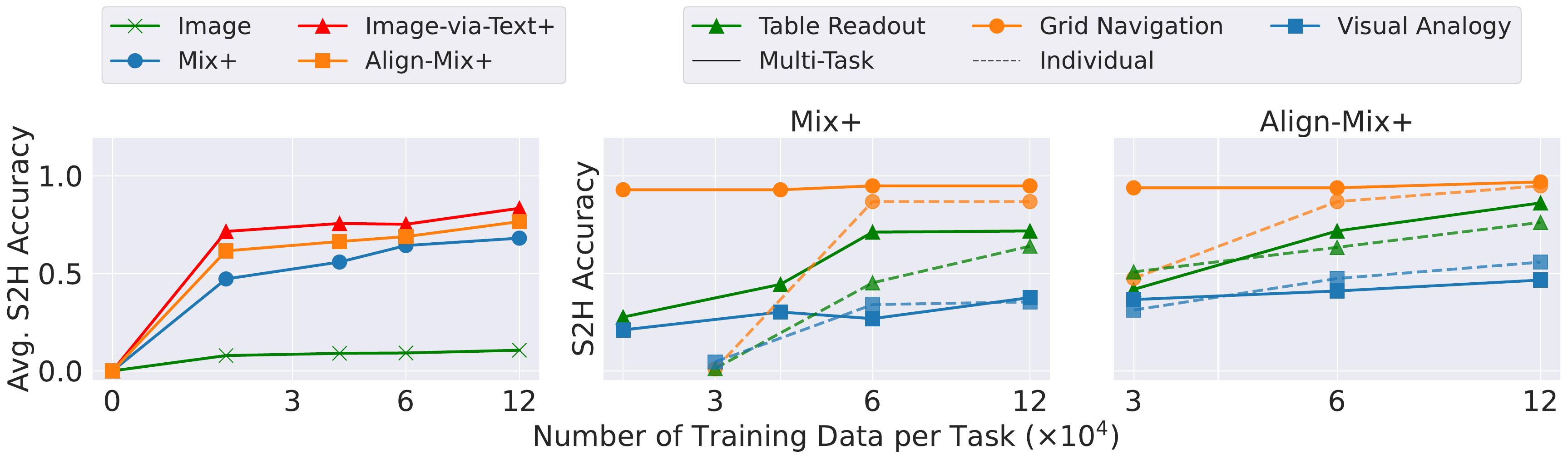}
        \end{subfigure}
    \caption{\textbf{Ablation on jointly training on all three {\nontextgeneralizable} tasks:} (Left) Average S2H Generalization on image; (Middle, Right) Comparison of Trained Jointly vs. Individually. Similar to training on each task individually, {\cmft} and {\cimageviatext} outperform {\smftimage}, and {\alcmft} matches the performance of {\cimageviatext}. Multi-task SFT boosts image {\simpletohard} for {\tableread} and {\gridnav}, while {\visualanalogy} performance remains unchanged or slightly declines, indicating task interactions drive the cross-modal transfer of reasoning capabilities in multi-task training. }
    \label{fig:multi_task}
\end{figure*}

\subsection{Ablation on repeated {\hard} examples}
\label{app:unique_samples}

\begin{figure}[t]
    \centering
    \includegraphics[width=0.7\linewidth]{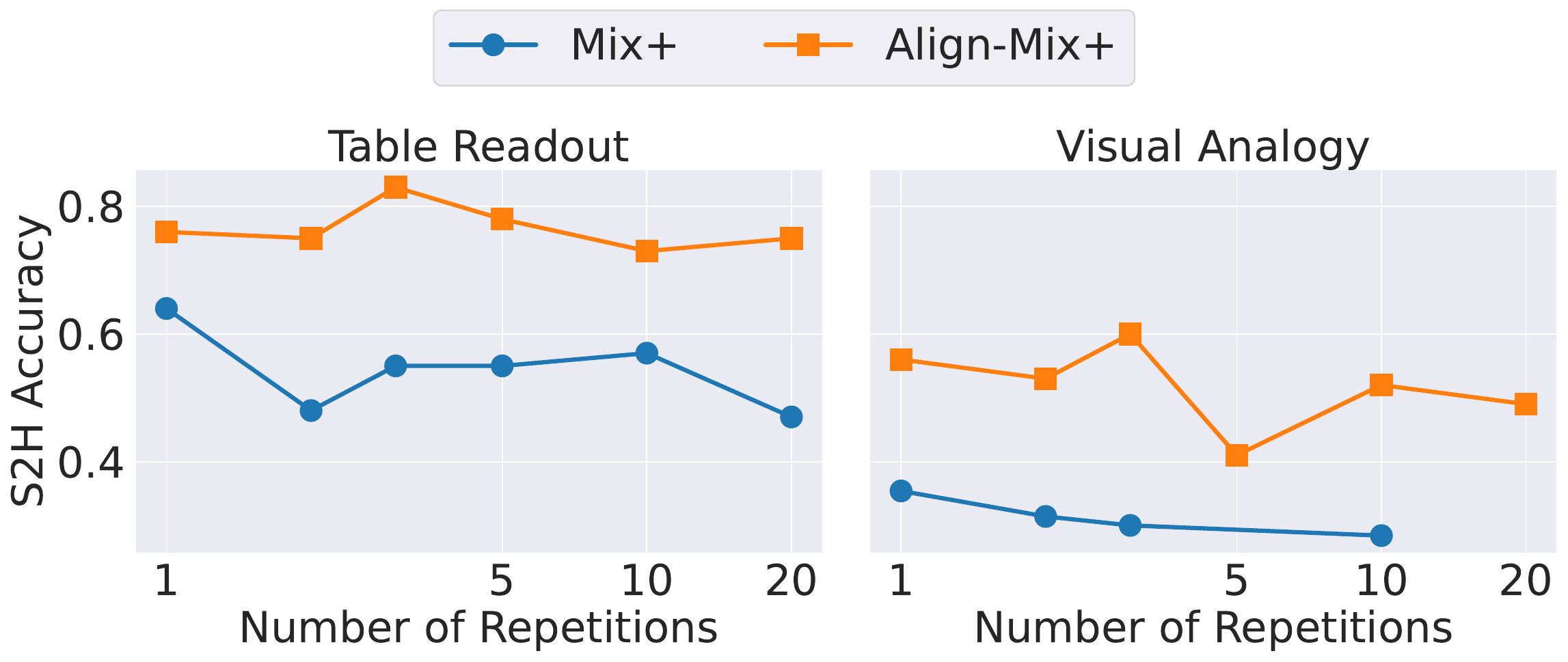}
    \caption{\textbf{Ablation on the number of repetitions of unique \hard{} examples, while maintaining the total amount of \hard{} training data, on {\tableread} and {\visualanalogy}:} Image {\simpletohard} degrades with more repetitions of {\hard} {\smfttext} examples, with the effect on {\cmft} being more drastic. Here, the amount of training data is fixed at $12 \times 10^4$, with $6 \times 10^4$ examples sampled from the \hard{} task. Interestingly, performance of {\alcmft} peaks at $3\times$ repetitions, implying the number of unique {\hard} {\smfttext} examples can be reduced by $3\times$ for {\alcmft}. }
    \label{fig:repeat_hard}
\end{figure}

In the experiments reported in the main paper (summarized in \cref{fig:main-results}), we kept all {\hard} {\smfttext} examples unique. In \cref{fig:repeat_hard}, we present the ablation where we repeat each {\hard} {\smfttext} example during training, while keeping the total number of training data fixed. Our primary observations are:

\begin{itemize}
    \item Repeating {\hard} {\smfttext} examples harms the performance of {\cmft}. Halving the number of unique {\hard} {\smfttext} examples and repeating each example $2$ times can drop the performance on {\hard}-image by at least $10\%$p on {\tableread}.
    
    \item On the other hand, {\alcmft} is quite robust to repetitions on {\tableread}. The number of unique {\hard} {\smfttext} examples can be reduced by $10\times$ (and repeating each example $10\times$) with the performance on  {\hard}-image dropping by no more than $1$-$2\%$p.
    
    \item On {\visualanalogy}, while the performance of {\alcmft} drops with large number of repetitions, the drop in performance is within $1$-$2\%$p if the number of repetitions is up to $3$.
    
    \item Interestingly, the image {\simpletohard} performance reaches its peak at exactly $3$ repetitions for {\alcmft} on both {\tableread} and {\visualanalogy}. This suggests that we may only require $3\times$ less unique {\hard} {\smfttext} examples than reported in \cref{fig:main-results}. 
\end{itemize}

\clearpage
\section{Interpretability Experiments}
\label{sec:interp_gradient}

We use gradient attribution to identify which pixel in the image is important when generating each token in the CoT. For a given data $\data \in \inputset$ and its corresponding image format $\dataimage$, we label the set of pixels in the image as $\{\dataimage_j\}$. For a given gold output $\vy = \{\reasoningtrace(\data), f(\data)\}$, we label the sequence of CoT tokens as $\{y_k\}$, where $\vy_{:k}$ refers to the subsequence of the CoT tokens, up to the $k$-th token. 

For each pixel $\dataimage_j \in \dataimage$ and each CoT token $y_k$, we compute the attribution score as:

\begin{align*}
    \text{Pixel Attribute Score:} \quad\quad  \left\langle \nabla_{\dataimage_j} \evalloss( \model(\{ \dataimage, \vy_{:k} \}), y_k), \quad \dataimage_j \right\rangle 
\end{align*}

Informally, on {\smftimage} examples, we take the gradient of the loss of the model's output (up to the $k$-th CoT token) with respect to each pixel, and project on the pixel values. Pixels that show positive alignment with the gradients are marked important for the model's prediction for the token $y_k$. 

In \cref{fig:tableread_influence_cmft}, we plot the pixel attribute values, averaged across tokens that correspond to different segments of a highlighted path of an example image from {\tableread}. We observe that {\alcmft} improves over {\cmft} models by having more focused and concise pixel attributes around the path of highlighted cells and their corresponding row/column names. In \cref{fig:ar_lossy_influence_ctalcmft}, we also show pixel attribute scores on {\visualanalogy}, where the pixel attributes are more aligned with objects of interest scattered around the grid.

\clearpage

\begin{figure*}[!ht]
    \centering
    \begin{subfigure}[t]{0.32\linewidth}
        \centering
        \includegraphics[width=1.\linewidth]{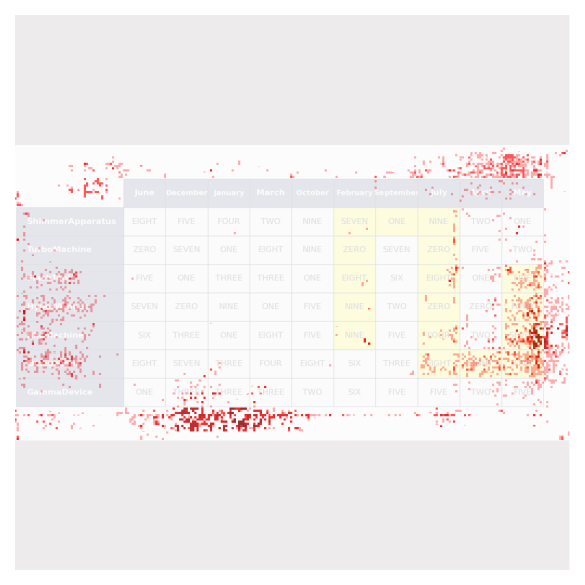}
    \end{subfigure} \hfill
    \begin{subfigure}[t]{0.32\linewidth}
        \centering
        \includegraphics[width=1.\linewidth]{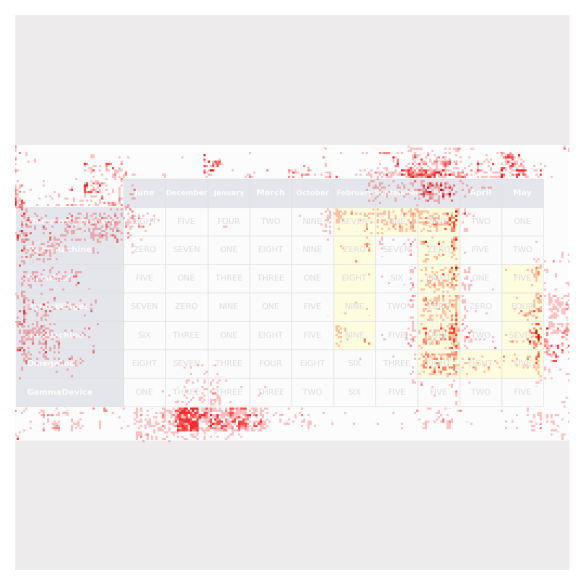}
    \end{subfigure} \hfill
    \begin{subfigure}[t]{0.32\linewidth}
        \centering
        \includegraphics[width=1.\linewidth]{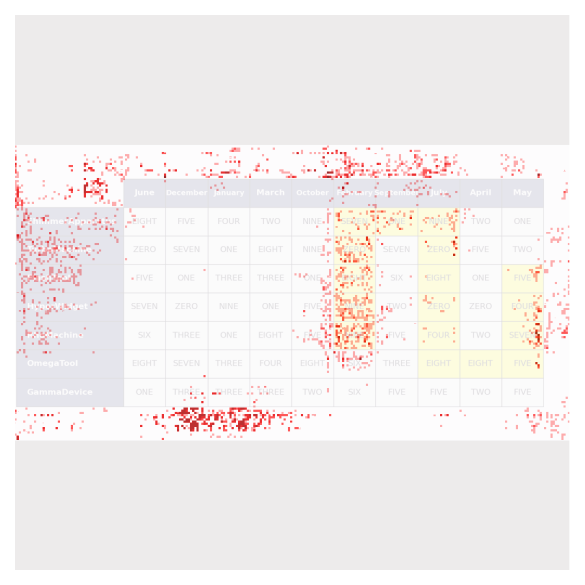}
    \end{subfigure} \hfill
    \begin{subfigure}[t]{0.32\linewidth}
        \centering
        \includegraphics[width=1.\linewidth]{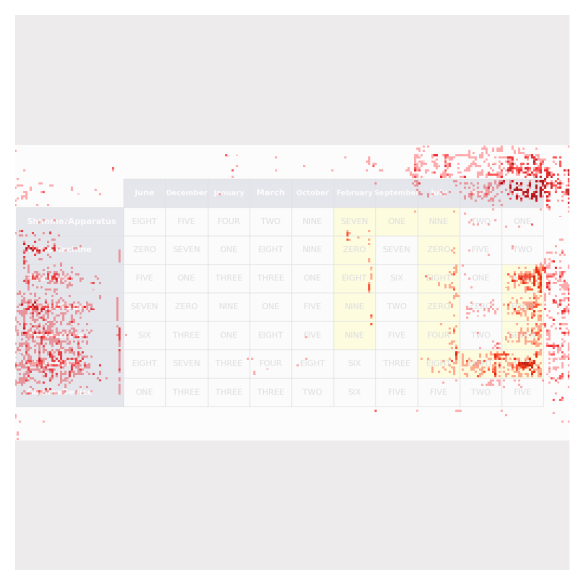}
    \end{subfigure} \hfill
    \begin{subfigure}[t]{0.32\linewidth}
        \centering
        \includegraphics[width=1.\linewidth]{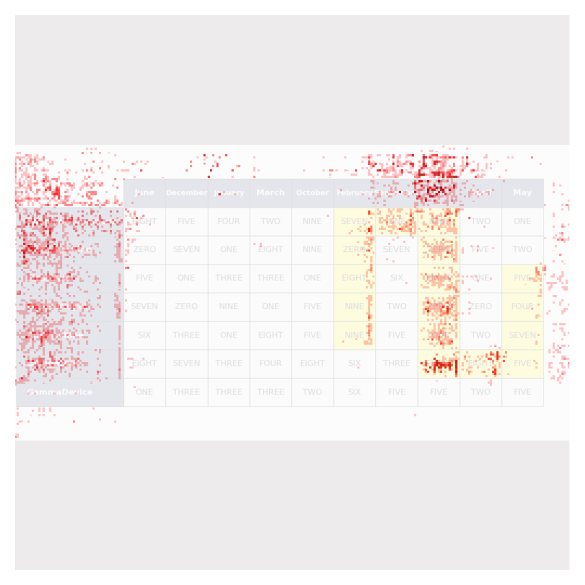}
    \end{subfigure} \hfill
    \begin{subfigure}[t]{0.32\linewidth}
        \centering
        \includegraphics[width=1.\linewidth]{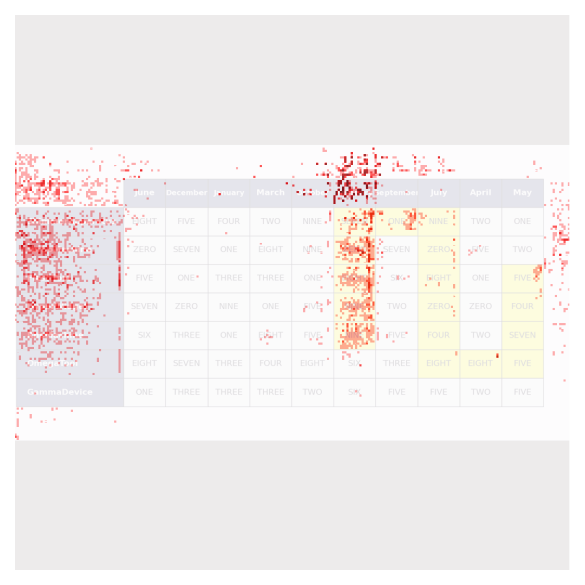}
    \end{subfigure} \hfill
    \caption{\textbf{Visualization of pixel attribute scores on {\tableread}:} (Top) {\cmft}; (Bottom) {\alcmft}. Models are trained with $24 \times 10^4$ training data. Pixel attribute scores are averaged across CoT tokens that belong to the first 5 pixels roughly in the $10$th column (left), the next 6 cells in the $8$th column (middle), and the last $6$ cells in the $6$-th column (right). We show the top-$1\%$ pixels with the highest pixel attribution scores (marked as red). {\cmft} has more diffused pixel attributions in the image, while {\alcmft} focuses more on the path of cells (and their corresponding row/column names).}
    \label{fig:tableread_influence_cmft}
\end{figure*}

\begin{figure*}[!ht]
    \centering
    \begin{subfigure}[t]{0.32\linewidth}
        \centering
        \includegraphics[width=1.\linewidth]{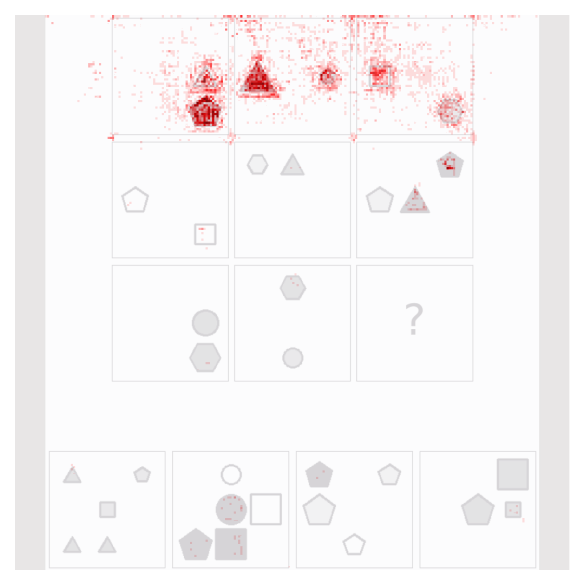}
    \end{subfigure} \hfill
    \begin{subfigure}[t]{0.32\linewidth}
        \centering
        \includegraphics[width=1.\linewidth]{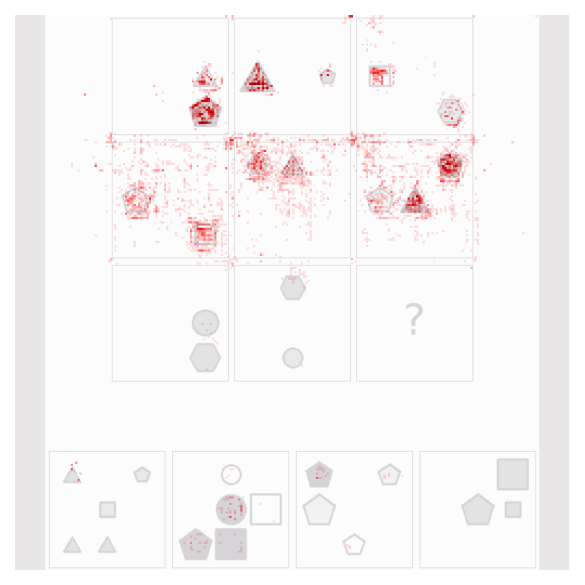}
    \end{subfigure} \hfill
    \begin{subfigure}[t]{0.32\linewidth}
        \centering
        \includegraphics[width=1.\linewidth]{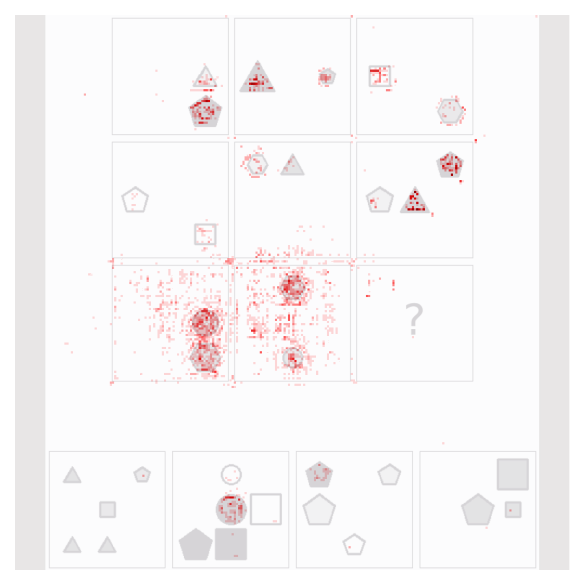}
    \end{subfigure} \hfill
    \caption{\textbf{Visualization of pixel attribute scores on {\visualanalogy}:} The model is trained with $ 12 \times 10^4$ training data of {\contalcmft}. Pixel attribute scores are averaged across CoT tokens that belong to Example 1 (left), Example 2 (middle), and the query (right) respectively. We show the top-$1\%$ pixels with the highest pixel attribution scores (marked as red). The pixel attributes are focused on relevant objects across the grid. Interestingly, when reading relevant object attributes in Example 2, the model still attends to objects from Example 1.}
    \label{fig:ar_lossy_influence_ctalcmft}
\end{figure*}

\clearpage
\section{Analysis of Failure Modes}
\label{sec:failure_modes}

In this section, we briefly discuss the common failure modes of models trained on our synthetic data, when evaluated on examples from the {\hard} split.

\subsection{\tableread}
We analyze the outputs of {\smfttext} on \hard-text, {\smftimage} on \hard-image, and {\cmft} on both \hard-text and \hard-image, where all models have been trained on $24 \times 10^4$ examples. 

Since the models perform almost perfectly on the {\simple} examples, where the total length of the sequence is around $12$, one may expect the models to read off the first $12$ numbers from tables of the {\hard} split equivalently well but start making errors after the sequence length it was trained on. We find that this is not the case by analyzing the index of the first error; i.e., how many numbers the model reads off correctly before making the first mistake. Although the average index of the first error is around $14.7$, about $56\%$ of incorrect generations (equivalently, $26\%$ of total generations) contain a mistake before the $12$th number in the sequence. 

\begin{figure*}[!ht]
    \centering
    \includegraphics[width=0.45\linewidth]{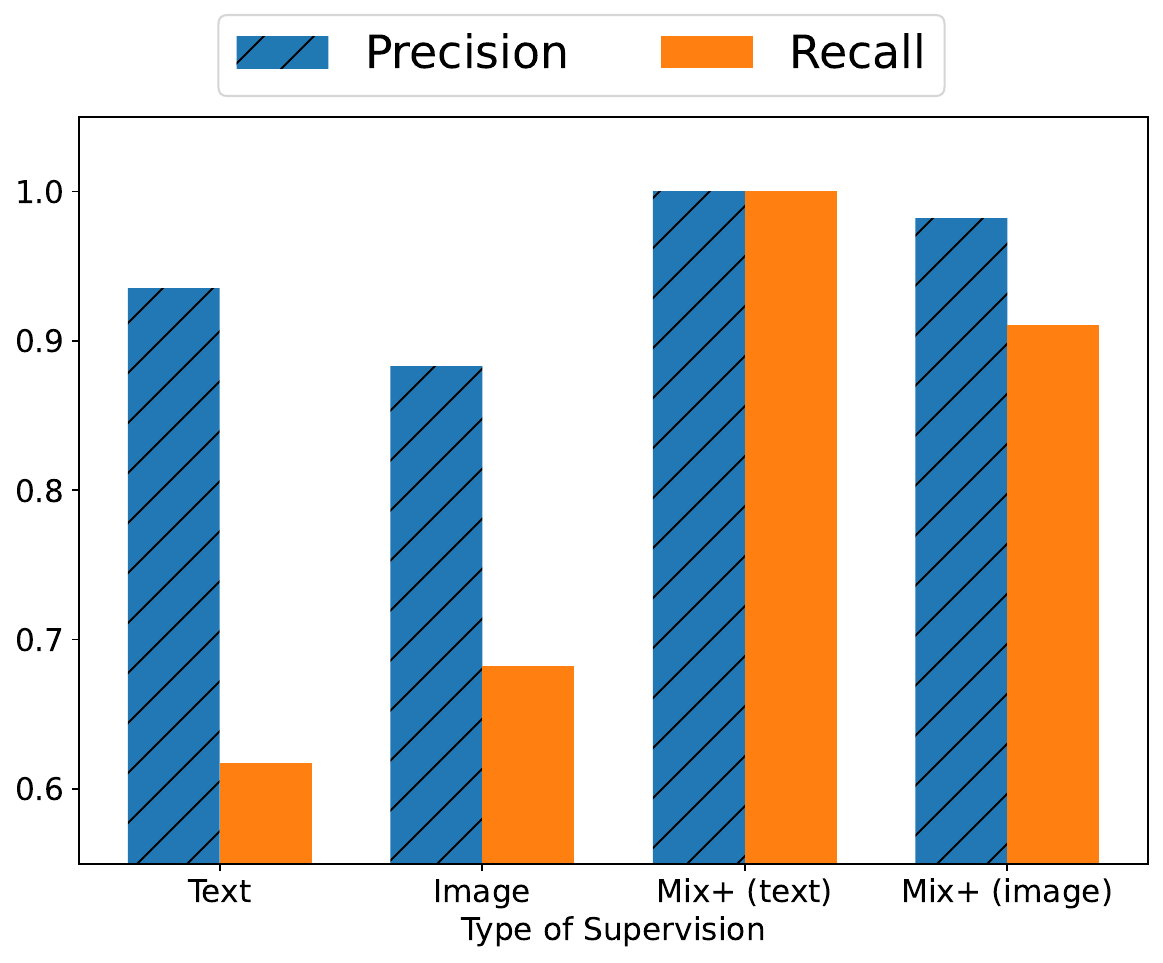}
    \includegraphics[width=0.45\linewidth]{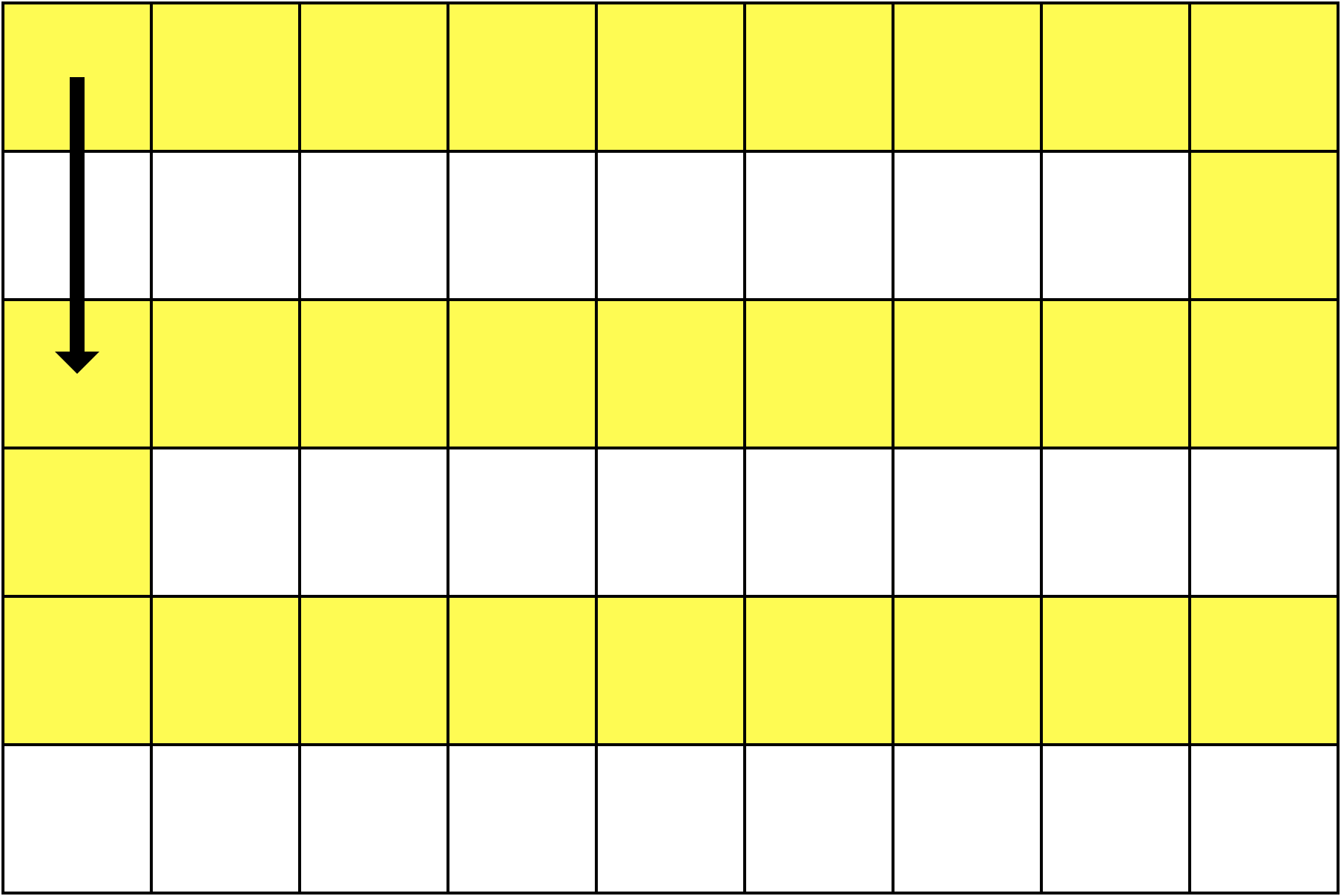}
    \caption{\textbf{Analysis of failure modes on {\tableread}:} (Left) Precision and Recall; (Right) Example of a common mistake. Models are trained on $24 \times 10^4$ examples of {\smfttext}, {\smftimage} and {\cmft} supervision and evaluated on corresponding inputs from \hard. Models often hallucinate a ``shortcut.'' In this case, precision would be $12/13$ and recall would be $12/29$.}
    \label{fig:table_readout_failure}
\end{figure*}

To further analyze the behavior of the model when it makes a mistake, we extend the definition of precision and recall:
\begin{equation*}
    \text{Precision } = \frac{\text{Total \# correctly listed}}{\text{Total \# listed}} \quad \quad \text{Recall } = \frac{\text{Total \# correctly listed}}{\text{Total \# highlighted}}
\end{equation*}
where we take the sum in the numerator and denominator across all test examples and mark a cell as correctly listed only if the model generation contains it, regardless of the exact position in the sequence. See left of \cref{fig:table_readout_failure} for the evaluation results. Note that for {\smfttext} and {\smftimage}, precision is significantly higher than recall, meaning that it rarely hallucinates that a cell is highlighted (when it is not), but it fails to list off many of the numbers that were highlighted. We find that this is mainly because once the model derails from the highlighted path, it just moves directly towards the destination cell, until it rejoins the path, unintentionally creating a ``shortcut'' that skips around 15 cells on the original path on average. See right of \cref{fig:table_readout_failure} for a visualization. However, the recall improves significantly on both {\hard}-text and {\hard}-image when trained with {\cmft}.

\subsection{\gridnav}
In \cref{fig:GW_failure}, we analyze the outputs of {\smfttext} supervision on \hard-text, {\smftimage} supervision on \hard-image, and {\cmft} supervision on \hard-image, where all models have been trained on a varying number of examples. 

A successful evaluation on {\gridnav} requires completing multiple intermediate subtasks. The model first needs to correctly identify the source and destination cells from the grid and parse the row/column indices. We observe that the models can easily learn this subtask. Under any of the three types of supervision, the model can get at least $98\%$ accuracy on parsing the location of the source and destination cells with only $1.5 \times 10^4$ examples. With $6 \times 10^4$ or more examples, the accuracy is always $100\%$. 

Next, we analyze whether the model returns a sequence of actions that leads from the source to the destination (ignoring any object or obstacle). We observe that there is some ``phase transition'' at $3 \times 10^4$ examples, where the model's accuracy on this subtask increases sharply. However, whereas {\cmft}  continues to improve accuracy on this subtask, exceeding $90\%$ at $6 \times 10^4$ examples, {\smfttext} and {\smftimage} supervision fail to achieve $90\%$ even with $24 \times 10^4$ examples. 

We then analyze the average fraction of objects collected while navigating the grid. The evaluation on this subtask also follows a similar ``phase transition'' at $3 \times 10^4$ examples. However, whereas {\cmft} immediately achieves $90\%$ at $3 \times 10^4$ examples and continues to improve to $96\%$ at $24 \times 10^4$ examples, {\smfttext} and {\smftimage} supervision fail to improve beyond $50$-$70\%$. This subtask becomes a strong bottleneck for {\smfttext} and {\smftimage} supervision which prevents them from improving {\simpletohard} performance. 

Finally, we analyze the average number of obstacles that the model passes through. Across any of the three types of supervision, the metric improves with more training data. However, this metric drops as low as 0.12 for {\cmft} at $24 \times 10^4$ examples, whereas {\smfttext} supervision only achieves 0.78 and {\smftimage} supervision achieves 0.67.

\begin{figure*}[!ht]
    \centering
    \includegraphics[width=0.9\linewidth]{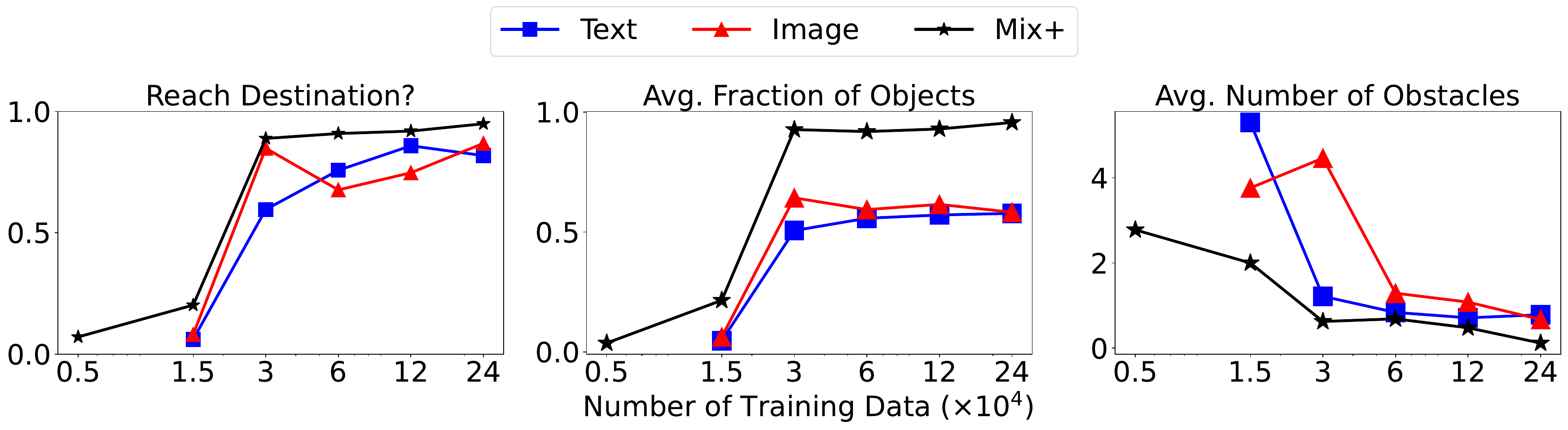}
    \caption{\textbf{Analysis of failure modes on {\gridnav}:} (Left) Whether model generates a sequence of actions that leads to the destination; (Middle) Average fraction of objects collected; (Right) Average number of obstacles passed through. Models trained with {\smfttext} and {\smftimage} fail to improve beyond a certain threshold for all three subtasks.}
    \label{fig:GW_failure}
\end{figure*}

\subsection{\visualanalogy}
We analyze the outputs of {\smfttext} on \hard-text, {\smftimage} on \hard-image, and {\cmft} on both \hard-text and \hard-image, where all models have been trained on $12 \times 10^4$ examples. Specifically, we analyze the CoT trace, focusing on the following structural steps as introduced in \cref{app:eval_visual_analogy} earlier:
\begin{itemize}
    \item To reason about examples:
    \begin{enumerate}
        \item given an attribute (e.g. \texttt{shape\_type}), the model first needs to correctly enumerate the attribute values (e.g. \texttt{circle}) for each image in the examples;
        \item the model then needs to decide whether the values in all three images of that example are consistent with a logical relation (e.g. \texttt{XOR});
        \item after repeating the process for both in-context examples, the model summarizes the two relational patterns $(d_1,r_1)$ and $(d_2,r_2)$ for the examples;
        \item finally, the model needs to identify the target relation $r_1=r_2=r_\text{query}$ from the examples.
    \end{enumerate}
    \item To reason about the query: the model needs to correctly enumerate the attribute values for each image in the query similarly.
    \item To reason about the options:
    \begin{enumerate}
        \item assuming the query when combined with each option follows a relational pattern (domain $d$, relation $r$) (e.g. (\texttt{line type}, \texttt{XOR}), the model needs to identify the correct values of the attribute domain $d$ for each option image and the correct relation $r$;
        \item the model also needs to reason whether the identified relation $r$ is the desired target relation $r_\text{query}$.
    \end{enumerate}
\end{itemize}

\paragraph{{\cmft} supervision enables significant improvement on reasoning steps that require compositional generalization where {\smfttext} and {\smftimage} supervision fail:} As shown in \cref{tab:visual_analogy_failure_modes}, we observe that models trained with {\smfttext} and {\smftimage} struggle primarily to identify the correct held-out relational pattern $(d_i,r_i)$ for in-context examples, and in particular to recognize $d_1\ne d_2$, that is, the two examples vary along \emph{different} attributes, with both error rates $\ge 80\%$. These two sources of error correspond exactly to the differences between the {\simple} and {\hard} split of {\visualanalogy}, which requires the model to generalize in a compositional manner. With {\cmft} supervision, the model significantly improves on these steps with a much smaller error rate of $42.2\%$ in identifying the held-out $(d_i,r_i)$ and $23.8\%$ in recognizing $d_1\ne d_2$. 

\paragraph{{\visualanalogy} focuses more on abstract relational reasoning rather than object detection:} We observe that even with a consistently higher error rate in identifying attribute values, models with {\cmft} supervision can achieve a lower error rate in both CoT and exact match compared to their counterparts with {\smfttext} and {\smftimage} supervision. This makes sense since reasoning depends more on identifying the correct \emph{logical relation} than on identifying the correct \emph{attribute values}. Although achieving the latter can be an important reasoning step, it is not a necessary condition to arrive at the correct solution.

We also note that the error rate of CoT can be higher than the error rate in exact match. This indicates that in some cases the model can still arrive at the correct solution even though it makes slight mistakes in the reasoning trace: for example, it can still conclude with the correct relational pattern without identifying all the attribute values correctly.

\paragraph{Even with {\cmft} supervision, the model still exhibits sensitivity to CoT templates and hallucinations:} Interestingly, we find that the error rate in identifying values of \texttt{size}, \texttt{quantity}, and \texttt{position} consistently similar. Upon manual inspection of the model output, we find that models fail to switch between different reasoning templates about shape and line objects: while the general templates for the two object types are similar, the model needs to reason about five attributes for shapes and only \texttt{type} and \texttt{color} for lines. With {\cmft} supervision, models can still be sensitive to this small difference in CoT templates and hallucinate about undefined \texttt{size}, \texttt{quantity}, and \texttt{position} attributes of the line objects. This highlights that models with {\cmft} supervision are still brittle.

\begin{table}[!t]
    \caption{\textbf{Analysis of failure modes on {\visualanalogy}:} Models are trained on $12 \times 10^4$ examples of {\smfttext}, {\smftimage}, and {\cmft} supervision and evaluated on corresponding {\hard} inputs. $^*$ means the evaluation is considered in-domain, as {\cmft} supervision contains {\hard}-text examples in training. To evaluate the entire CoT (second to last row), we check if the generated output contains all the correct values in reasoning steps about the examples, query, and options. The main sources of errors for each type of supervision are highlighted.}
    \label{tab:visual_analogy_failure_modes}
    \centering
    \begin{tabular}{c c | c | c | c}
        \toprule
        \multicolumn{2}{c |}{\multirow{2}{*}{Types of failures}} & \multicolumn{3}{| c }{Error rate ($\%$)} \\ 
        \cline{3-5} & & {\smfttext} (text) & {\smftimage} (image) & {\cmft} (text$^*$ / image) \\
        \midrule
        \multirow{8}{*}{Reasoning about examples} & type values & $0.0$ & $0.0$ & $0.0$ / $0.0$ \\ & color values & $0.0$ & $0.0$ & $0.0$ / $0.0$ \\ & size values & $29.6$ & $26.6$ & $0.0$ / $39$ \\ & quantity values & $29.6$ & $26.2$ & $0.0$ / $37.4$ \\ & position values & $29.6$ & $26.2$ & $0.0$ / $37.4$ \\ & held-out $(d_i,r_i)$ & $\mathbf{86.8}$ & $\mathbf{81.0}$ & $0.0$ / $42.2$ \\ & $d_1\ne d_2$ & $\mathbf{86.8}$ & $\mathbf{80.8}$ & $0.0$ / $23.8$ \\ & relation & $35.2$ & $34.4$ & $0.0$ / $0.8$ \\
        \midrule
        \multirow{5}{*}{Reasoning about query} & type values & $0.0$ & $0.0$ & $0.0$ / $0.0$ \\ & color values & $0.0$ & $0.0$ & $0.0$ / $0.0$ \\ & size values & $0.4$ & $7.6$ & $0.0$ / $16.8$ \\ & quantity values & $0.4$ & $7.6$ & $0.0$ / $16.2$ \\ & position values & $0.4$ & $7.6$ & $0.0$ / $16.2$ \\
        \midrule
        \multirow{4}{*}{Reasoning about options} & attribute domain & $79.8$ & $65.2$ & $0.4$ / $44.2$ \\ & attribute values & $8.4$ & $32.0$ & $0.0$ / $\mathbf{65.0}$ \\ & relation & $71.4$ & $\mathbf{82.4}$ & $0.2$ / $45.0$ \\ & identify solution & $45.2$ & $51.8$ & $0.2$ / $21.0$ \\
        \midrule
        \multicolumn{2}{c |}{CoT} & $100.0$ & $100.0$ & $0.4$ / $79.8$ \\
        \midrule
        \multicolumn{2}{c |}{Exact match} & $100.0$ & $100.0$ & $0.4$ / $64.6$ \\
        \bottomrule
    \end{tabular}
\end{table}

\clearpage

\begin{figure}[!h]
    \centering
    \includegraphics[scale=0.28]{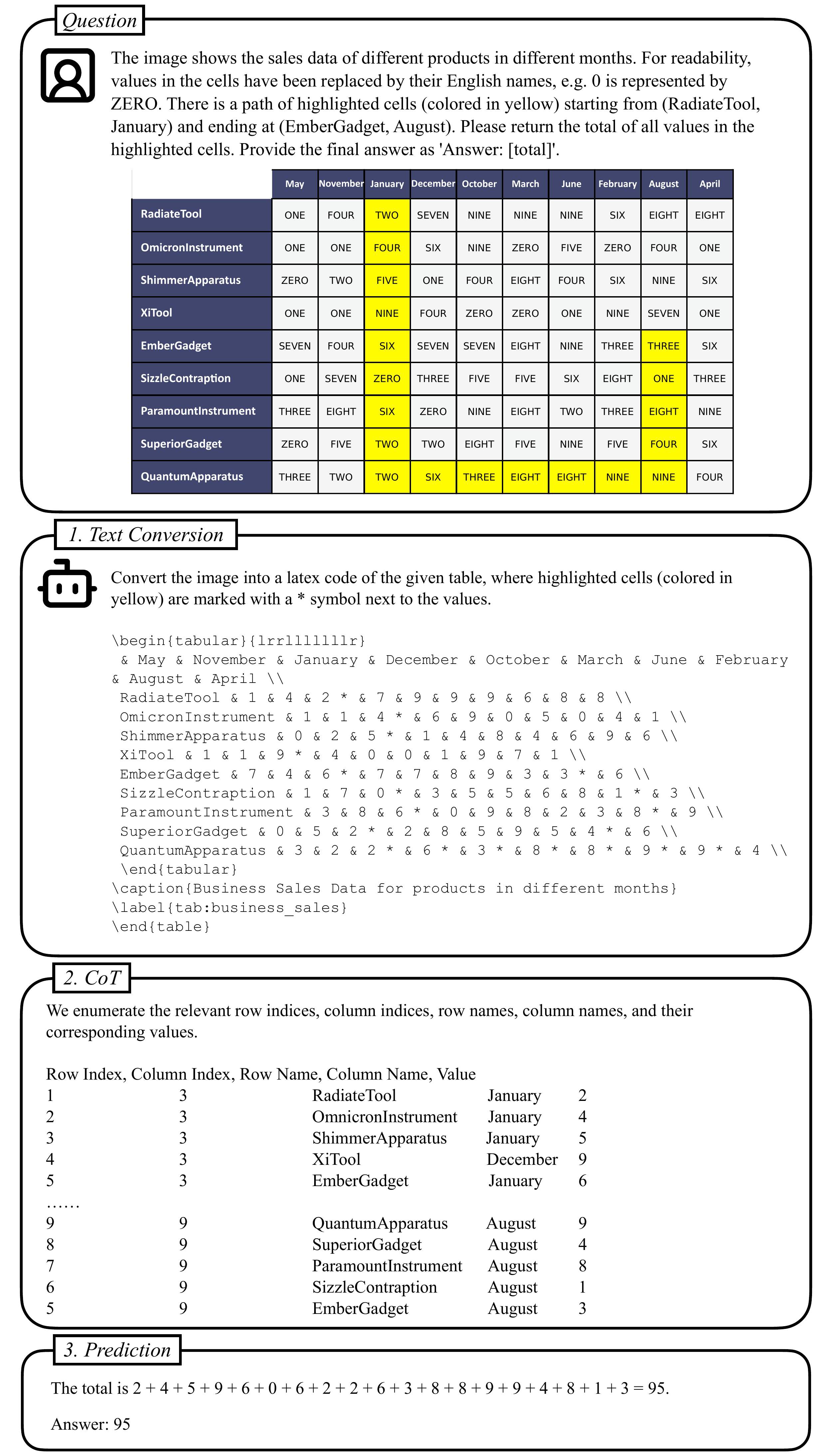}
    \caption{\textbf{A {\simple} example from {\tableread}}.}
    \label{fig:table-id}
\end{figure}

\begin{figure}[!h]
    \centering
    \includegraphics[scale=0.28]{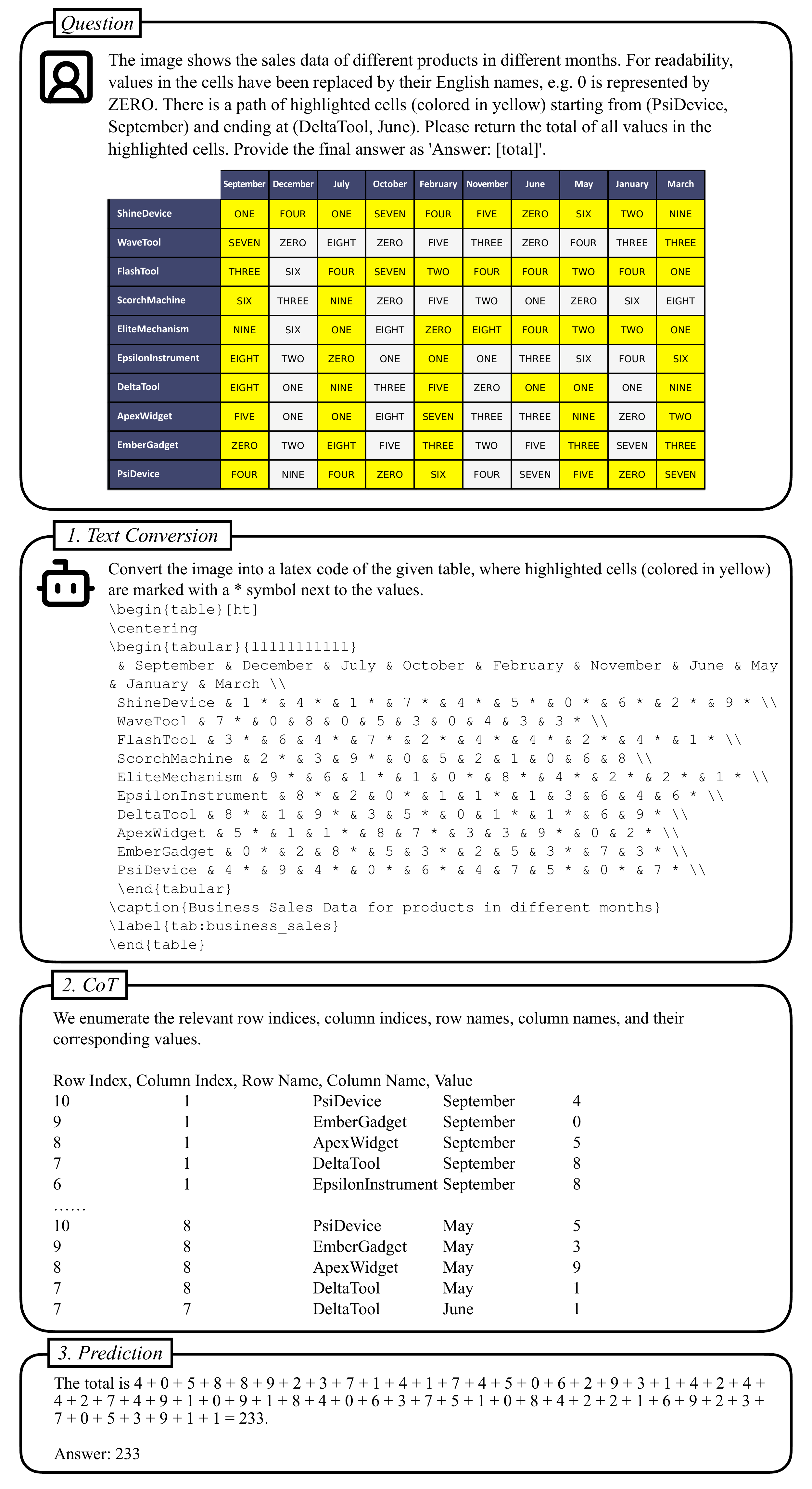}
    \caption{\textbf{A {\hard} example from {\tableread}}.}
    \label{fig:table-ood}
\end{figure}

\begin{figure}[!h]
    \centering
    \includegraphics[scale=0.32]{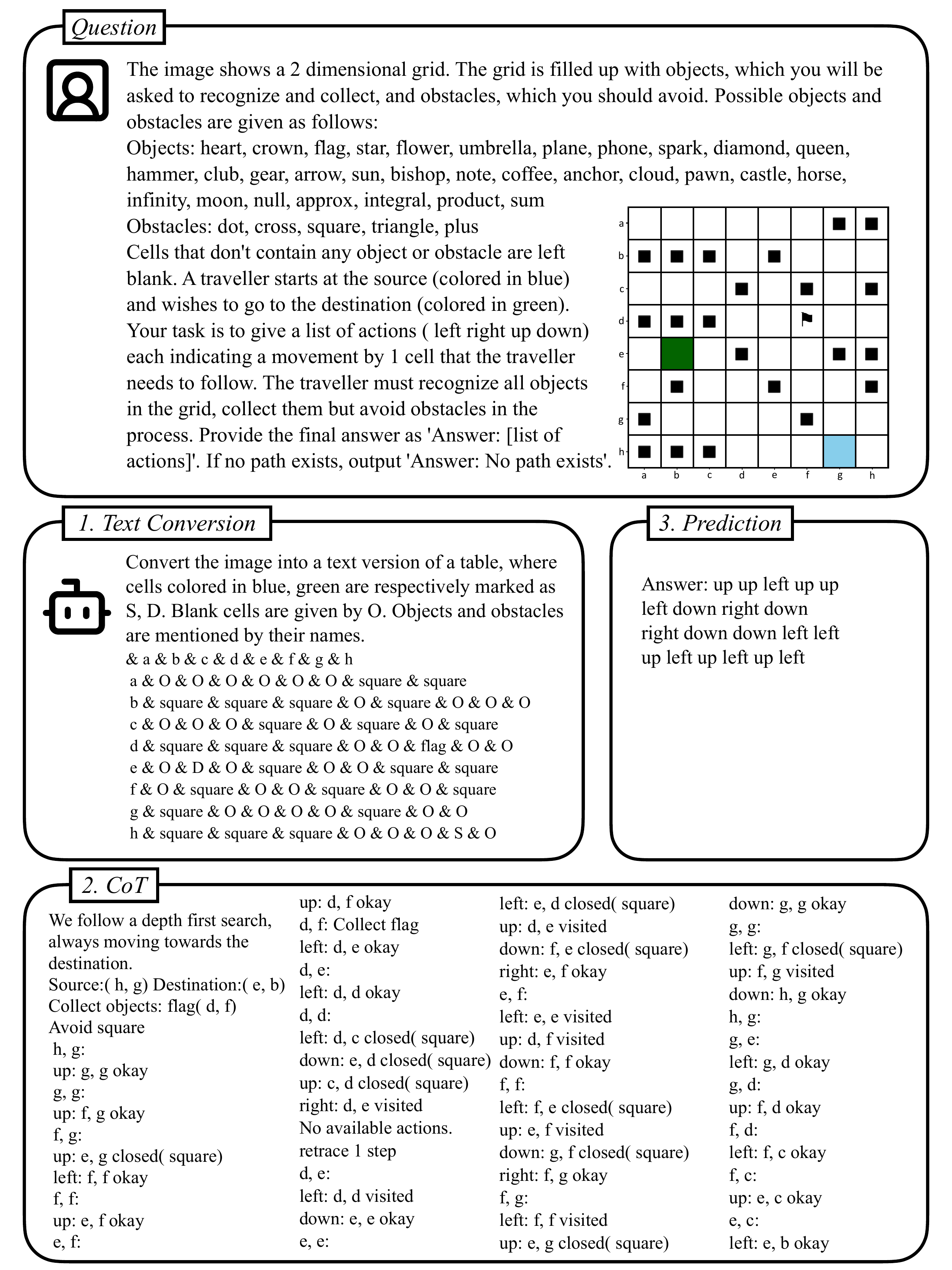}
    \caption{\textbf{A {\simple} example from {\gridnav}}.}
    \label{fig:grid-id}
\end{figure}

\begin{figure}[!h]
    \centering
    \includegraphics[scale=0.31]{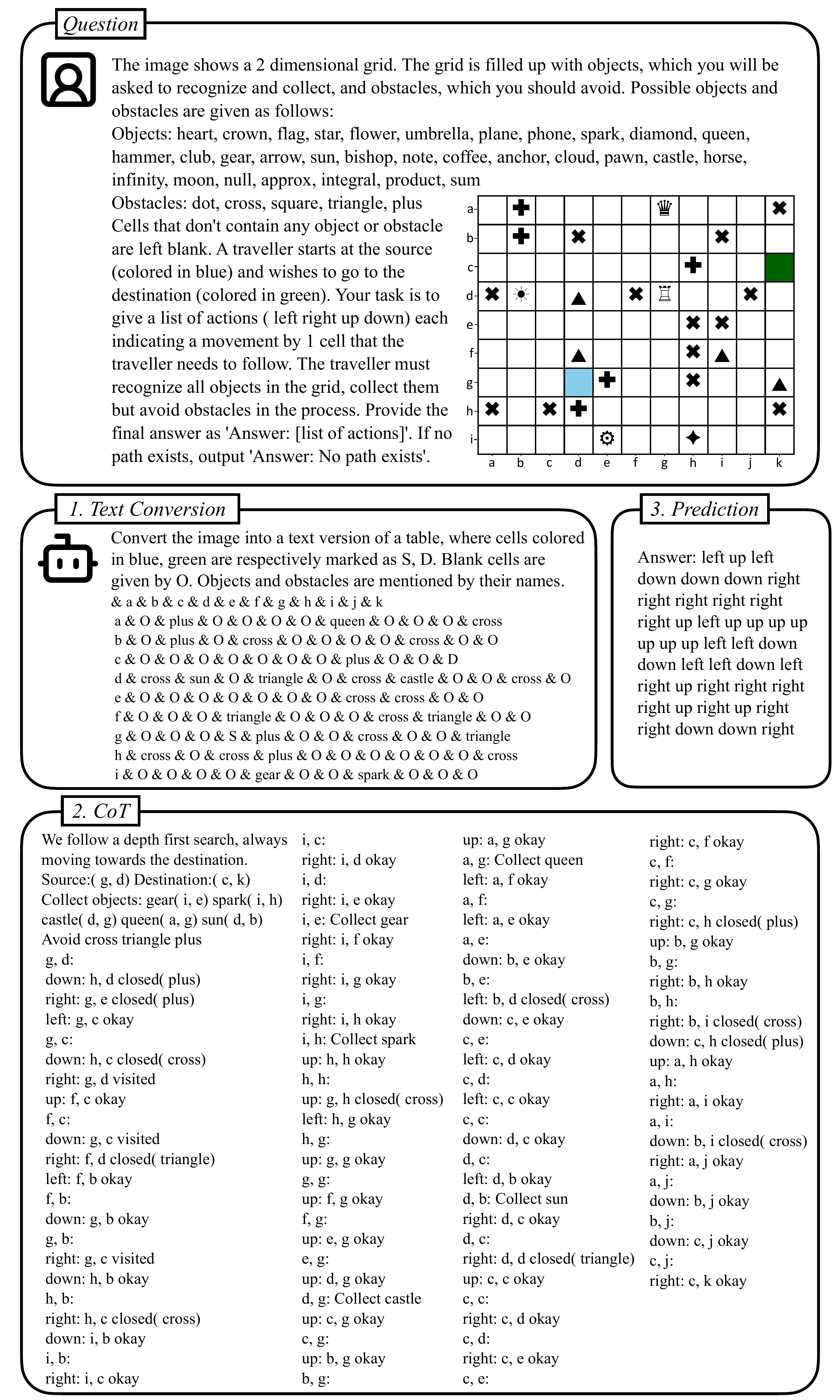}
    \caption{\textbf{A {\hard} example from {\gridnav}}.}
    \label{fig:grid-ood}
\end{figure}

\begin{figure}[!h]
    \centering
    \includegraphics[scale=0.32]{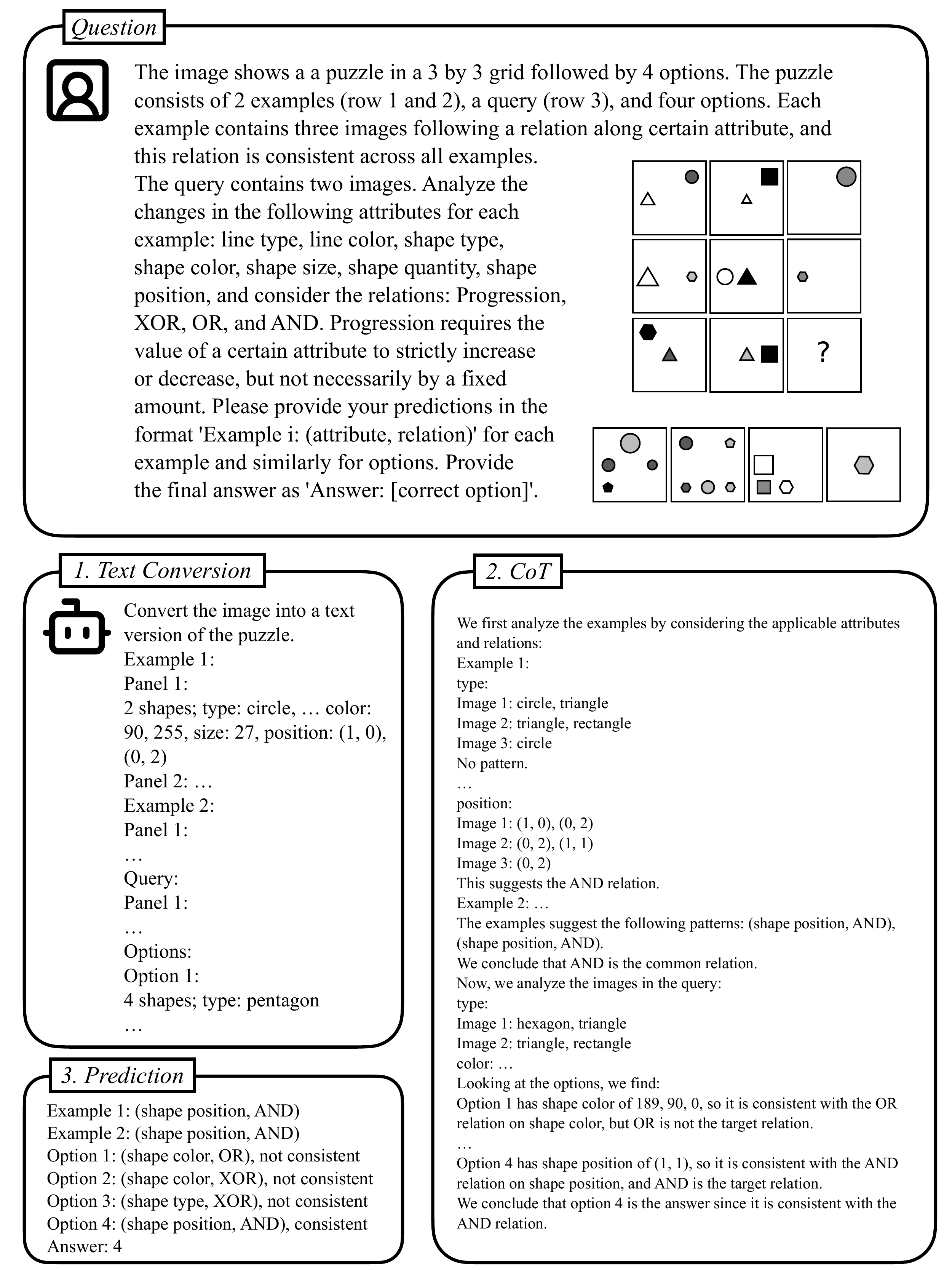}
    \caption{\textbf{A {\simple} example from {\visualanalogy}:} The common relation is $r = \texttt{AND}$ and the domains are $d_1 = d_2 = d_\text{query} = \texttt{shape quantity}$, and the combinations $(d, r)$ are not in the held-out set $\mathcal{S}= \{(\texttt{line type}, \texttt{XOR}),$ $(\texttt{line color}, \texttt{OR}),$ $(\texttt{shape type}, \texttt{AND}),$ $(\texttt{shape size}, \texttt{XOR}),$
    $(\texttt{shape color}, \texttt{Progression}),$ $(\texttt{shape position}, \texttt{OR}),$ $(\texttt{line type}, \texttt{AND}),$ $(\texttt{line color}, \texttt{Progression})\}$.}
    \label{fig:ar-domain-transfer-heldout-id}
\end{figure}

\begin{figure}[!h]
    \centering
    \includegraphics[scale=0.32]{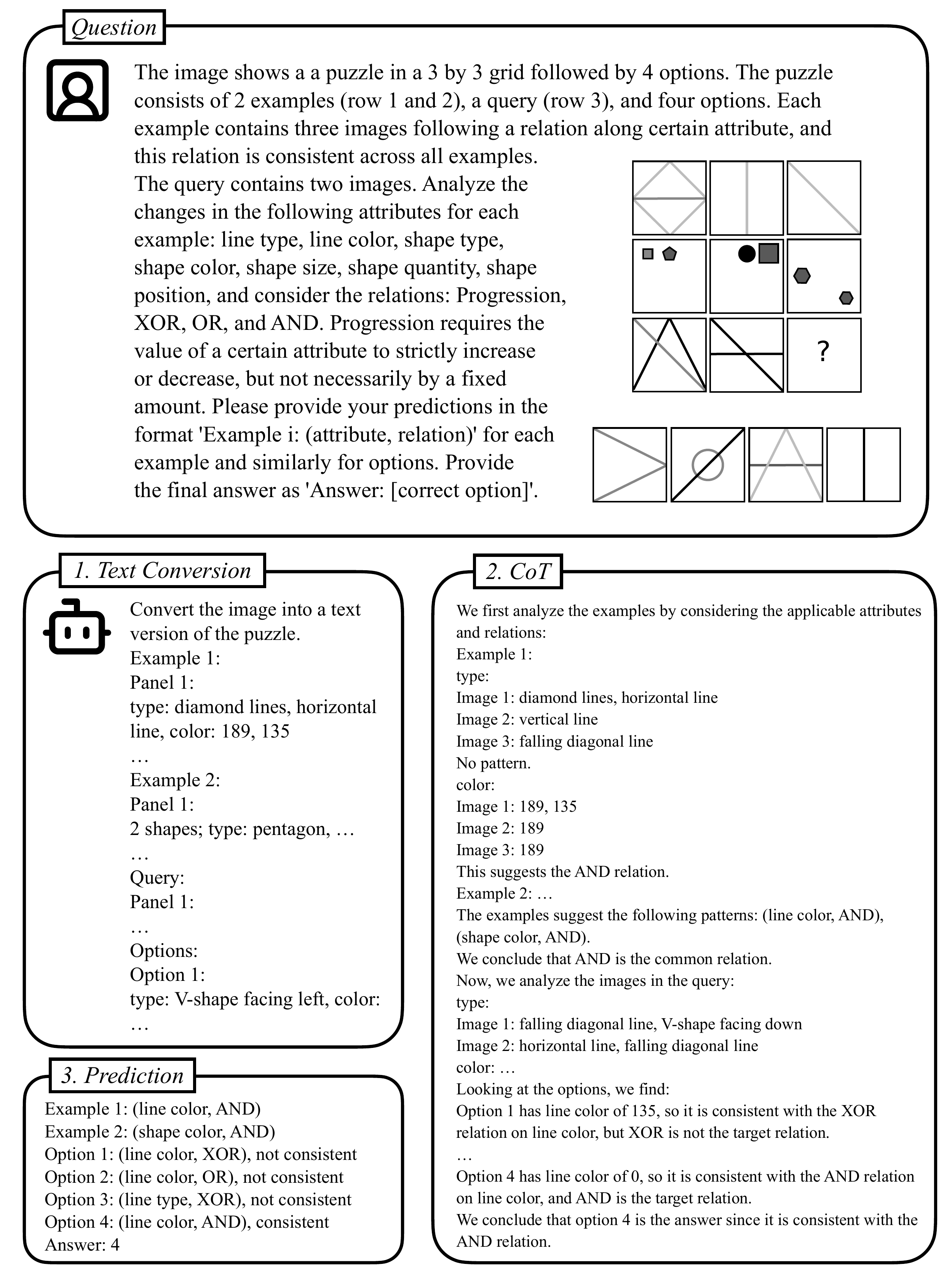}
    \caption{\textbf{A {\hard} example from {\visualanalogy}:} 
    The common relation is $r = \texttt{AND}$ and the domains are distinct: $d_1 = \texttt{line color}$, $d_2 =\texttt{shape position}$, $d_\text{query} = \texttt{line color}$, and the combinations $(d, r)$ are in the held-out set $\mathcal{S}= \{(\texttt{line type}, \texttt{XOR}),$ $(\texttt{line color}, \texttt{OR}),$ $(\texttt{shape type}, \texttt{AND}),$ $(\texttt{shape size}, \texttt{XOR}),$
    $(\texttt{shape color}, \texttt{Progression}),$ $(\texttt{shape position}, \texttt{OR}),$ $(\texttt{line type}, \texttt{AND}),$ $(\texttt{line color}, \texttt{Progression})\}$. Note that the pattern for the confounding options may not be in $\mathcal{S}$.}
    \label{fig:ar-domain-transfer-heldout-ood}
\end{figure}

\end{document}